\documentclass[10pt,journal,compsoc]{IEEEtran}
\usepackage{ifpdf}
\ifCLASSOPTIONcompsoc
  \usepackage[nocompress]{cite}
\else
  \usepackage{cite}
\fi
\ifCLASSINFOpdf
\else
\fi
\usepackage{microtype}
\usepackage{graphicx}
\usepackage{overpic}
\usepackage{subfigure}
\usepackage{booktabs} % for professional tables
\usepackage{multirow}
\usepackage{hyperref}
\usepackage{amsmath}
\usepackage{amssymb}
\usepackage{mathtools}
\usepackage{amsthm}
\theoremstyle{plain}
\newtheorem{theorem}{Theorem}[section]

\newtheorem{lemma}[theorem]{Lemma}
\newtheorem{corollary}[theorem]{Corollary}
\theoremstyle{definition}

\newtheorem{assumption}[theorem]{Assumption}
\newtheorem{remark}[theorem]{Remark}
\usepackage{algorithm}
\usepackage{algorithmic}
\usepackage{multirow}
\usepackage{threeparttable}
\begin{document}
%
% paper title
% Titles are generally capitalized except for words such as a, an, and, as,
% at, but, by, for, in, nor, of, on, or, the, to and up, which are usually
% not capitalized unless they are the first or last word of the title.
% Linebreaks \\ can be used within to get better formatting as desired.
% Do not put math or special symbols in the title.
\title{Efficient Federated Learning via Local Adaptive Amended Optimizer with Linear Speedup
}
%
%
% author names and IEEE memberships
% note positions of commas and nonbreaking spaces ( ~ ) LaTeX will not break
% a structure at a ~ so this keeps an author's name from being broken across
% two lines.
% use \thanks{} to gain access to the first footnote area
% a separate \thanks must be used for each paragraph as LaTeX2e's \thanks
% was not built to handle multiple paragraphs
%
%
%\IEEEcompsocitemizethanks is a special \thanks that produces the bulleted
% lists the Computer Society journals use for "first footnote" author
% affiliations. Use \IEEEcompsocthanksitem which works much like \item
% for each affiliation group. When not in compsoc mode,
% \IEEEcompsocitemizethanks becomes like \thanks and
% \IEEEcompsocthanksitem becomes a line break with idention. This
% facilitates dual compilation, although admittedly the differences in the
% desired content of \author between the different types of papers makes a
% one-size-fits-all approach a daunting prospect. For instance, compsoc 
% journal papers have the author affiliations above the "Manuscript
% received ..."  text while in non-compsoc journals this is reversed. Sigh.

\author{Yan Sun, Li Shen, Hao Sun, Liang Ding, and Dacheng Tao,~\IEEEmembership{Fellow,~IEEE}% <-this % stops a space
\IEEEcompsocitemizethanks{
\IEEEcompsocthanksitem Prof Dacheng Tao is partially supported by Australian Research Council Project FL-170100117.
\IEEEcompsocthanksitem Yan Sun is with the University of Sydney, Australia. 
\(Email: woodenchild95@outlook.com\)

\IEEEcompsocthanksitem Li Shen is with the JD Explore Academy, China.
\(Email: mathshenli@gmail.com\).

\IEEEcompsocthanksitem Hao Sun is with the University of Science and Technology of China, China.
\(Email: ustcsh@mail.ustc.edu.cn\).

\IEEEcompsocthanksitem Liang Ding is with the JD Explore Academy, China 
\(Email: liangding.liam@gmail.com\).

\IEEEcompsocthanksitem Dacheng Tao is with the University of Sydney, Australia. 
Email: dacheng.tao@gmail.com  

\IEEEcompsocthanksitem Corresponding author: Li Shen
}% <-this % stops an unwanted space
\thanks{Manuscript received April 19, 2005; revised August 26, 2015.}}

% note the % following the last \IEEEmembership and also \thanks - 
% these prevent an unwanted space from occurring between the last author name
% and the end of the author line. i.e., if you had this:
% 
% \author{....lastname \thanks{...} \thanks{...} }
%                     ^------------^------------^----Do not want these spaces!
%
% a space would be appended to the last name and could cause every name on that
% line to be shifted left slightly. This is one of those "LaTeX things". For
% instance, "\textbf{A} \textbf{B}" will typeset as "A B" not "AB". To get
% "AB" then you have to do: "\textbf{A}\textbf{B}"
% \thanks is no different in this regard, so shield the last } of each \thanks
% that ends a line with a % and do not let a space in before the next \thanks.
% Spaces after \IEEEmembership other than the last one are OK (and needed) as
% you are supposed to have spaces between the names. For what it is worth,
% this is a minor point as most people would not even notice if the said evil
% space somehow managed to creep in.

% The paper headers
\markboth{Journal of \LaTeX\ Class Files,~Vol.~14, No.~8, August~2015}%
{Shell \MakeLowercase{\textit{Sun et al.}}: FedLADA: Towards Efficient Federated Learning via Local Adaptive Amended Optimizer with Linear Speedup
}
% The only time the second header will appear is for the odd numbered pages
% after the title page when using the twoside option.
% 
% *** Note that you probably will NOT want to include the author's ***
% *** name in the headers of peer review papers.                   ***
% You can use \ifCLASSOPTIONpeerreview for conditional compilation here if
% you desire.

% The publisher's ID mark at the bottom of the page is less important with
% Computer Society journal papers as those publications place the marks
% outside of the main text columns and, therefore, unlike regular IEEE
% journals, the available text space is not reduced by their presence.
% If you want to put a publisher's ID mark on the page you can do it like
% this:
%\IEEEpubid{0000--0000/00\$00.00~\copyright~2015 IEEE}
% or like this to get the Computer Society new two part style.
%\IEEEpubid{\makebox[\columnwidth]{\hfill 0000--0000/00/\$00.00~\copyright~2015 IEEE}%
%\hspace{\columnsep}\makebox[\columnwidth]{Published by the IEEE Computer Society\hfill}}
% Remember, if you use this you must call \IEEEpubidadjcol in the second
% column for its text to clear the IEEEpubid mark (Computer Society jorunal
% papers don't need this extra clearance.)

% use for special paper notices
%\IEEEspecialpapernotice{(Invited Paper)}

% for Computer Society papers, we must declare the abstract and index terms
% PRIOR to the title within the \IEEEtitleabstractindextext IEEEtran
% command as these need to go into the title area created by \maketitle.
% As a general rule, do not put math, special symbols or citations
% in the abstract or keywords.

\IEEEtitleabstractindextext{
\begin{abstract}
Adaptive optimization has achieved notable success for distributed learning while extending adaptive optimizer to federated Learning (FL) suffers from severe inefficiency, including (i) rugged convergence due to inaccurate gradient estimation in global adaptive optimizer; (ii) client drifts exacerbated by local over-fitting with the local adaptive optimizer. In this work, we propose a novel momentum-based algorithm via utilizing the global gradient descent and locally adaptive amended optimizer to tackle these difficulties. Specifically, we incorporate a locally amended technique to the adaptive optimizer, named Federated Local ADaptive Amended optimizer (\textit{FedLADA}), which estimates the global average offset in the previous communication round and corrects the local offset through a momentum-like term to further improve the empirical training speed and mitigate the heterogeneous over-fitting. Theoretically, we establish the convergence rate of \textit{FedLADA} with a linear speedup property on the non-convex case under the partial participation settings. Moreover, we conduct extensive experiments on the real-world dataset to demonstrate the efficacy of our proposed \textit{FedLADA}, which could greatly reduce the communication rounds and achieves higher accuracy than several baselines.
\end{abstract}
}
% make the title area
\maketitle

% To allow for easy dual compilation without having to reenter the
% abstract/keywords data, the \IEEEtitleabstractindextext text will
% not be used in maketitle, but will appear (i.e., to be "transported")
% here as \IEEEdisplaynontitleabstractindextext when the compsoc 
% or transmag modes are not selected <OR> if conference mode is selected 
% - because all conference papers position the abstract like regular
% papers do.
\IEEEdisplaynontitleabstractindextext
% \IEEEdisplaynontitleabstractindextext has no effect when using
% compsoc or transmag under a non-conference mode.

% For peer review papers, you can put extra information on the cover
% page as needed:
% \ifCLASSOPTIONpeerreview
% \begin{center} \bfseries EDICS Category: 3-BBND \end{center}
% \fi
%
% For peerreview papers, this IEEEtran command inserts a page break and
% creates the second title. It will be ignored for other modes.
\IEEEpeerreviewmaketitle
\IEEEraisesectionheading{\section{Introduction}\label{Introduction}}
\IEEEPARstart{F}{ederated} learning is a distributed machine learning framework for collaboratively training the global model without sharing the private dataset \cite{FL_root}. With the increasing privatization and localization of information, FL has gradually become one of the most practical and efficient methods to implement large-scale distributed training and protect data privacy \cite{FL_intro3,FL_intro2,FL_intro1}. In the federated cross-device settings, a global model is usually trained by hundreds or thousands of clients. The expensive communication costs among the cluster have become the main bottleneck, especially on the global server. To reduce the total amount of communication costs, \cite{FL_root} propose a classic and efficient algorithm FedAvg, via applying local stochastic gradient descent (local SGD) strategy \cite{localsgd} and partial participation to exchange information after several local training iterations among a subset of randomly sampled clients in each communication round. The global server aggregates the local differences and performs standard gradient descent. Referring to this scheme, a series of SGD-based two-stage optimization algorithms \cite{SlowMo, FedDyn, FedADC} are proposed to implement large-scale training in FL settings.

While SGD-based algorithms have seen great success in FL, applying a more efficient adaptive optimizer in FL has become an important topic worthy of our attention. In recent years, the adaptive optimizer has been widely studied as an alternative to vanilla SGD, which can reduce the number of training iterations and achieve better performance on many tasks. Several works focus on embedding the adaptive optimizer into the FL framework and benefit from the adaptivity of the learning rate. \cite{fedadam} incorporate the adaptive optimizer on the global server to update parameters through the gradient estimation aggregated from the multiple local stochastic gradients. Global adaptive optimizer inherits the advantages of general adaptive algorithms and demonstrates stronger generalization ability in FL training. \cite{local_adaAlter} and \cite{FedLamb} respectively perform different local adaptive optimizer in FL frameworks with the adaptive adjustment of layer-wise and dimension-wise in local optimization, which converges faster in  deep neural networks. Local adaptive optimizer improves the training speed in local clients and effectively reduces the number of communication rounds for convergence.

Though adaptive optimizers both on the global server and local clients exhibit amazing performance and excellent potential in practical FL applications, there are still daunting challenges in training practical non-convex deep network scenarios. Our experiments indicate that the global adaptive optimizer has an adverse impact on convergence speed, which is much slower than SGD-based algorithms. Inaccurate gradient estimation from local clients' differences introduces a larger variance in the calculation of second-order momenta and leads to instability of the training process. Local adaptive optimizer that effectively improves the convergence speed suffers from the negative implication of significant over-fitting. The heterogeneous dataset yields huge gaps between aggregated local optimum and global optimum as client drifts mentioned in \cite{SCAFFOLD} by $\mathbf{x}^{*}\neq \frac{1}{m}\sum_{i}\mathbf{x}_{i}^{*}$ where $\mathbf{x}^{*}$ represents for the optimum of the objective function. Therefore, the local heterogeneous dataset causes the unsatisfactory performance of the global model. 
% In our experiments, 
In federated deep model training, we empirically reveal the lower generalization problem caused by local over-fitting of directly applying the local adaptive optimizer compared to SGD-based algorithms.
% lower generalization is observed as directly applying local adaptive optimizer than SGD-based algorithms in federated deep model training.

To tackle the aforementioned heterogeneous over-fitting of applying the local adaptive optimizer, we propose a momentum-based algorithm named FedLADA. We incorporate a novel local amended technique with the local adaptive optimizer which estimates the global average offset in the previous communication round and correct the local offset with a momentum-like term as introduced in \cite{quasi-momentum,FedCM}. In the local training, our method can modify the current direction through the exponential average of past global offsets, which effectively improves the stability of applying the adaptive optimizer on local clients. On the global server, it performs one-step gradient descent as the aggregation of local training. This technique forces the local offset direction toward to global optimum and effectively alleviates the influence of local over-fitting. Compared to the existing works, the local amended technique considers the average offset as the global direction instead of the average gradient, which is more suitable for the adaptive optimizer in FL training. Theoretically, we provide the proof of our proposed algorithm with a linear speedup property on the heterogeneous non-convex and $L$-smooth finite-sum objective functions under the partial participating setting. We compare the performance of applying local amended techniques in the global and local adaptive optimizer, respectively. Extensive experiments on CIFAR-10/100 and TinyImagenet show that our proposed FedLADA method achieves faster convergence speed and higher generalization accuracy in training deep neural networks than several baselines.

To the end, we summarize our main contributions as:
\begin{itemize}
    \item We explore the major challenges of applying adaptive methods in FL framework and summarize them as \textit{rugged convergence} (induced by applying global adaptive optimizer) and \textit{client drifts} (exacerbated by local over-fitting of applying local adaptive optimizer).
    \item We proposed a novel and communication-efficient FedLADA algorithm in FL, which incorporates the local adaptive amended technique on the local adaptive optimizer to alleviate the negative impact of client drifts caused by over-fitting on the heterogeneous datasets and maintain high convergence speed.
    \item We analyze the convergence rate of proposed FedLADA for non-convex and $L$-smooth objective functions under partial participating setting, which achieves a linear speedup of $\mathbf{O}(\frac{1}{\sqrt{SKT}})$.
    %\item We speculate through comparative experiments that global adaptive optimizer schemes, e.g. for FedAdam, are difficult to be further improved by local amended techniques.
    \item We conduct extensive experiments on CIFAR-10/100 and TinyImagenet datasets to verify the effectiveness of FedLADA, which achieves faster convergence speed in deep network training (approximately 1.2$\times$ than vanilla local adaptive optimizer and 1.5$\times$ than the best SGD-based baseline) and higher test accuracy.
\end{itemize}

\section{Related Work}
\label{related_work}
\textbf{Federated Learning.} Since \cite{FL_root} firstly propose the FL framework and FedAvg algorithm to address the key challenges of communication bottleneck and heterogeneous dataset with the theoretical analysis \cite{favg_convergence1,favg_convergence2,favg_convergence3} of linear speedup property, a series of SGD-based methods are proposed to implement large-scale training in FL. \cite{FedProx} introduce the FedProx algorithm to further tackle heterogeneity in FL through a proximal term to limit the difference of local training. However, it uses a proxy term to force the local update toward the last global state, which will perform worse under the limited local interval $K$. Motivated by the significant effects of variance reduction techniques in stochastic optimization, \cite{SCAFFOLD} propose SCAFFOLD which applies SVRG \cite{SVRG} to alleviate the client drifts. It introduces an additional term to correct the local SGD optimizer toward a global estimation. FedNova \cite{FedNova} considers different local steps on asynchronous aggregation settings and averages the normalized local offset. It is a variant of FedAvg with allowing 
asynchronous updates. \cite{FedPD} first incorporates Primal-Dual on the local clients to be adaptive to different levels of local heterogeneity. It adopts the averaged local dual variable as the correction which could effectively improve the local consistency. However, it relies on full participation. To allow partial participation, \cite{FedDyn} focuses on consistency and proposes FedDyn which forces the local objective optimum close to the global optimum. It employs the mean of the dual variable of all clients and the average of the selected local clients at communication round $t$ to update the global state. \cite{local_momentum} analyze the efficiency of parallel local restarted momentum. \cite{SlowMo} introduce SlowMo to apply the global momentum update to yield improvements in optimization and test accuracy. \cite{FedCM, FedADC} apply average global gradient as a client-level momentum term to achieve better performance. They both successfully explore the potential of the momentum terms in FL. Inspired by the advantages of 1) effective global correction could reduce the biases during the updates; and 2) enhancing local consistency could improve the global training efficiency, we study the efficient local adaptive optimizer in FL with amended techniques.

The two major challenges that FL faces are the huge costs of information exchange and communication in massively parallel computing and the huge bias of training on the local heterogeneous dataset without directly sharing data. Though the current large-scale federated architectures are based on distributed frameworks, efficient information transmission bandwidth creates barriers to communication costs. At the same time, due to the effect of heterogeneity, FL has to solve the trade-off of balancing global model precision and local model precision. \cite{L2GD} introduce the impact of measuring local and global optimums in FL. Applying local intervals and partial participation settings can effectively reduce communication costs while it hurts the convergence speed and precision severely. \cite{t1,t2,t4,GD} prove the excellent properties of local GD and SGD. \cite{appendix_1,appendix_2} provide the differences between applying local intervals and larger mini-batches on convergence. \cite{yurochkin2019bayesian} analyze the Bayesian nonparametric models in FL. \cite{appendix_3} proves the convergence of local SGD for random intervals and gives the upper bound on the theoretical convergence of linearly increasing intervals. Based on local theoretical guarantees, more efficient algorithms are tapped into the FL framework \cite{FedSplit,FedDR,FedMAGA}. \cite{FedPuring,FedHN} investigate the solutions to non-aggregated global models. With the wide applications of first-order gradient descent-based methods, recently how to apply more efficient optimizers has become a rookie. \cite{t3,appendix_4} focus more on federated adaptive gradients. \cite{distributed_adam} apply adam optimizer to distributed systems. \cite{decentralized_adaptive,local_adaAlter,local_adaptive} apply adaptive optimizer on local clients to improve performance. \cite{adam_rmsprop} prove convergence conditions in adaptive optimization and \cite{quantized_adam,adam_convergence_convexity} apply error-feedback and quantization to adam to further improve communication efficiency. \cite{yu2021fed2} adopt an aligned module to improve the training efficiency. \cite{li2022federated} propose the position-aware neurons in FL to avoid the negative impacts.  \cite{FedNL,hessian1} explore second-order methods in federated learning for more efficient training.

\textbf{Adaptive Optimization.} Adaptive methods in FL greatly benefit from the adaptivity on the heterogeneous dataset. \cite{adagrad,adapt1,adam,adapt2,adapt3} study several adaptive methods on non-FL settings. A lot of powerful variants are proposed including Adagrad \cite{adagrad}, Adadelta \cite{adadelta}, Adam \cite{adam}, Amsgrad \cite{amsgrad} and Nadam, etc. To be adapted to different tasks. adaptive methods have achieved more excellent empirical performance than SGD. \cite{fedadam} incorporate adaptive optimizer on the global server in FL framework to accelerate the convergence speed in deep network training. \cite{local_adaAlter} apply the AdaAlter optimizer on the local clients with the lazily updated denominators. \cite{average_v} indicate the second-order momenta of local Amsgrad must be averaged to avoid divergence in the training process. \cite{local_preconditioner} prove the inconsistency leads to  non-vanishing gaps in a toy quadratic example and update the global model by averaging the inverse of the local pre-conditioner matrices. Compared with these works, our proposed method benefits from the fast convergence speed of the local adaptive optimizer and takes advantage of the locally amended technique to mitigate over-fitting on non-$iid$ dataset.
\section{Methodology}
\label{method}

\begin{algorithm}[tb]  
  \renewcommand{\algorithmicrequire}{\textbf{Input:}}
  \renewcommand{\algorithmicensure}{\textbf{Output:}}
  \caption{Global Federated Adaptive}  
  \begin{algorithmic}[1]\label{global}
  \REQUIRE $\mathbf{x}_{0}$, $\beta_{1}$, $\beta_{2}$, mapping function $h_{t}$
  \ENSURE global parameters $\mathbf{x}^{T}$
    \FOR{$t = 0, 1, 2, \cdots, T-1$}
    \STATE sample subset $\mathcal{S}^{t}$ from $[m]$ 
      \FOR{client $i \in \mathcal{S}^{t}$ parallel}
        \STATE communicate $\mathbf{x}^{t}$ as $\mathbf{x}_{i,0}^{t}$
        \FOR{$\tau= 0, 1, 2, \cdots, K-1$}
          \STATE compute stochastic gradient $\mathbf{g}_{i,\tau}^{t}$
          \STATE $\mathbf{x}_{i,\tau+1}^{t}=\mathbf{x}_{i,\tau}^{t}-\eta_{l}\mathbf{g}_{i,\tau}^{t}$
        \ENDFOR
        \STATE communicate $\hat{\mathbf{g}}_{i}^{t}=\mathbf{x}_{i,0}^{t}-\mathbf{x}_{i,K}^{t}$
      \ENDFOR
      \STATE $\hat{\mathbf{g}}_{t}=\textbf{Average}_{i}(\hat{\mathbf{g}}_{i}^{t})$
      \STATE $\mathbf{m}_{t+1}=(1-\beta_1)\mathbf{m}_{t}+\beta_1\hat{\mathbf{g}}_{t}$
      \STATE $\mathbf{v}_{t+1}=(1-\beta_2)\mathbf{v}_{t}+\beta_2\hat{\mathbf{g}}_{t}\odot\hat{\mathbf{g}}_{t}$
      \STATE $\hat{\mathbf{v}}_{t+1}=h_{t}(\{\mathbf{v}_{t}\})$
      \STATE $\mathbf{x}_{t+1}=\mathbf{x}_{t}-\eta_{g}\mathbf{m}_{t+1}/(\sqrt{\hat{\mathbf{v}}_{t+1}} )$
    \ENDFOR
    \STATE \textbf{Output:} $\mathbf{x}_{T}$
  \end{algorithmic}  
\end{algorithm}  

In this section, we introduce some preliminaries and our proposed FedLADA. We will explain the algorithm flow and the implicit meaning of the local amended technique. Firstly we define some notations for convenience.
\subsection{Preliminary and Notations}
\textbf{Problem setup.} We consider a finite-sum non-convex optimization problem $F$ : $\mathbb{R}^{d}\rightarrow\mathbb{R}$ of the form:
\begin{equation}
\label{core-problem}
    F\left( \mathbf{x} \right) = \frac{1}{m} \sum_{i=1}^{m} F_{i} \left( \mathbf{x} \right), \quad  \mathbf{x}^{*} := \mathop{\arg\min}_{\mathbf{x} \in \mathbb{R}^{d}} F(\mathbf{x}),
\end{equation}
where $F_{i}(\mathbf{x})=\mathbb{E}_{\varepsilon_{i}\sim\mathcal{D}_{i}}F_{i}(\mathbf{x},\varepsilon_{i})$ is the local objective function at $i$-th client and $\varepsilon_{i}$ denotes the randomly sampled dataset obeying distribution $\mathcal{D}_{i}$. $m$ is the total number of local clients. In FL problems, $\mathcal{D}_{i}$ may differ across local clients.\\

\textbf{Notations.} $m, S$ are the number of total clients and active clients per round, respectively. We denote $\mathcal{S}^{t}$ as the set of active clients randomly sampled from the whole clients. $K$ is the number of local iterations and $T$ is the total communication rounds. $(\cdot)_{i,\tau}^{t}$ denotes the $i$-th client's variable $(\cdot)$ at $\tau$-th iteration in $t$-th round. $\mathbf{x}\in\mathbb{R}^{d}$ is the model parameters. $\mathbf{g}_{i,\tau}^{t}=\nabla F_{i}(\mathbf{x}_{i,\tau}^{t},\varepsilon_{i,\tau}^{t})$ is the stochastic gradient computed by the sampled data $\varepsilon_{i,\tau}^{t}$ ($\varepsilon$ will be omitted if it is fully sampled). $\mathbf{m}$ is the momentum term (exponential average of past stochastic gradients). $\mathbf{v}$ is the second-order momentum term (exponential average of past square stochastic gradients). $\hat{\mathbf{v}}$ is the element-wise historical maximum value of $\mathbf{v}$. We denote $\langle\cdot,\cdot\rangle$ as the inner product and $\odot$ as the Hadamard product between vectors. $\Vert \cdot \Vert$ is the Euclidean norm. $\{\cdot\}$ means a set of variables. If there are no specific instructions, addition, subtraction, division, power, and square root between vectors are element-wised.

\begin{algorithm}[tb]
  \renewcommand{\algorithmicrequire}{\textbf{Input:}}
  \renewcommand{\algorithmicensure}{\textbf{Output:}}
  \caption{Local Federated Adaptive}  
  \begin{algorithmic}[1]\label{local}
  \REQUIRE $\mathbf{x}_{0}$, $\beta_{1}$, $\beta_{2}$, mapping function $h_{t}$
  \ENSURE global parameters $\mathbf{x}^{T}$
    \FOR{$t = 0, 1, 2, \cdots, T-1$}  
      \STATE sample subset $\mathcal{S}^{t}$ from $[m]$ 
      \FOR{client $i \in \mathcal{S}^{t}$ parallel}
        \STATE communicate $\mathbf{x}_{t}$ as $\mathbf{x}_{i,0}^{t}$
        \FOR{$\tau= 0, 1, 2, \cdots, K-1$}
          \STATE compute stochastic gradient $\mathbf{g}_{i,\tau}^{t}$
          \STATE $\mathbf{m}_{i,\tau+1}^{t}=(1-\beta_{1})\mathbf{m}_{i,\tau}^{t}+\beta_{1}\mathbf{g}_{i,\tau}^{t}$
          \STATE $\mathbf{v}_{i,\tau+1}^{t}=(1-\beta_2)\mathbf{v}_{i,\tau}^{t}+\beta_2\mathbf{g}_{i,\tau}^{t}\odot\mathbf{g}_{i,\tau}^{t}$
          \STATE $\hat{\mathbf{v}}_{i,\tau+1}^{t}=h_{t}(\{\mathbf{v}_{i,\tau}^{t}\})$
          \STATE $\mathbf{x}_{i,\tau+1}^{t}=\mathbf{x}_{i,\tau}^{t}-\eta_{l}\mathbf{m}_{i,\tau+1}^{t}/(\sqrt{\hat{\mathbf{v}}_{i,\tau+1}^{t}} )$
        \ENDFOR
        \STATE communicate $\hat{\mathbf{g}}_{i}^{t}=\mathbf{x}_{i,0}^{t}-\mathbf{x}_{i,K}^{t}$
      \ENDFOR
      \STATE $\hat{\mathbf{g}}_{t}=\textbf{Average}_{i}(\hat{\mathbf{g}}_{i}^{t})$
      \STATE $\mathbf{x}_{t+1}=\mathbf{x}_{t}-\eta_{g}\hat{\mathbf{g}}_{t}$
    \ENDFOR
    \STATE \textbf{Output:} $\mathbf{x}_{T}$
  \end{algorithmic}  
\end{algorithm}  

\subsection{Federated Adaptive Optimizer}
We introduce the adaptive update as follows:
\begin{equation}
\textbf{Adaptive:}\left\{
\begin{aligned}
\mathbf{m}_{t+1} & =  (1-\beta_{t})\mathbf{m}_{t}+\beta_{t}\mathbf{g}_{t} \\
\mathbf{v}_{t+1} & = (1-\beta_2)\mathbf{v}_{t}+\beta_2\mathbf{g}_{t}\odot\mathbf{g}_{t} \\
\hat{\mathbf{v}}_{t+1} & =  h_{t}(\{\textbf{v}_{t}\}) \\
\mathbf{x}_{t+1} & =  \mathbf{x}_{t} - \eta_{t}\mathbf{m}_{t+1}/\left(\sqrt{\hat{\mathbf{v}}_{t+1}}\right)
\end{aligned}
\right.
\end{equation}
where $h_{t}(\cdot)$ denotes the different mapping function.\\

\textbf{Global Federated Adaptive.} In the federated learning framework, FedAdam \cite{fedadam} firstly applies the global adaptive optimizer with local SGD to update the model parameters. This idea stems from using the average of local gradients, which has been proven as the robust alternative of the true global gradient in SGD-based federated optimization. It adopts the local-SGD optimization for several steps and then communicates to the global server. The server aggregates the local offset as the vanilla gradient term in the adaptive optimizer, which could be considered as a quasi-gradient on the global server. Details can be referred to  Algorithm~\ref{global}. Global adaptive optimizer benefits from the element-wise change of the inaccurate gradient estimation revised by the second-order momenta term.

\begin{figure}[h]
    \setlength{\abovecaptionskip}{0.cm}
	\begin{minipage}[t]{0.495\linewidth}
		\centering
		\includegraphics[width=0.99\linewidth]{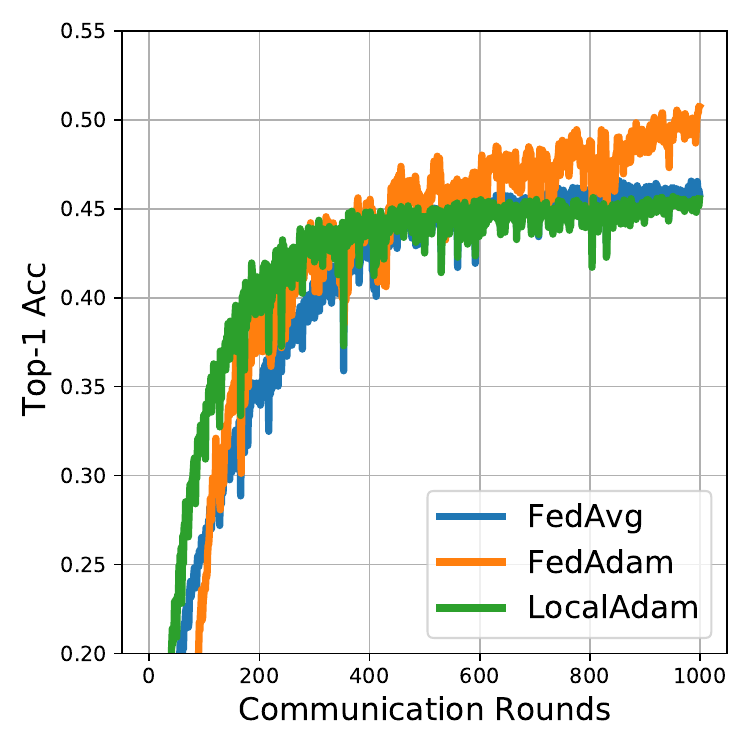}
	\end{minipage}
	\begin{minipage}[t]{0.495\linewidth}
		\centering
		\includegraphics[width=0.99\linewidth]{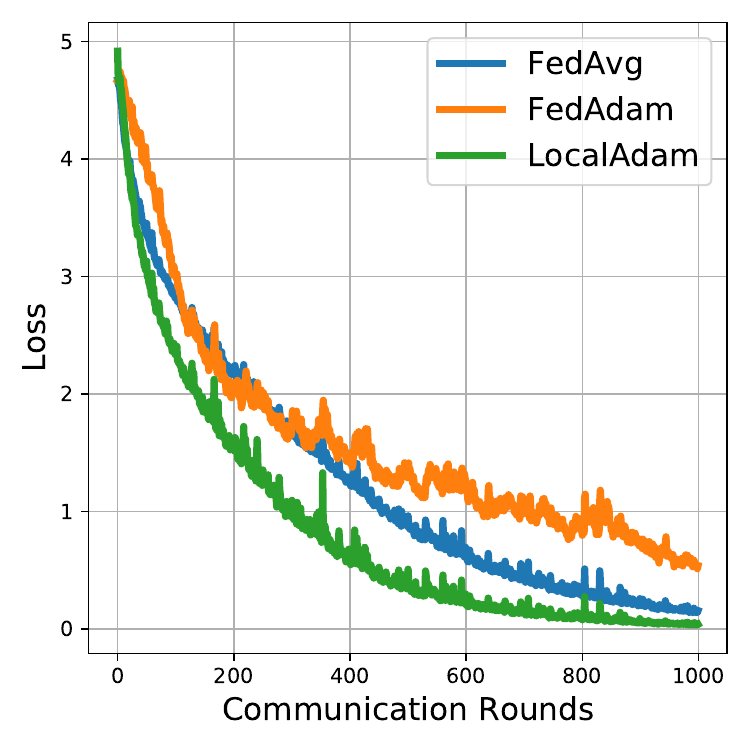}
	\end{minipage}
	%\vskip -0.1in
	\caption{Test accuracy (left) and training loss (right) of FedAvg, FedAdam and LocalAdam on CIFAR100. LocalAdam converges the fastest than FedAvg under the same setups. FedAdam can prevent the local overfitting to some extent, while from the perspective of optimization, its efficiency is extremely low.}
	\label{compare_global_local}
	%\vskip -0.1in
\end{figure}

\textbf{Local Federated Adaptive.} As shown in Algorithm~\ref{local}, local adaptive schemes apply aggregation (or gradient descent) on the global server and local adaptive optimizer to update the parameters instead. Each local training process solves the sub-problem as SGD-based methods. Federated local adaptive schemes show the same advantages as training in a single node. FedAdam can only complete one adaptive adjustment after local $K$ iterations. The error of the local training process is gradually accumulated due to the inaccurate gradient estimation. As updated on the global server, the accumulated error has an adverse impact on the adaptive correction of the gradient especially when $K$ is large enough. Though we focus more on the number of global communication rounds and ignore the local training costs in FL framework, the limitation of $K$ affects the training convergence rate and the total number of communication rounds $T$ to a certain extent. Local adaptive is more like an upgraded version of FedAvg which replaces local SGD with an adaptive optimizer. LocalAdam converges faster than FedAvg while it performs worse in generalization. Local heterogeneity is the challenging difficulty in FL framework which forces each local model parameter to be closer to the local optimum. Therefore, without local correction, the higher efficiency of local optimizer intensifies local over-fitting and exacerbates serious client drifts, as shown in Figure~\ref{compare_global_local}.

\subsection{FedLADA Algorithm}
In this part, we will introduce our proposed method and explain how it lessens the negative impact of heterogeneity and reduces the communication rounds. The core inspiration behind our proposed method is to apply the global change as a local gradient correction into the adaptive optimizer to control the client drifts. We consider that a global server receives and aggregates the local change $(\mathbf{x}_{i,0}^{t} - \mathbf{x}_{i,K}^{t})$ of client $i \in \mathcal{S}^{t}$ at the end of each training round $t$, and then sends the global parameters $\mathbf{x}^{t+1}$ to each client $i \in \mathcal{S}^{t+1}$ at the beginning of each round $t+1$ for training. The local active clients start their updating in parallel.

\begin{algorithm}[tb]
	\renewcommand{\algorithmicrequire}{\textbf{Input:}}
	\renewcommand{\algorithmicensure}{\textbf{Output:}}
	\caption{FedLADA}
	\begin{algorithmic}[1]
		\REQUIRE Initial parameters $\mathbf{x}^{0}$, local learning rate $\eta_{l}$, global learning rate $\eta_{g}$, communication rounds $T$, local iterations $K$, $\mathbf{g}_{a}^{0}=0$, $\hat{\mathbf{v}}^{0}=\epsilon_{v}^{2}$, amended weight $\alpha$, first-order momenta weight $\beta_{1}$, second-order momenta weight $\beta_{2}$
		\ENSURE global parameters $\mathbf{x}^{T}$
		\FOR{$t = 0, 1, 2, \cdots, T-1$}
		% \STATE $\mathbf{x}_{t+1} = \mathop{\arg\min} \mathbf{u}(\mathbf{x},
		% \mathbf{x}_{t}) $
		\STATE randomly select active clients-set $\mathcal{S}^{t}$ at round $t$
		\FOR {client $i \in \mathcal{S}^{t}$ parallel}
		\STATE transfer $\mathbf{x}^{t}$ and $\mathbf{v}^{t}$ to client $i$ and set
		\STATE $\mathbf{x}_{i,0}^{t}=\mathbf{x}^{t}, \mathbf{m}_{i,0}^{t}=\mathbf{0}, \hat{\mathbf{v}}_{i,0}^{t}=\mathbf{v}^{t}$
		\FOR{$\tau = 1, 2, \cdots, K-1$}
		\STATE compute unbiased stochastic gradient $\mathbf{g}_{i,\tau}^{t}$
		\STATE $\mathbf{m}_{i,\tau}^{t} = \beta_{1} \mathbf{m}_{i,\tau-1}^{t} + (1 - \beta_{1}) \mathbf{g}_{i,\tau}^{t}$
		\STATE $\mathbf{v}_{i,\tau}^{t} = \beta_{2} \mathbf{v}_{i,\tau-1}^{t} + (1 - \beta_{2}) \mathbf{g}_{i,\tau}^{t}\odot\mathbf{g}_{i,\tau}^{t}$
		\STATE $\hat{\mathbf{v}}_{i,\tau}^{t} = \max(\mathbf{v}_{i,\tau}^{t}, \hat{\mathbf{v}}_{i,\tau-1}^{t})$
		\STATE $\vartheta_{i,\tau}^{t} = 1/\sqrt{\hat{\mathbf{v}}_{i,\tau}^{t}}$
		\STATE $\mathbf{x}_{i,\tau}^{t} = \mathbf{x}_{i,\tau-1}^{t} - \eta_{l}\big(\alpha\mathbf{m}_{i,\tau}^{t}\odot\vartheta_{i,\tau}^{t} + (1-\alpha)\mathbf{g}_{a}^{t}\big)$
		\ENDFOR
		\STATE communicate $\mathbf{x}_{i,0}^{t} - \mathbf{x}_{i,K}^{t}$ and $\hat{\mathbf{v}}_{i,K}^{t}$ to the server
		\ENDFOR
		\STATE $\mathbf{v}^{t+1} = \frac{1}{S}\sum_{i \in \mathcal{S}^{t}}\hat{\mathbf{v}}_{i,K}^{t}$
		\STATE $\mathbf{x}^{t+1} = \mathbf{x}^{t} - \eta_{g}\frac{1}{S}\sum_{i\in\mathcal{S}^{t}}(\mathbf{x}_{i,0}^{t} - \mathbf{x}_{i,K}^{t})$
		\STATE $\mathbf{g}_{a}^{t+1} = \frac{1}{\eta_{g}\eta_{l}K}(\mathbf{x}^{t} - \mathbf{x}^{t+1})$
		\ENDFOR
	\end{algorithmic}
	\label{algorithm_fedLada}
\end{algorithm}

In Algorithm~\ref{algorithm_fedLada}, the global parameters $\mathbf{x}^{0}$ is randomly generated. We set the initial momentum term $\mathbf{m}_{i,0}^{t}=\mathbf{0}$ in each round $t$ to eliminate local historical gradient information. Each $\hat{\mathbf{v}}_{i,K}^{t}$ is averaged on the global server to guarantee the convergence \cite{average_v} and $\hat{\mathbf{v}}^{0}$ is set as $\epsilon_{v}^{2}$. To further reduce the adverse effect of local heterogeneity, we introduce a momentum-like term $\mathbf{g}_{a}^{t}$, the average of the global change in round $t-1$, to correct the local update direction. More details can be referred to the algorithm paradigm.

The same as FedAvg, $\mathbf{g}_{a}^{t}$ averages each local change of active clients and serves as the direction for the global update at round $t$:
\begin{equation}
    \mathbf{g}_{a}^{t} = \frac{1}{\eta_{g}\eta_{l}K}(\mathbf{x}^{t}-\mathbf{x}^{t+1}).
\end{equation}
\cite{FedCM,quasi-momentum} indicate the potential ability of averaged gradient-level momentum-like terms to suppress local heterogeneity both in centralized and decentralized FL systems. In a local adaptive optimizer, due to the precondition vector or matrix, it is difficult to recover gradient information during optimization. A natural improvement is to apply the $\mathbf{g}_{a}$ to guide the local update directly. We have:
\begin{equation}
    \mathbf{g}_{a}^{t} = \alpha\frac{1}{SK}\sum_{i,\tau}\mathbf{m}_{i,\tau}^{t}\odot\vartheta_{i,\tau}^{t}+(1-\alpha)\mathbf{g}_{a}^{t-1}.
\end{equation}
Here, $\mathbf{g}_{a}$ is an exponential average of each previous local update direction. It should be noted that $\mathbf{g}_{a}$ is different from the global momentum mentioned in \cite{SlowMo} which aggregates the average local change on the global optimizer and aim to improve the generalization performance. The local training process can not directly benefit from the global momentum, which provides an initial point with historical gradient information for clients. Compared to local momentum, $\mathbf{g}_{a}$ retains information of other clients instead of unilateral changes. $\mathbf{g}_{a}$ mimics a more precise approximation of the global optimization directions instead of gradients, which avoids the impact of second-order momenta in the local training. Unlike vanilla SGD, an adaptive optimizer applies element-wised adjustment on the gradients for descent. The direct use of gradient correction cannot effectively solve heterogeneous client drifts, and it introduces turmoil on the convergence, as shown in section~\ref{experiments}. We need to emphasize that the global offset estimation $\mathbf{g}_{a}$ in algorithm~\ref{algorithm_fedLada} is the accumulation of adaptive gradients, which avoids calculating the weighted average gradient from the momentum term.

\subsection{Convergence Analysis}
\label{convergence_analysis}
In this part, we give the theoretical analysis of our proposed FedLada algorithm. Firstly we state some standard assumptions for the non-convex function $F$.
\begin{assumption}
\label{smoothness}
(Smoothness) \textit{The non-convex $F_{i}$ is a $L$-smooth function for all $i\in[m]$, i.e., $\Vert\nabla F_{i}(\mathbf{x})-\nabla F_{i}(\mathbf{y})\Vert\leq L\Vert\mathbf{x}-\mathbf{y}\Vert$, for all $\mathbf{x},\mathbf{y}\in\mathbb{R}^{d}$.}
\end{assumption}
\begin{assumption}
\label{bounded_stochastic_gradient_I}
(Bounded Stochastic Gradient 1) \textit{$\mathbf{g}_{i}^{t}=\nabla F_{i}(\mathbf{x}_{i}^{t}, \varepsilon_{i}^{t})$ computed by using a sampled mini-batch data $\varepsilon_{i}^{t}$ in the local client $i$ is an unbiased estimator of $\nabla F_{i}$ with bounded variance, i.e., $\mathbb{E}_{\varepsilon_{i}^{t}}[\mathbf{g}_{i}^{t}]=\nabla F_{i}(\mathbf{x}_{i}^{t})$ and $\mathbb{E}_{\varepsilon_{i}^{t}}\Vert g_{i}^{t} - \nabla F_{i}(\mathbf{x}_{i}^{t})\Vert^{2} \leq \sigma_{l}^{2}$, for all $\mathbf{x}_{i}^{t}\in\mathbb{R}^{d}$.}
\end{assumption}
\begin{assumption}
\label{bounded_stochastic_gradient_II}
(Bounded Stochastic Gradient 2) \textit{Each element of stochastic gradient $\mathbf{g}_{i}^{t}$ is bounded, i.e., $\Vert\mathbf{g}_{i}^{t}\Vert_{\infty}=\Vert F_{i}(\mathbf{x}_{i}^{t},\varepsilon_{i}^{t})\Vert_{\infty}\leq G_{g}$, for all $\mathbf{x}_{i}^{t}\in\mathbb{R}^{d}$ and any sampled mini-batch data $\varepsilon_{i}^{t}$.}
\end{assumption}
\begin{assumption}
\label{bounded_heterogeneity}
(Bounded Heterogeneity) \textit{The dissimilarity between local clients is bounded on the gradients, i.e., $\Vert\nabla F_{i}(\mathbf{x})-\nabla F(\mathbf{x})\Vert^{2}\leq\sigma_{g}^{2}$, for all $\mathbf{x}\in\mathbb{R}^{d}$.}
\end{assumption}

According to Assumption \ref{bounded_stochastic_gradient_II}, the exponential average gradient $\Vert \mathbf{m}_{i,k}^{t}\Vert_{\infty}=\Vert(1-\beta_{1})\sum_{\tau=1}^{k}\beta_{1}^{k-\tau}\mathbf{g}_{i,\tau}^{t}\Vert_{\infty}\leq G_{g}$ holds. Noting that in Assumption \ref{bounded_stochastic_gradient_II} the local gradient is element-wised bounded, thus $\nabla F$ satisfies $\Vert \nabla F\Vert_{\infty}\leq G_{g}$ as a constant upper bound. We use $\Vert\nabla F\Vert^{2}\leq\sigma_{u}^{2}\leq dG_{g}^{2}$ instead ($\Vert \nabla F\Vert^{2}=\sum_{j=1}^{d}(\nabla F)_{j}\leq\sum_{j=1}^{d}\Vert \nabla F\Vert_{\infty}$). About the heterogeneity Assumption \ref{bounded_heterogeneity}, \cite{SCAFFOLD} apply two other assumptions, $(G,B)$-BGD and $\delta$-BHD. The dominant term of the convergence rate is generally the same under these three assumptions. We use the common gradient bound condition in our proof. According to the algorithm\ref{algorithm_fedLada}, $\hat{\mathbf{v}}$ is updated as:
\begin{equation}
    \hat{\mathbf{v}}_{i,\tau}^{t}=\max(\mathbf{v}_{i,\tau}^{t}, \hat{\mathbf{v}}_{i,\tau-1}^{t}).
\end{equation}
It can be seen that $\hat{\mathbf{v}}$ is a non-decreasing sequence. Then we can upper bound $G_{\vartheta}$ as:
\begin{align*}
    G_{\vartheta} \
    &= \max\left\{\Vert\vartheta_{i,\tau}^{t}\Vert_{\infty}\right\}=\max\left\{\Vert\frac{1}{\sqrt{\hat{\mathbf{v}}}_{i,\tau}^{t}}\Vert_{\infty}\right\}\\
    &=\max\left\{\Vert\frac{1}{\max(\sqrt{(1-\beta_{2}^{K})(\mathbf{g}_{i,\tau}^{t})^{2}},\epsilon_{v})}\Vert_{\infty}\right\}.
\end{align*}
For $\sqrt{(1-\beta_{2}^{K})(\mathbf{g}_{i,\tau}^{t})^{2}}$ is bounded as:
\begin{equation}
     \Vert\sqrt{(1-\beta_{2}^{K})(\mathbf{g}_{i,\tau}^{t})^{2}}\Vert_{\infty}\leq \Vert\mathbf{g}_{i,\tau}^{t}\Vert_{\infty}.
\end{equation}
Thus we bound $G_{\vartheta}$ as $\frac{1}{G_{g}}\leq G_{\vartheta} \leq\frac{1}{\epsilon_{v}}$.

We consider the partial participating settings.
\begin{theorem}
When the learning rate $\eta_{l}$ and $\eta_{g}$ satisfy that $\eta\eta_{l} \leq \min\{\frac{1}{4L},  \frac{\beta_{1}S(m-1)(1+\mu)}{32(m-S)LG_{\vartheta}}\}$ and $\eta_{l} \leq \frac{1}{2\sqrt{2\alpha(1-\beta_{1})K}G_{\vartheta}L}$ where $G_{\vartheta}$ is a constant bounded above. Let the Assumptions above hold and let partial participating ratio equal to $\frac{S}{m}$ to randomly sample an active client-set, the sequence $\{\mathbf{z}^{t} = \frac{1}{\alpha}\mathbf{x}^{t} - \frac{1-\alpha}{\alpha}\mathbf{x}^{t-1}\}$, where $\mathbf{x}^{t}$ is the global model parameters generated by Algorithm\ref{algorithm_fedLada} at round $t$, satisfies:
\begin{equation}
    \frac{1}{T}\sum_{t=0}^{T-1}\mathbb{E}\Vert\nabla F(\mathbf{z}^{t})\Vert^{2}\leq \frac{F(\mathbf{z}^{0})-f_{*}}{\eta\eta_{l}CG_{\vartheta}T} + \Psi,
\end{equation}
where $f_{*}$ is the optimum of function F and $C=\frac{1}{2}\left(1-\frac{\beta_1}{(1-\beta_1)K}\right)$ is a constant. The $\Psi$ is:
\begin{align*}
    \Psi
    &= \eta\eta_{l}\Big(\frac{2\beta_{1}^{2}G_{\vartheta}L\sigma_{l}^{2}}{CSK}+\frac{dG_{g}^{4}   L}{\epsilon_{v}^{4}CG_{\vartheta}T}\Big)+\frac{dG_{g}^{4}}{2\epsilon_{v}^{3}CG_{\vartheta}T} \\
    &+\eta_{l}^{2}\Big(\frac{L^{2}C_{2}G_{c}}{C} + \frac{L^{2}C_{3}\sigma^{2}}{C}\Big) + \eta^{2}\eta_{l}^{2}\frac{(1-\alpha)^{2}L^{2}G_{c}}{\alpha^{2}C},
\end{align*}
where $C_{1} = \frac{\alpha(1+a)(1+\frac{1}{a})}{1+a-\beta_{1}}$, $C_{2} = 6(1-\alpha)(1+\frac{1}{a})KG_{\vartheta}^{2}$ and $C_{3} = 24(1-\beta_{1})KG_{\vartheta}^{2}C_{1}$. $G_{c}=\frac{2G_{g}^{4}}{\epsilon_{v}^{4}} + \frac{2\beta_{1}^{2}G_{\vartheta}^{2}\sigma_{l}^{2}}{SK} + 4G_{\vartheta}^{2}(\sigma_{g}^{2}+\sigma_{u}^{2})$ is the constant combination of the variance and upper bound of gradients, and $\sigma^{2}$ is defined as $\sigma^{2} \!=\!\sigma_{l}^{2}\!+\!\sigma_{g}^{2}\!+\!\sigma_{u}^{2}$.\\
\label{convergence_rate}
\end{theorem}

\begin{corollary}
    When $\eta\eta_{l}$ satisfies the conditions in Theorem \ref{convergence_rate}, let $\eta\eta_{l}=\mathbf{O}(\sqrt{\frac{SK}{T}})$ and $\eta_{l}=\mathbf{O}(\sqrt{\frac{1}{KT}})$ for $\eta_{l} \leq \frac{1}{2\sqrt{2\alpha(1-\beta_{1})K}G_{\vartheta}L}$, the convergence rate of the sequence $\{\mathbf{z}^{t}\}$ generated in Algorithm\ref{algorithm_fedLada} under the partial participating is:
    \begin{equation*}
        \frac{1}{T}\sum_{t=0}^{T-1}\mathbb{E}\Vert\nabla F(\mathbf{z}^{t})\Vert^{2}=\mathbf{O}\left(\frac{1}{\sqrt{SKT}}+\frac{1}{KT}+\frac{1}{T}+\frac{1}{T^{3/2}}\right).
    \end{equation*}
\end{corollary}

\begin{remark}
    The theoretical convergence rate upper bound of Theorem~\ref{convergence_rate} includes a vanishing part as $T$ increases and the constant part of the variance caused by the stochastic method and heterogeneity dataset.\\
\end{remark}
\begin{remark}
\label{111}
    When the communication rounds $T$ is large enough, the dominant term of the convergence rate achieves a linear speedup of $\mathbf{O}(\frac{1}{\sqrt{SKT}})$. Which means to achieves the precision of $\epsilon$, at least $\mathbf{O}(\frac{1}{SK\epsilon^{2}})$ communication rounds are required on non-convex $L$-smooth objective functions.\\
\end{remark}
\begin{remark}
\label{alpha}
    Amended weight $\alpha$ balance the impact of $\sigma^{2}$ and $G_{c}$, the last three terms in Theorem \ref{convergence_rate}, which contains the $\alpha$, $\alpha(1-\alpha)$ and $\big(\frac{1-\alpha}{\alpha}\big)^{2}$ as coefficients. $\exists \alpha \in (0, 1)$ which can minimize the sum of the last three terms. In our experiments, we extensively search many $\alpha$ to verify this.
\end{remark}
\section{Experiments}
\label{experiments}
In this section, we present some empirical challenges of applying adaptive optimizers in FL framework. We demonstrate that the proposed FedLADA algorithm can outperform the SGD-based and vanilla local adaptive baselines empirically with partial participation and heterogeneous dataset.
\begin{table*}[t]
\caption{ResNet-18 convergence speed of different methods on CIFAR10/100 and TinyImageNet. ``Acc.'' represents the target accuracy on the dataset of the corresponding mode. ``Rounds'' is the minimum number of communication rounds required to achieve the target accuracy.}
\label{speed}
%\vskip 0.05in
\begin{center}
\begin{small}
\begin{sc}
\begin{tabular}{llcccccc}
\toprule
 & &\multicolumn{2}{c}{CIFAR10}&\multicolumn{2}{c}{CIFAR100}&\multicolumn{2}{c}{TinyImageNet}\\
\midrule
Mode & Methods & Acc.($\%$) & Rounds & Acc.($\%$) & Rounds & Acc.($\%$) & Rounds\\
\midrule
 & FedAvg     &  & 977 (1.70$\times$) &  & 1198 (1.94$\times$) &  & 1474 (1.67$\times$)\\
 & FedProx    &  & 913 (1.59$\times$) &  & 1166 (1.89$\times$) &  & 1508 (1.71$\times$)\\
 & SCAFFOLD   &  & 740 (1.29$\times$) &  & 1017 (1.65$\times$) &  & 1338 (1.52$\times$)\\
Train & FedCM & 98.0 & 1429 (2.48$\times$) & 98.0 & 966 (1.56$\times$) & 98.0 & 1754 (1.99$\times$)\\
 & FedAdam    &  & 2554 (4.44$\times$)&  & 2778 (4.50$\times$) &  & 2834 (3.22$\times$)\\
 & LocalAdam  &  & \textbf{575} &  & 746 (1.20$\times$) &  & 971 (1.10$\times$)\\
 & \textbf{Our} &  & 765 (1.33$\times$) &  & \textbf{618} &  & \textbf{878}\\
\midrule
 & FedAvg     &  & $\infty$ &  & $\infty$ &  & $\infty$\\
 & FedProx    &  & $\infty$ &  & $\infty$ &  & $\infty$\\
 & SCAFFOLD   &  & 1177 (2.04$\times$) &  & 677 (1.51$\times$) &  & 494 (1.83$\times$)\\
Test & FedCM  & 84.0 & 1211 (2.11$\times$) & 51.0 & 721 (1.61$\times$) & 38.0 & 455 (1.69$\times$)\\
 & FedAdam    &  & $\infty$ &  & 1114 (2.49$\times$) &  & 1397 (5.19$\times$)\\
 & LocalAdam  &  & $\infty$ &  & $\infty$ &  & $\infty$\\
 & \textbf{Our} &  & \textbf{598} &  & \textbf{447} &  & \textbf{269}\\
\bottomrule
\end{tabular}
\end{sc}
\end{small}
\end{center}
%\vskip -0.1in
\end{table*}
\subsection{Experimental Settings}\label{setting}
\textbf{Setup.} We conduct extensive experiments on CIFAR-10/100 and TinyImageNet datasets. CIFAR-10 dataset consists of total 60K 32$\times$32 color images in 10 classes, with 6K images per class. There are 50K training images and 10K test images. CIFAR-100 \cite{CIFAR100} includes 100 categories of 50K training images and 10K test images. The format is the same as CIFAR10. TinyImageNet (MicroImageNet) contains 200 classes of total 110K 64$\times$64 color images, 100K for training, and 10K for testing, which is a miniature of ImageNet close to the real-world dataset. For non-$iid$ dataset, we closely follow \cite{Dirichlet} and sample the label ratios from the Dirichlet distribution with concentration parameter equal to 0.6, which is a common setting to split heterogeneous datasets. We adopt the ResNet-18 \cite{resnet} as the backbone and use the group normalization \cite{GN,hendrycks2016gaussian,biswas2021smu} instead of batch normalization \cite{FLGN}. In order to fairly compare the performance of the various methods, we fix random seeds. We report the best-performing hyperparameter selections in our experiments.
\begin{table}[H]
\centering
\caption{Introduction of Dataset.}
%\vskip 0.15in
\begin{tabular}{|c|c|c|c|c|}
\hline
             & Train  & Test  & Size  & Class \\ \hline
CIFAR10      & 50000  & 10000 & 32*32 & 10    \\ \hline
CIFAR100     & 50000  & 10000 & 32*32 & 100   \\ \hline
TinyImageNet & 100000 & 10000 & 64*64 & 200   \\ \hline
\end{tabular}
\label{dataset}
\end{table}

\noindent
\textbf{Implementation.} Local learning rate $\eta_{l}$ is set as 0.1 in local SGD-based methods and 0.001 in FedLADA. Global learning rate $\eta_{g}$ is set as 1.0 in global SGD-based methods (average of local offsets) and 0.1 in FedAdam. Learning rate decay is set as 0.998 per communication round, which is the same as \cite{FedDyn,FedCM}. The Minibatch size is set as 50 for all datasets. Weight decay is set 0.001 in local SGD-based methods and 0.01 in FedLADA. Local epochs are set as 5 on CIFAR-10 and 2 on CIFAR-100/TinyImageNet. Amended weight $\alpha$ is set as 0.1 on CIFAR-10 and 0.05 on CIFAR-100/TinyImageNet. First-order and second-order momentums $\beta_{1}$ and $\beta_{2}$ are set as 0.9 and 0.99 for the adaptive optimizer, respectively. $\epsilon_{v}$ is set as 1$e$-8. The partial participating rate is set as 10$\%$ per round.

We test a lot of major hyperparameters selections including global learning rate $\eta_{l}$, global learning rate $\eta_{g}$, local intervals $K$, and partial participating rate. For different dataset, we have different options for the best-performing. We split the dataset into 100 local datasets and 1 total test dataset on the global server. Bathsize is set as 50. The dataset pre-process includes common normalization and random cropping on the extra 4/8-pixel padding images. The local learning rate is selected from 0.1 for local SGD-based methods or 0.001 for local adaptive methods. Global learning rate $\eta_{g}$ is selected from 1.0 for global SGD-based methods, which means average aggregation on the global server, and 0.1 for FedAdam. We apply the exponential decayed learning rate as mentioned in \cite{FedDyn,FedCM} and set the decaying coefficient as 0.998 per round. Also emulating their experiments, we apply the $L$-2 weight decay coefficient $\lambda$ as $1e$-3 for SGD-based methods and $1e$-2 for adaptive methods. In the adaptive algorithm, in order to prevent the operation failure caused by division by 0, we set the initial second-order momenta term as $1e$-2 for FedAdam and $1e$-8 for FedLADA. Amended weight $\alpha$ controls the importance of global direction when local clients updates. Partial participating rate is selected from 5$\%$ to 50$\%$. We test and set different values on different datasets.\\\\
\textbf{Baselines.} We fairly compare the performance of applying the adaptive optimizer on either the server or local clients in the FL framework and analyze their major differences. We compare FedLADA with several efficient state-of-the-art (SOTA) baselines including FedAvg \cite{FL_root}, FedProx \cite{FedProx}, SCAFFOLD \cite{SCAFFOLD}, FedCM \cite{FedCM} and FedAdam \cite{fedadam}. FedAvg is a general baseline in FL. FedProx utilizes proximal operators to correct local objective functions and achieves a faster convergence rate. SCAFFOLD introduces a variance reduction technique to alleviate the client drifts. FedCM incorporates the client-level momentum on the local update to tackle the heterogeneity. FedAdam first uses adaptive server optimization in FL. We report both training loss and test accuracy to demonstrate the optimization and generalization capabilities of different methods.
\begin{figure*}[t]
	\centering
	\subfigure[Training loss on CIFAR-10 (L), CIFAR-100 (M) and TinyImageNet (R).]{
		\begin{minipage}[b]{0.99\textwidth}
			\includegraphics[width=0.33\textwidth]{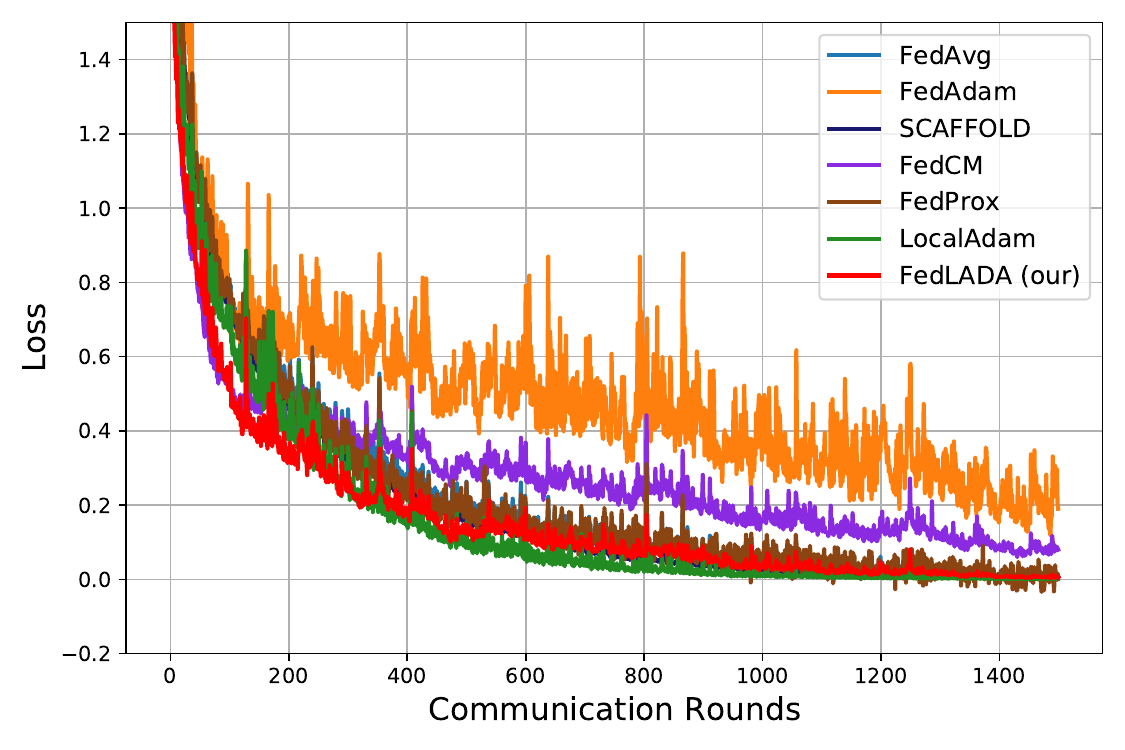} 
			\includegraphics[width=0.33\textwidth]{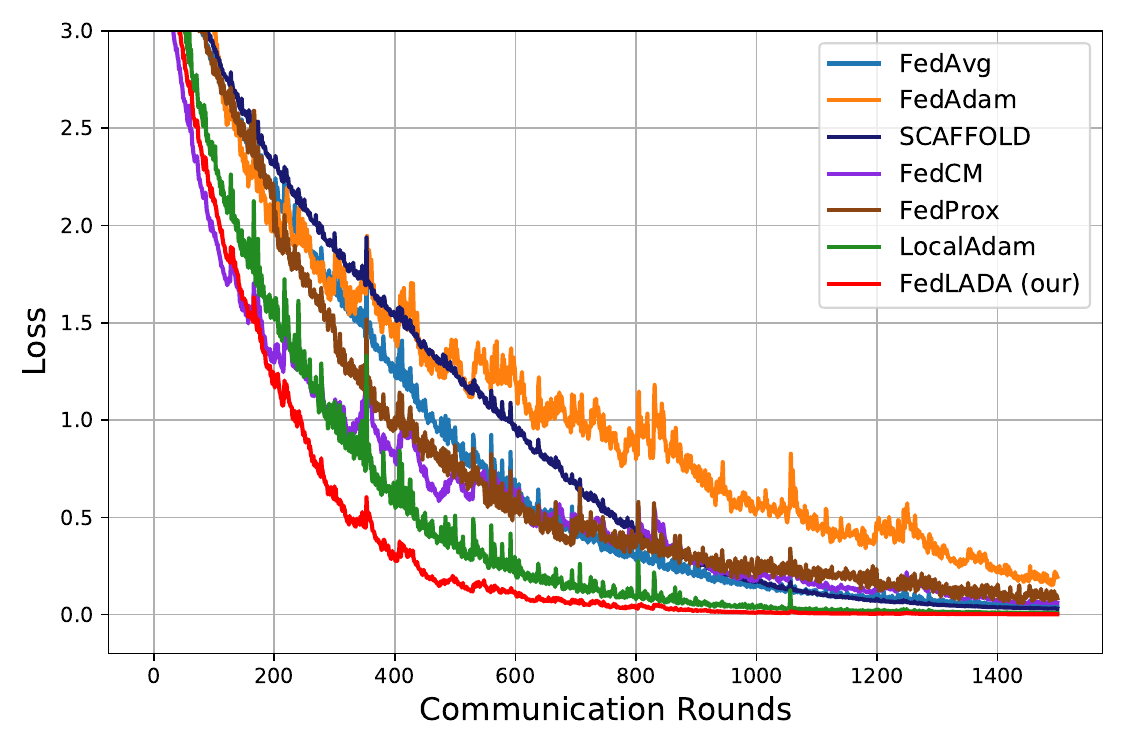}
			\includegraphics[width=0.33\textwidth]{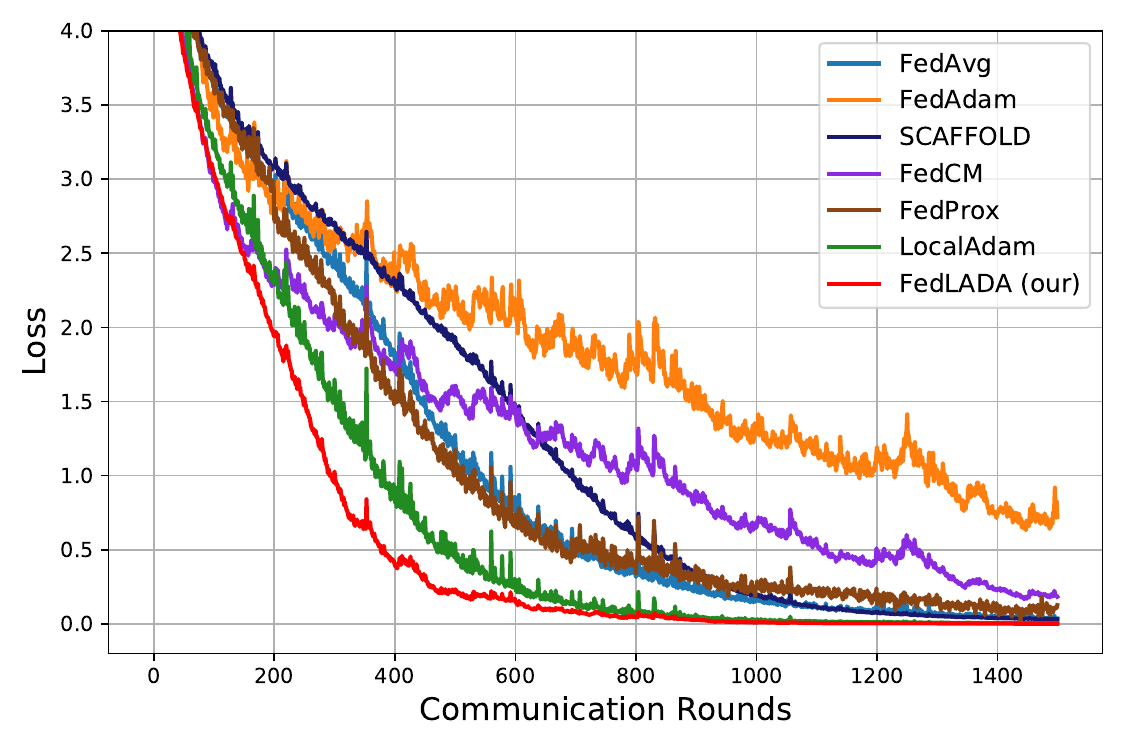}
		\end{minipage}
	}
	%\vskip -0.05in
    \subfigure[Top-1 Accuracy on CIFAR-10 (L), CIFAR-100 (M) and TinyImageNet (R).]{
    	\begin{minipage}[b]{0.99\textwidth}
   		 	\includegraphics[width=0.33\textwidth]{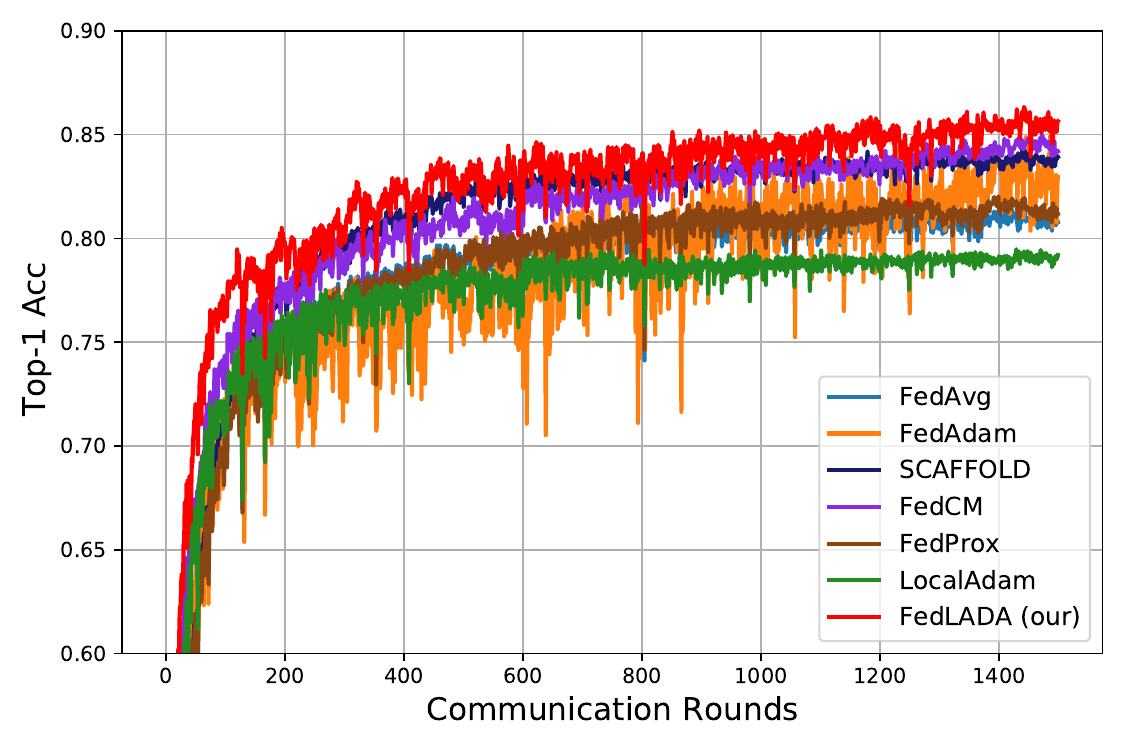}
		 	\includegraphics[width=0.33\textwidth]{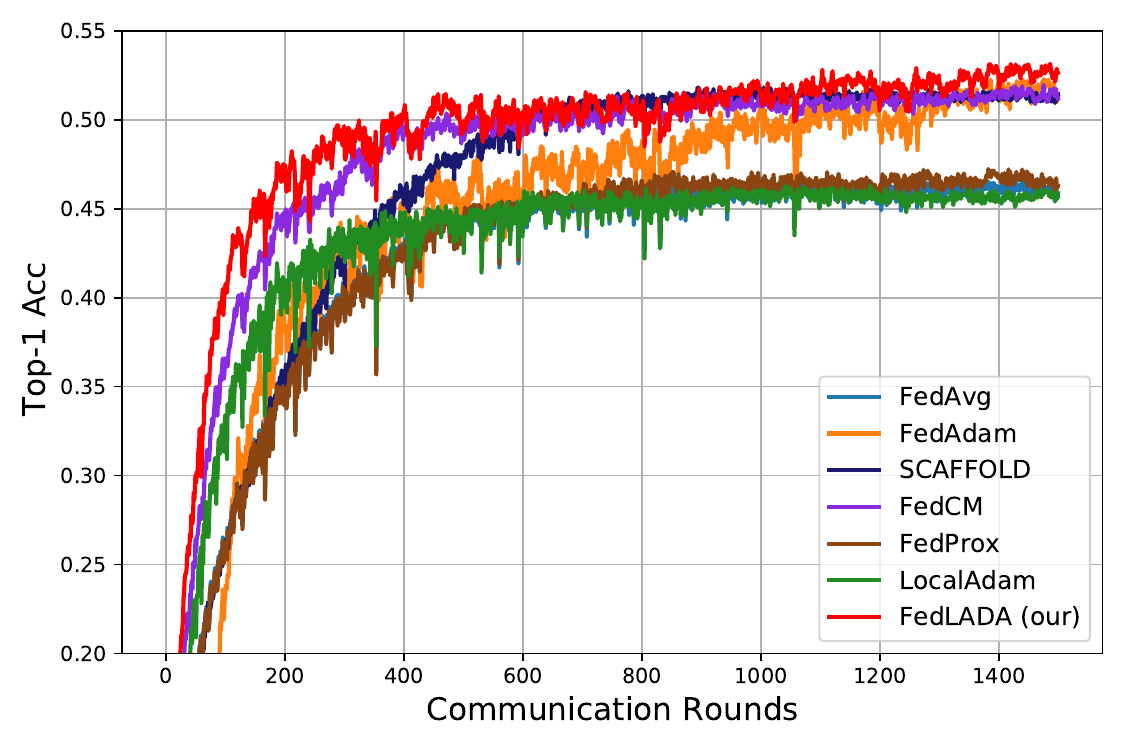}
		 	\includegraphics[width=0.33\textwidth]{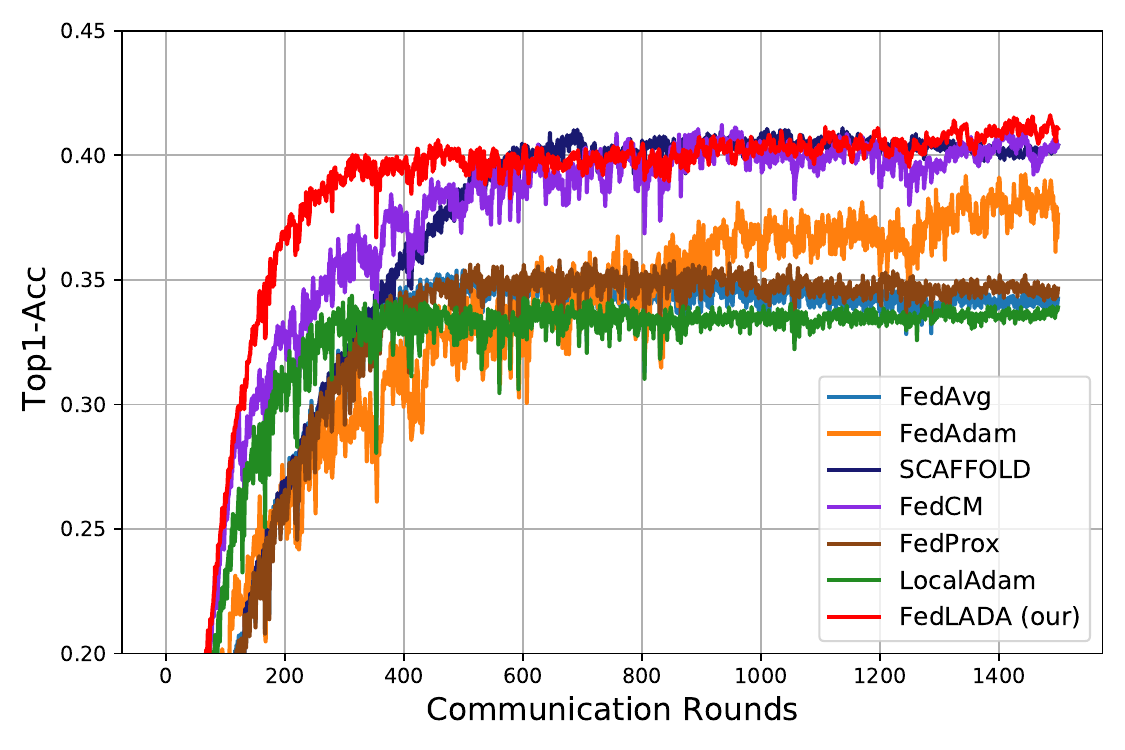}
    	\end{minipage}
	}
	%\vskip -0.05in
	\caption{The training loss and top-1 accuracy in communication rounds of our proposed FedLADA and other baselines. LocalAdam applies adam optimizer on the local clients and SGD on the global server, which is a local adaptive method. Each method updates for 1500 communication rounds. For a fair comparison, the same optimizer is trained with the same hyperparameters.}
	\label{total_res}
\end{figure*}

\subsection{Experiments on CIFAR10/100 and TinyImageNet}\label{Experiments}

\begin{table}[t]
\caption{ResNet-18 performance of different methods distinguish by optimizer combination on TinyImageNet after 1500 rounds training. (+A) represents the amended optimizer. FedAdam(*) is FedAdam incorporated by client-level momentum.}
\label{acc}
%\vskip 0.05in
\begin{center}
\begin{small}
\begin{sc}
\begin{tabular}{llccc}
\toprule
Mode & Global & Local & Acc.($\%$)\\
\midrule
 FedAvg    & SGD & SGD & 34.5\\
 FedCM    & SGD & SGD(+A) & +6.6\\
\midrule
 FedAdam   & Adam & SGD & +4.3\\
 FedAdam(*)  & Adam & SGD(+A) &+0.5\\
\midrule
 LocalAdam  & SGD & Adam & -0.4\\
 \textbf{Our}  & \textbf{SGD} & \textbf{Adam(+A)} & \textbf{+7.4}\\
\bottomrule
\end{tabular}
\end{sc}
\end{small}
\end{center}
%\vskip -0.1in
\end{table}

In Table~\ref{speed}, $\infty$ means impossible in the training process. The training convergence speed of FedAdam is the slowest. LocalAdam enjoys a faster convergence speed than the whole baselines, while it suffers from heterogeneous over-fitting. Our proposed FedLADA inherits the efficiency of local adaptive and achieves approximately 1.1$\times$ convergence speed improvements than LocalAdam, over 1.5$\times$ than the best SGD-based method, and over 3$\times$ than FedAdam.

Figure~\ref{total_res} shows the performance of ResNet-18 trained on CIFAR-10/100 and TinyImageNet. Our proposed FedLADA outperforms other methods both on convergence speed and top-1 test accuracy. Local adaptive amended optimizer is stable and robust on the FL framework. On TinyImageNet, FedLADA improves approximately 7.4$\%$ ahead of FedAvg, 7.8$\%$ ahead of LocalAdam, and slightly a little than other SGD-based baselines while reducing about one-third of communication rounds over best-performing SGD-based methods and achieves 1.1$\times$ faster than LocalAdam. On CIFAR100, FedCM performs stronger generalization performance among SGD-based methods, which improves approximately 6$\%$ ahead of FedAvg. Our proposed FedLADA outperforms FedCM by 1$\%$ after 1500 rounds. FedAdam shows strong generalization on CIFAR100, while the convergence speed is still the bottleneck. From the standpoint of optimization, FedLADA converges 3$\times$ faster than FedAdam. CIFAR10 is a relatively simple challenge. FedLADA prevails slightly over FedCM, and achieves a similar convergence speed to many local SGD-based methods. The improvement of accuracy is about 4.2$\%$ over FedAvg and 7.5$\%$ over LocalAdam. 
It can be seen that in simple tasks such as CIFAR10, the client drifts caused by local heterogeneous over-fitting is more serious. The locally amended technique can effectively alleviate this difficulty in FL framework.

Table~\ref{acc} presents the performance of different methods according to combinations of the optimizer. Compared with the global adaptive optimizer, the local adaptive optimizer benefits more from the amended technique. Our proposed FedLADA further improves the efficiency of the local adaptive optimizer while utilizing the locally amended technique to alleviate the heterogeneous over-fitting. This inspires us to rethink the FL framework. It is a trade-off between generalization and convergence even with the help of efficient optimizers. How to reasonably apply more efficient optimizers to help federated optimization is a question worth pondering. And, effectively improving both training and communication efficiency is still a promising study.
\subsection{Sensitivity for Hyperparameters}
\begin{figure}[t]
\centering
\setlength{\abovecaptionskip}{0.cm}
	    \begin{minipage}[t]{0.49\linewidth}
		    \centering
		    \subfigure{
		    \includegraphics[width=0.99\linewidth]{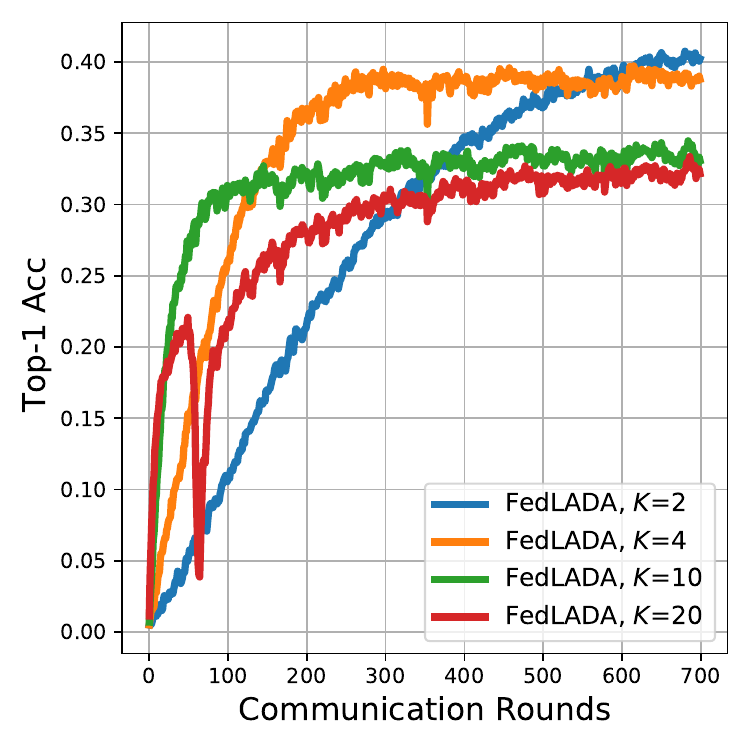}
		    }
		    \end{minipage}
		 \begin{minipage}[t]{0.49\linewidth}
		    \centering
		    \subfigure{
		    \includegraphics[width=0.99\linewidth]{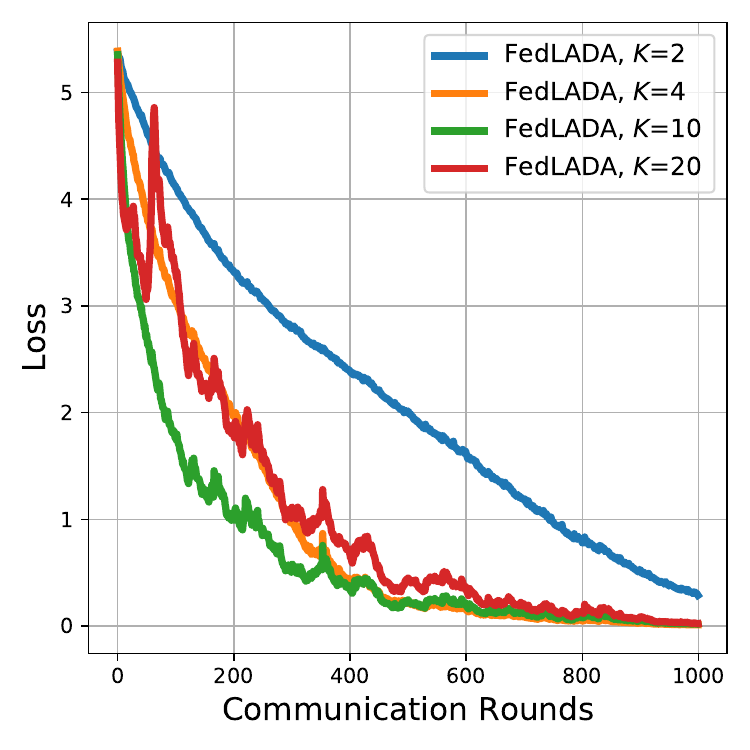}
		    }
	    \end{minipage}
	%\vskip -0.05in
	\caption{Test accuracy (Left) and training loss (Right) of FedLADA with different intervals $K$ on TinyImageNet. $K$ is set as 2, 4, 10, 20, and other hyperparameters are set as mentioned above.}
	\label{different_K}
	%\vskip -0.1in
\end{figure}

\begin{table}[t]
\caption{Performance of FedLADA with different intervals $K$ on TinyImageNet. Top1-Acc is the best accuracy on the test set and Rounds means the communication rounds to achieve the best accuracy.}
\label{K}
%\vskip 0.05in
\begin{center}
\begin{small}
\begin{sc}
\begin{tabular}{ccccc}
\toprule
 K & 2 & 4 & 10 & 20\\
\midrule
 Top1-Acc ($\%$) & $\textbf{41.9}$ & 40.2 & 36.2 & 33.8\\
 Rounds   & 878 & $\textbf{614}$ & 959 & 1327\\
\bottomrule
\end{tabular}
\end{sc}
\end{small}
\end{center}
%\vskip -0.1in
\end{table}
\textbf{Local Intervals $K$} measures the progress of local optimization. In Theorem~\ref{convergence_rate}, when $T$ is large enough, increasing $K$ can help the global model to achieve higher convergence speed. Table~\ref{K} shows the performance of $K=2,4,10,20$. Increasing $K$ will not only improve the convergence speed but also leads to a stronger negative impact of local heterogeneous over-fitting. It is a trade-off to balance the convergence and the generalization, as shown in Table~\ref{K}, which should be selected properly. Figure~\ref{different_K} shows the effects of different intervals $K$. In FL frameworks, usually, $K$ is used as a local computing cost and we do not pay more attention to this. And for some specific methods like FedProx and FedDyn, $K$ is expected to be large enough to approach the sub-optimum of the proxy-objective functions. However, in most deep training of FL, $K$ is a trade-off of balancing the training convergence speed and local over-fitting. When $K=2$, FedLADA performs a lower training convergence speed while its performance of generalization is the best of the four curves. When $K$ is increased to 4, it performs approximately 2.2$\times$ faster than $K=2$ which obeys the dominant term $\mathbf{O}(\frac{1}{\sqrt{SKT}})$ in our theoretical analysis. As $K$ continues to increase, the speedup property exists but is gradually insignificant, and the performance of generalization drops sharply. We analyze the reasons as follows:
\begin{itemize}
    \item From the experimental point of view, larger $K$ means more updates on the local dataset, which forces the local parameter $\mathbf{x}_{i,\tau}^{t}$ closed to local optimum $\mathbf{x}_{i}^{*}$. This causes severe client drifts that hurt the generalization. 
    \item In the theoretical analysis, we consider the second-dominant term of $\mathbf{O}(\frac{K}{T})$. When $K$ is increased as $\mathbf{O}(\Big(\frac{T}{S}\Big)^{\frac{1}{3}})$, the second-dominant term will be the main influence of the convergence rate instead of $\mathbf{O}(\frac{1}{\sqrt{SKT}})$. This phenomenon is also verified in the experiments. When $K$ increases from 10 to 20, the convergence speed does not continue to increase and the training process becomes unstable due to the strong client drifts.
\end{itemize}
For the federated non-convex optimization, most frameworks are eager to search for a suitable $K$ to balance the trade-off of convergence rate and generalization performance.\\

\begin{figure}[t]
    \setlength{\abovecaptionskip}{0.cm}
	\begin{minipage}[t]{0.495\linewidth}
		\centering
		\includegraphics[width=0.99\linewidth]{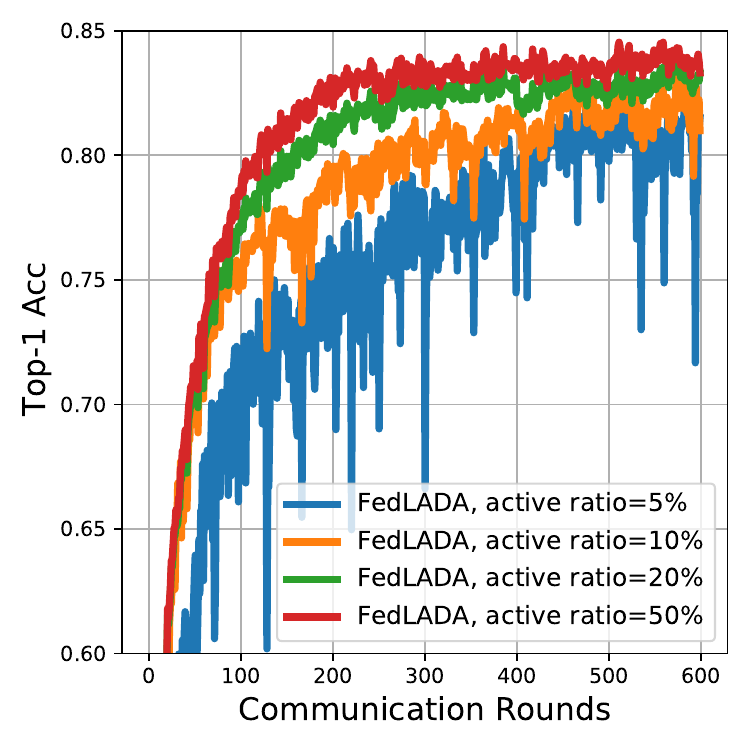}
	\end{minipage}
	\begin{minipage}[t]{0.495\linewidth}
		\centering
		\includegraphics[width=0.99\linewidth]{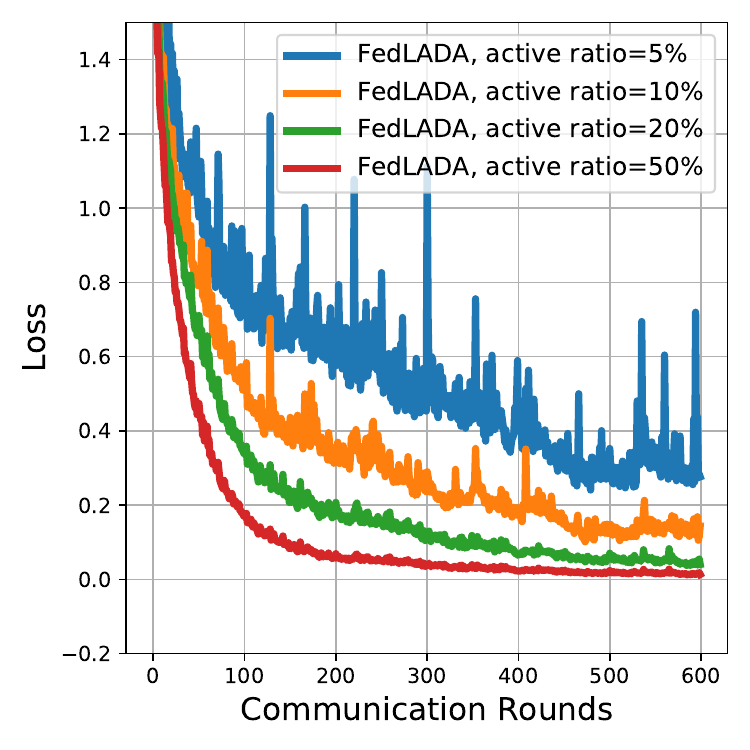}
	\end{minipage}
	%\vskip -0.1in
	\caption{Test accuracy (left) and training loss (right) of FedLADA with different partial participating ratios on CIFAR10.}
	\label{2extraexp}
	%\vskip -0.1in
\end{figure}

\textbf{Partial Participation} is a practical technique to reduce communication costs. We prove in the Remark~\ref{111} that the dominant term of the Algorithm~\ref{algorithm_fedLada} satisfies the property of linear speedup convergence when $T$ is large enough. The improvement of increasing active clients is obvious when the participation ratio is low and gradually weakens nearly full participation, as shown in Figure~\ref{2extraexp}. The ratios are set as 5$\%$, 10$\%$, 20$\%$, and 50$\%$ respectively.

\begin{table}[h]
\caption{FedLADA performance of different $\alpha$ on TinyImageNet.}
\label{table:alpha}
%\vskip 0.05in
\begin{center}
\begin{small}
\begin{sc}
\begin{tabular}{ccccccc}
\toprule
 $\alpha$ & 0.01 & $\textbf{0.05}$ & 0.1 & 0.2 & 0.5 & 1.0\\
\midrule
 Acc ($\%$)   & 29.5 & $\textbf{41.9}$ & 40.1 & 38.8 & 36.5 & 34.1\\
\bottomrule
\end{tabular}
\end{sc}
\end{small}
\end{center}
%\vskip -0.1in
\end{table}
\textbf{Amended Weight $\alpha$} determines how much historical global information to use as a guide for the current round of local updates. According to the Remark~\ref{alpha}, the best value for $\alpha$ is a constant belonging to $(0,1)$. We test different values of $\alpha$ and show the results in Table~\ref{table:alpha}.
$\alpha=1.0$ represents the local-adaptive scene. Intuitively, small $\alpha$ results in insufficient learnable local information, introducing slow convergence speed and training oscillation on the global server, while large $\alpha$ makes FedLADA degenerate into a local-adaptive scene that suffers from severe over-fitting. In the experiments, the performance achieves the best when the value of $\alpha$ is set as 0.05 on TinyImageNet.

\subsection{Summary of Finetuning Hyperparameters} Finetuning the hyperparameters will bring some tricky benefits to the experiments. We fix most of the hyperparameters to the same value in different methods in order to avoid controversial results due to the specific hyperparameters, which ensures a fair comparison of the performance between the various methods as far as possible. Although the following tricks were not used in our reported experiments, we summarize these as follows:
\begin{itemize}
    \item (Batchsize) Typically, batchsize is a trade-off to balance the convergence speed and test accuracy, which means larger batchsize brings faster convergence speed while being easy to fall into local over-fitting. When training on a small dataset like CIFAR10, we obverse the accuracy and training speed both benefits from smaller batchsize, which is approximately 0.7$\%$ improvement when batchsize decreases from 50 to 25 on FedAvg and 0.4$\%$ improvement on SCAFFOLD.
    \item (Local Intervals) In FL frameworks, we usually do not care about the number of local intervals, which is defined as the local training costs. Some works point out that certain conditions should be satisfied between the total rounds $T$ and the local interval $K$ to guarantee the upper bound of the convergence rate. In actual training, larger $K$ is easier to fall into local over-fitting. For different datasets, the best-performing $K$ is different. The results of our tests show that choosing a suitable $K$ can bring at least 1$\%$ improvement to the test performance.
    \item (Learning Rate) Learning rate is a common factor affecting training performance. Usually, a larger learning rate results in faster training speed and lower generalization performance. This phenomenon is obvious in the experiments of FedAvg and local adaptive optimizers. However, our experiments on CIFAR100 confirm that reducing the initial value of the learning rate and increasing the learning rate decay can improve the generalization performance, while on TinyImageNet it does not hold. In addition, adaptive methods benefit more from larger initial learning rates and smaller learning rate decay due to the influence of the second-order momenta term.
\end{itemize}
\section{Discussion: Global v.s. Local Adaptive}
\label{Global_vs_Local}
In this part, we show the difference between applying an adaptive optimizer on the global server and local clients in FL framework and reveal the potential instructive meaning beyond the experiments, which can inspire us to design a more efficient FL framework.

We describe the training loop from the perspective of the global optimizer as: (1) the global optimizer provides local clients with staged initial parameters $\mathbf{x}^{t}$; (2) the pseudo global gradient is calculated as the averaged local gradients amended by last pseudo gradient after $K$ local updates; (3) global optimizer updates parameters with pseudo gradient. FedAdam converges much slower than FedCM as shown in Figure~\ref{total_res} and Table~\ref{speed}. We compare FedCM with the amended FedAdam. Their difference is that the global optimizer is SGD for FedCM and Adam for FedAdam. Both apply local SGD with client-level momentum. Global learning rate $\eta_{g}$ is set as 0.1 in vanilla FedAdam. 
\begin{figure}[t]
    \setlength{\abovecaptionskip}{0.cm}
	\begin{minipage}[t]{0.495\linewidth}
		\centering
		\subfigure[Different $\alpha$ with $\eta=0.1$]{
		\includegraphics[width=0.99\linewidth]{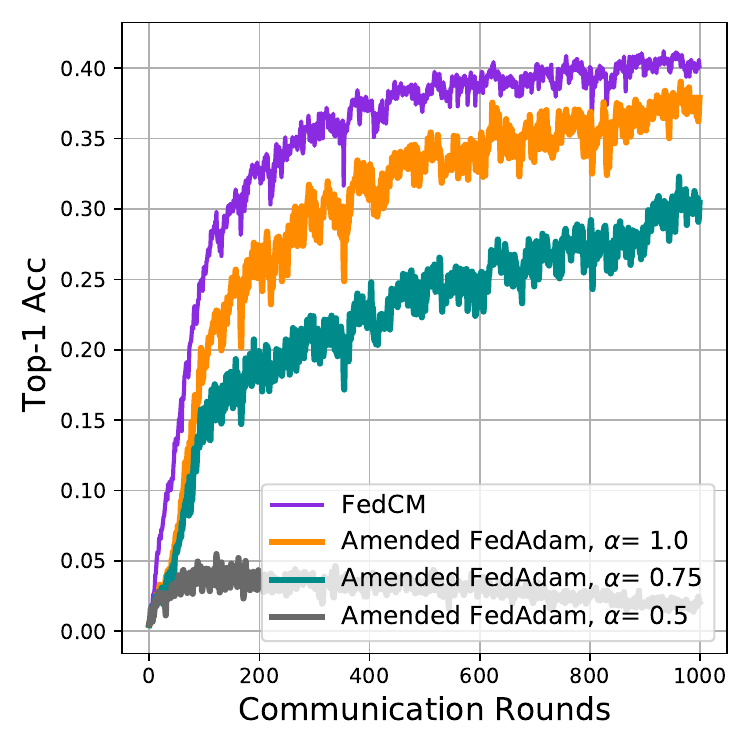}
		}
	\end{minipage}
	\begin{minipage}[t]{0.495\linewidth}
		\centering
		\subfigure[Different $\eta_{g}$ with $\alpha=0.1$]{
		\includegraphics[width=0.99\linewidth]{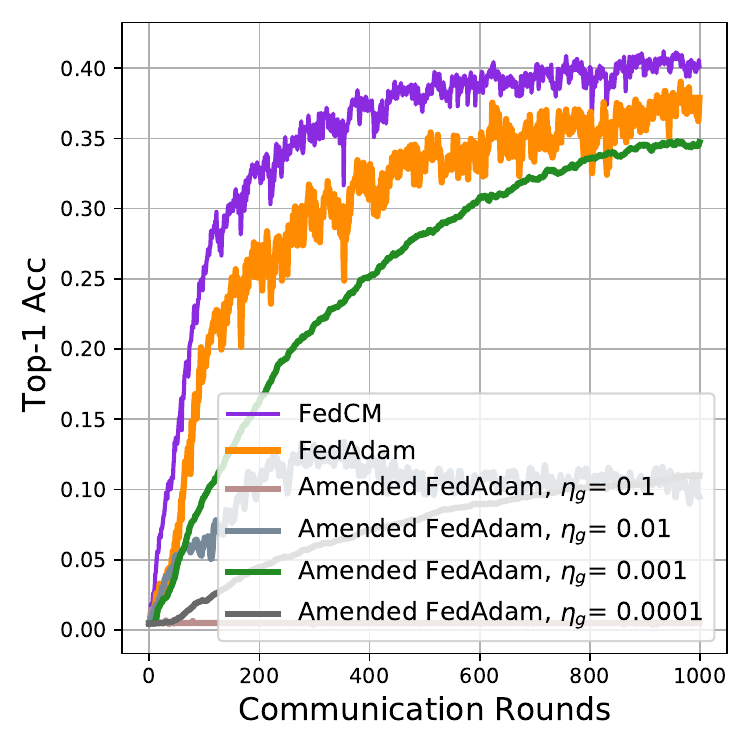}
		}
	\end{minipage}
	\vskip -0.05in
	\caption{Performance of amended FedAdam and FedCM of (a) different $\alpha$ with fixed $\eta_{g}=0.1$ and (b) different $\eta_{g}$ with fixed $\alpha=0.1$ on TinyImageNet. We test for selecting the best hyperparameters.}
	\label{Fedadam_cm}
	\vskip -0.1in
\end{figure}

 Unfortunately, global adaptive methods like FedAdam could not efficiently benefit from the local amended technique. We test the sensitivity of the performance on the coefficient $\alpha$ and global learning rate $\eta_g$ respectively. As shown in Fig.~\ref{Fedadam_cm}~(a), $\alpha=1.0$ represents the FedAdam method. When the amended coefficient decreases, the accuracy decreases extremely. We also search for the optimal global learning rate from $\left[0.1, 0.01, 0.,001, 0.0001\right]$. Fig.~\ref{Fedadam_cm}~(b) indicates that even adopting the best selection still makes the training process much slower than the vanilla FedAdam. These phenomena validate our analysis above and demonstrate the locally amended term cannot effectively revise the bias of the global adaptive update. Applying both global and local adaptive optimizers has very large local biases in FL. As studied in~\cite{local_adaptive,local_adaAlter}, local adaptive optimizer leads to serious inconsistency across local clients due to the local heterogeneity. Therefore, compared with the local SGD optimizer~\cite{localsgd}, the local adaptive optimizer generates more biases when they are aggregated on the global server. Then, as claimed in~\cite{fedadam}, the global adaptive optimizer uses the aggregated local update as the quasi-gradient for the global model. When local updates are extremely far away from each other, the second-order momenta of this quasi-gradient becomes extremely unstable. This also leads to the fact that the two-stage adaptive optimizer is still a very difficult challenge in federated scenarios.

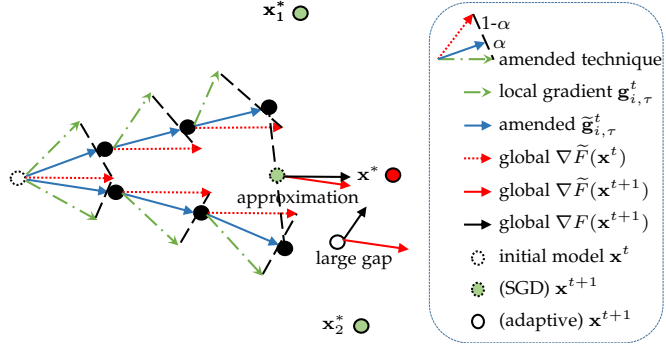
\begin{figure}[t]
	\begin{minipage}[t]{0.99\linewidth}
		\centering
		\begin{overpic}[width=0.99\textwidth]{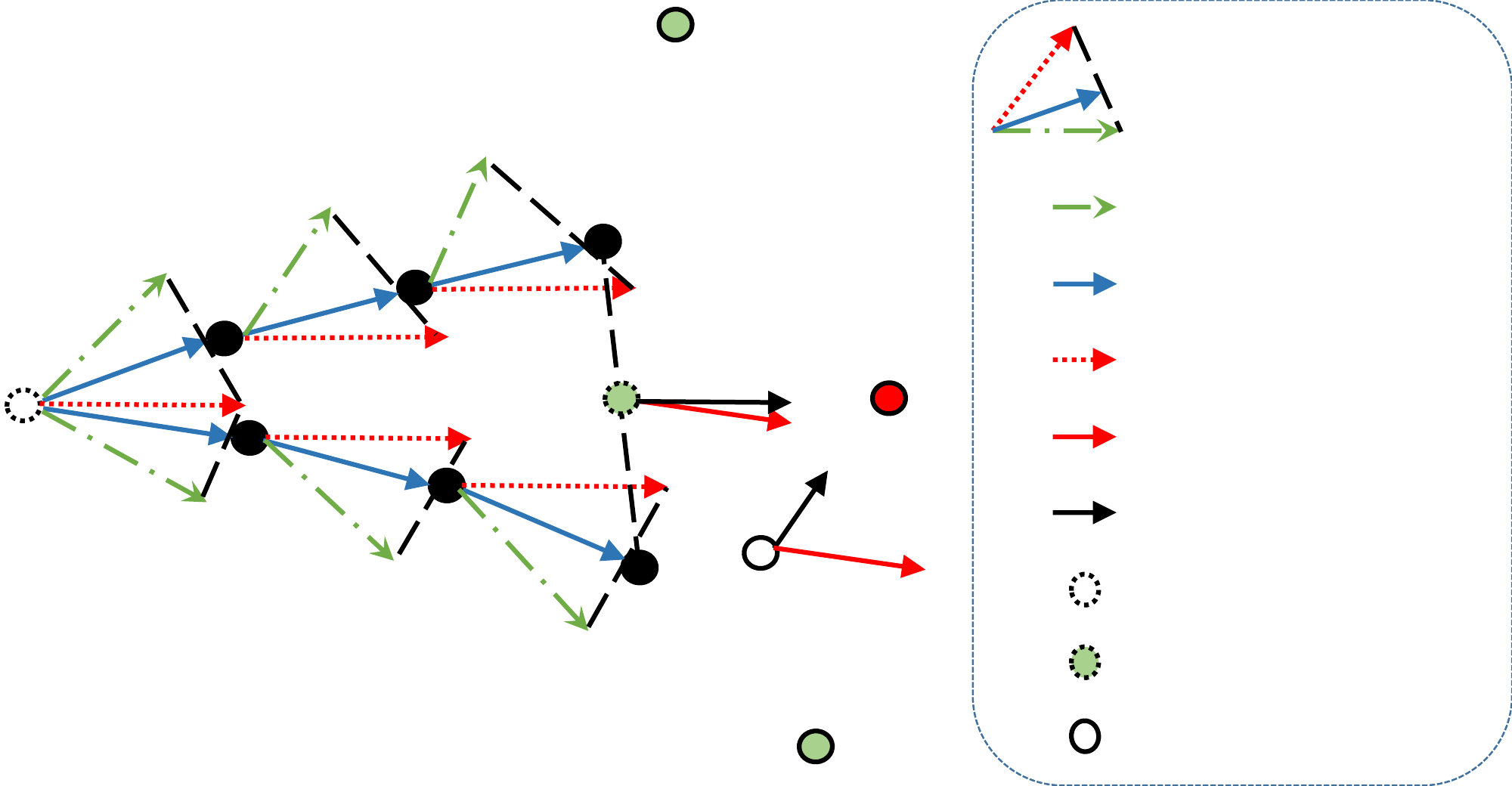}
		    \put(75,42.5){\scriptsize{amended technique}}
		    \put(75,37.5){\scriptsize{local gradient $\mathbf{g}_{i,\tau}^{t}$}}
		    \put(75,32.5){\scriptsize{amended $\widetilde{\mathbf{g}}_{i,\tau}^{t}$}}
		    \put(75,27.5){\scriptsize{global $\nabla \widetilde{F}(\mathbf{x}^{t})$}}
		    \put(75,22.5){\scriptsize{global $\nabla \widetilde{F}(\mathbf{x}^{t+1})$}}
		    \put(75,17.5){\scriptsize{global $\nabla F(\mathbf{x}^{t+1})$}}
		    \put(75,12.3){\scriptsize{initial model $\mathbf{x}^{t}$}}
		    \put(75,7.4){\scriptsize{(SGD) $\mathbf{x}^{t+1}$}}
		    \put(75,2.5){\scriptsize{(adaptive) $\mathbf{x}^{t+1}$}}

            \put(35,21.3){\scriptsize{approximation}}
		    \put(47,12){\scriptsize{large gap}}
		    \put(53.3,24.8){\scriptsize{$\mathbf{x}^{*}$}}
		    \put(39,50){\scriptsize{$\mathbf{x}_{1}^{*}$}}
		    \put(48,2.1){\scriptsize{$\mathbf{x}_{2}^{*}$}}
		    \put(72.3,48){\scriptsize{1-$\alpha$}}
		    \put(74,45){\scriptsize{$\alpha$}}
		    \put(39,50){\scriptsize{$\mathbf{x}_{1}^{*}$}}
		\end{overpic}
	\end{minipage}
	\vskip -0.05in
	\caption{A toy schematic to introduce the "matching dislocation" caused by global adaptive optimizer with local amended technique, where $m$=2 and $K$=3. $\mathbf{x}_{i}^{*}$ and $\mathbf{x}^{*}$ respectively represents for the local and global optimum. Amended gradient $\hat{\mathbf{g}}_{i,\tau}^{t}$ = $\alpha \mathbf{g}_{i,\tau}^{t}$+ (1-$\alpha)\nabla \widetilde{F}(\mathbf{x}_{i,\tau}^{t})$. (SGD) $\mathbf{x}^{t+1}$ and (adaptive) $\mathbf{x}^{t+1}$ respectively means the global model updated by global SGD and global adaptive optimizer. Since the global adaptive optimizer scales the pseudo gradient by second-order momenta $\vartheta$, there is a large gap between the true global gradient and the pseudo gradient at (adaptive) $\mathbf{x}^{t+1}$ as shown above. When it is set as the initial model at $t$+1 round, local client yields larger gaps between $\nabla \widetilde{F}(\mathbf{x}^{t+1})$ and $\nabla F(\mathbf{x}^{t+1})$.}
	\vskip -0.1in
	\label{notwork}
\end{figure}

 The core challenge is whether the pseudo gradient of $\mathbf{x}^{t}$ can truly and effectively represent the global descent direction at $\mathbf{x}^{t+1}$. For FedCM whose global optimizer is SGD, the global model inherits information from averaged gradient descent, in other words, the following formula satisfies:
 \begin{equation}
    \mathbb{E}[\nabla \widetilde{F}(\mathbf{x}^{t} - \eta_{g}\nabla \widetilde{F}(\mathbf{x}^{t}))] \approx \mathbb{E}[\frac{1}{mK}\sum_{i,\tau}^{K} \mathbf{g}_{i,\tau}^{t}],
 \end{equation}
 where $\nabla \widetilde{F}$ denotes the global pseudo gradient.
 For FedAdam embedded client-level momentum whose global optimizer applies adam, second-order momenta term changes the global direction. In the long run, the adaptivity does not hurt the convergence of the entire sequence $\{\mathbf{x}^{t}\}$. While under the same $\eta_{g}$, global parameters $\mathbf{x}^{t+1}=\mathbf{x}^{t}-\eta_{g}\nabla \widetilde{F}(\mathbf{x}^{t})\odot\widetilde{\vartheta}^{t}$ generated by global adam optimizer in FedAdam is far away from the global parameters generated by SGD in FedCM for the estimated second-order momenta term $\widetilde{\vartheta}^{t}$. The major effort of the locally amended technique is to introduce global direction as an auxiliary during local updates to avoid the local client falling into the local optimum completely. The gradient information at the new global parameters $\mathbf{x}^{t+1}$ generated by Adam is unknown and cannot be predicted efficiently by the local gradients from the previous round $t$, which leads to catastrophic 
 convergence. In short, the local SGD method can not effectively estimate the gradient at the point where the global parameters are updated by the non-SGD method, e.g. for Adam in our experiments, what we call “matching dislocation”. We illustrate this process with a simple illustration in Figure~\ref{notwork}.
When we decay the global learning rate $\eta_{g}$, the estimation error of the global gradient at $\mathbf{x}^{t+1}$ will be reduced when the distance between the two adjacent parameters $\mathbf{x}$ is close enough. However, the decayed global learning rate directly leads to a tardy training speed, which seriously hurts the convergence speed in practical scenarios. This gives us more insights into how to apply the efficient optimizer to serve FL frameworks. 
%A higher-performance FL system should implement both highly efficient local optimizer and powerful local amendment to both achieve higher convergence speed and alleviate local over-fitting.
\section{Conclusions}
In this paper, we explore the major challenges of applying efficient adaptive optimizers in the practical FL framework. We test the performance of global and local adaptive methods in FL and analyze the trade-off on convergence speed and generalization. We elaborate on the main difficulties when incorporating a global adaptive optimizer with a locally amended technique in deep network training. To tackle the heterogeneous over-fitting, we propose FedLADA, a novel local adaptive method adopted with a locally amended technique, which achieves linear speedup on non-convex settings in our theoretical proof. Extensive experiments on CIFAR10/100 and TinyImageNet verify the efficiency of FedLADA, which inherits the fast convergence speed of local adaptive optimizer while further improving the global generalization. This work inspires practical federated framework designs when applying the more efficient optimizer, e.g. for accelerated optimization and second-order optimization.

% if have a single appendix:
%\appendix[Proof of the Zonklar Equations]
% or
%\appendix  % for no appendix heading
% do not use \section anymore after \appendix, only \section*
% is possibly needed

% use appendices with more than one appendix
% then use \section to start each appendix
% you must declare a \section before using any
% \subsection or using \label (\appendices by itself
% starts a section numbered zero.)
%

% \appendices
% \section{Proof of the First Zonklar Equation}
% Appendix one text goes here.

% % you can choose not to have a title for an appendix
% % if you want by leaving the argument blank
% \section{}
% Appendix two text goes here.

% Can use something like this to put references on a page
% by themselves when using endfloat and the captionsoff option.
\ifCLASSOPTIONcaptionsoff
  \newpage
\fi

% trigger a \newpage just before the given reference
% number - used to balance the columns on the last page
% adjust value as needed - may need to be readjusted if
% the document is modified later
%\IEEEtriggeratref{8}
% The "triggered" command can be changed if desired:
%\IEEEtriggercmd{\enlargethispage{-5in}}

% references section

% can use a bibliography generated by BibTeX as a .bbl file
% BibTeX documentation can be easily obtained at:
% http://mirror.ctan.org/biblio/bibtex/contrib/doc/
% The IEEEtran BibTeX style support page is at:
% http://www.michaelshell.org/tex/ieeetran/bibtex/
%\bibliographystyle{IEEEtran}
% argument is your BibTeX string definitions and bibliography database(s)
%\bibliography{IEEEabrv,../bib/paper}
%
% <OR> manually copy in the resultant .bbl file
% set second argument of \begin to the number of references
% (used to reserve space for the reference number labels box)
% \begin{thebibliography}{1}

% \bibitem{IEEEhowto:kopka}
% H.~Kopka and P.~W. Daly, \emph{A Guide to \LaTeX}, 3rd~ed.\hskip 1em plus
%   0.5em minus 0.4em\relax Harlow, England: Addison-Wesley, 1999.

% \end{thebibliography}
\bibliographystyle{IEEEtran} 
\bibliography{main.bib}

% Generated by IEEEtran.bst, version: 1.14 (2015/08/26)
\begin{thebibliography}{10}
\providecommand{\url}[1]{#1}
\csname url@samestyle\endcsname
\providecommand{\newblock}{\relax}
\providecommand{\bibinfo}[2]{#2}
\providecommand{\BIBentrySTDinterwordspacing}{\spaceskip=0pt\relax}
\providecommand{\BIBentryALTinterwordstretchfactor}{4}
\providecommand{\BIBentryALTinterwordspacing}{\spaceskip=\fontdimen2\font plus
\BIBentryALTinterwordstretchfactor\fontdimen3\font minus
  \fontdimen4\font\relax}
\providecommand{\BIBforeignlanguage}[2]{{%
\expandafter\ifx\csname l@#1\endcsname\relax
\typeout{** WARNING: IEEEtran.bst: No hyphenation pattern has been}%
\typeout{** loaded for the language `#1'. Using the pattern for}%
\typeout{** the default language instead.}%
\else
\language=\csname l@#1\endcsname
\fi
#2}}
\providecommand{\BIBdecl}{\relax}
\BIBdecl

\bibitem{FL_root}
B.~McMahan, E.~Moore, D.~Ramage, S.~Hampson, and B.~A. y~Arcas,
  ``Communication-efficient learning of deep networks from decentralized
  data,'' in \emph{Proceedings of the 20th International Conference on
  Artificial Intelligence and Statistics, {AISTATS} 2017, 20-22 April 2017,
  Fort Lauderdale, FL, {USA}}, ser. Proceedings of Machine Learning Research,
  A.~Singh and X.~J. Zhu, Eds., vol.~54.\hskip 1em plus 0.5em minus 0.4em\relax
  {PMLR}, 2017, pp. 1273--1282.

\bibitem{FL_intro3}
Q.~Yang, Y.~Liu, T.~Chen, and Y.~Tong, ``Federated machine learning: Concept
  and applications,'' \emph{{ACM} Trans. Intell. Syst. Technol.}, vol.~10,
  no.~2, pp. 12:1--12:19, 2019.

\bibitem{FL_intro2}
T.~Li, A.~K. Sahu, A.~Talwalkar, and V.~Smith, ``Federated learning:
  Challenges, methods, and future directions,'' \emph{{IEEE} Signal Process.
  Mag.}, vol.~37, no.~3, pp. 50--60, 2020.

\bibitem{FL_intro1}
P.~Kairouz, H.~B. McMahan, B.~Avent, A.~Bellet, M.~Bennis, A.~N. Bhagoji, K.~A.
  Bonawitz, Z.~Charles, G.~Cormode, R.~Cummings, R.~G.~L. D'Oliveira,
  H.~Eichner, S.~E. Rouayheb, D.~Evans, J.~Gardner, Z.~Garrett,
  A.~Gasc{\'{o}}n, B.~Ghazi, P.~B. Gibbons, M.~Gruteser, Z.~Harchaoui, C.~He,
  L.~He, Z.~Huo, B.~Hutchinson, J.~Hsu, M.~Jaggi, T.~Javidi, G.~Joshi,
  M.~Khodak, J.~Kone{\v{c}}n{\'y}, A.~Korolova, F.~Koushanfar, S.~Koyejo,
  T.~Lepoint, Y.~Liu, P.~Mittal, M.~Mohri, R.~Nock, A.~{\"{O}}zg{\"{u}}r,
  R.~Pagh, H.~Qi, D.~Ramage, R.~Raskar, M.~Raykova, D.~Song, W.~Song, S.~U.
  Stich, Z.~Sun, A.~T. Suresh, F.~Tram{\`{e}}r, P.~Vepakomma, J.~Wang,
  L.~Xiong, Z.~Xu, Q.~Yang, F.~X. Yu, H.~Yu, and S.~Zhao, ``Advances and open
  problems in federated learning,'' \emph{Found. Trends Mach. Learn.}, vol.~14,
  no. 1-2, pp. 1--210, 2021.

\bibitem{localsgd}
S.~U. Stich, ``Local {SGD} converges fast and communicates little,'' in
  \emph{7th International Conference on Learning Representations, {ICLR} 2019,
  New Orleans, LA, USA, May 6-9, 2019}.\hskip 1em plus 0.5em minus 0.4em\relax
  OpenReview.net, 2019.

\bibitem{SlowMo}
J.~Wang, V.~Tantia, N.~Ballas, and M.~G. Rabbat, ``Slowmo: Improving
  communication-efficient distributed {SGD} with slow momentum,'' in \emph{8th
  International Conference on Learning Representations, {ICLR} 2020, Addis
  Ababa, Ethiopia, April 26-30, 2020}.\hskip 1em plus 0.5em minus 0.4em\relax
  OpenReview.net, 2020.

\bibitem{FedDyn}
D.~A.~E. Acar, Y.~Zhao, R.~M. Navarro, M.~Mattina, P.~N. Whatmough, and
  V.~Saligrama, ``Federated learning based on dynamic regularization,'' in
  \emph{9th International Conference on Learning Representations, {ICLR} 2021,
  Virtual Event, Austria, May 3-7, 2021}.\hskip 1em plus 0.5em minus
  0.4em\relax OpenReview.net, 2021.

\bibitem{FedADC}
E.~Ozfatura, K.~Ozfatura, and D.~G{\"{u}}nd{\"{u}}z, ``Fedadc: Accelerated
  federated learning with drift control,'' in \emph{{IEEE} International
  Symposium on Information Theory, {ISIT} 2021, Melbourne, Australia, July
  12-20, 2021}.\hskip 1em plus 0.5em minus 0.4em\relax {IEEE}, 2021, pp.
  467--472.

\bibitem{fedadam}
S.~J. Reddi, Z.~Charles, M.~Zaheer, Z.~Garrett, K.~Rush, J.~Kone{\v{c}}n{\'y},
  S.~Kumar, and H.~B. McMahan, ``Adaptive federated optimization,'' in
  \emph{9th International Conference on Learning Representations, {ICLR} 2021,
  Virtual Event, Austria, May 3-7, 2021}.\hskip 1em plus 0.5em minus
  0.4em\relax OpenReview.net, 2021.

\bibitem{local_adaAlter}
C.~Xie, O.~Koyejo, I.~Gupta, and H.~Lin, ``Local adaalter:
  Communication-efficient stochastic gradient descent with adaptive learning
  rates,'' \emph{CoRR}, vol. abs/1911.09030, 2019.

\bibitem{FedLamb}
B.~Karimi, X.~Li, and P.~Li, ``Fed-lamb: Layerwise and dimensionwise locally
  adaptive optimization algorithm,'' \emph{CoRR}, vol. abs/2110.00532, 2021.

\bibitem{SCAFFOLD}
S.~P. Karimireddy, S.~Kale, M.~Mohri, S.~J. Reddi, S.~U. Stich, and A.~T.
  Suresh, ``{SCAFFOLD:} stochastic controlled averaging for federated
  learning,'' in \emph{Proceedings of the 37th International Conference on
  Machine Learning, {ICML} 2020, 13-18 July 2020, Virtual Event}, ser.
  Proceedings of Machine Learning Research, vol. 119.\hskip 1em plus 0.5em
  minus 0.4em\relax {PMLR}, 2020, pp. 5132--5143.

\bibitem{quasi-momentum}
T.~Lin, S.~P. Karimireddy, S.~U. Stich, and M.~Jaggi, ``Quasi-global momentum:
  Accelerating decentralized deep learning on heterogeneous data,'' in
  \emph{Proceedings of the 38th International Conference on Machine Learning,
  {ICML} 2021, 18-24 July 2021, Virtual Event}, ser. Proceedings of Machine
  Learning Research, M.~Meila and T.~Zhang, Eds., vol. 139.\hskip 1em plus
  0.5em minus 0.4em\relax {PMLR}, 2021, pp. 6654--6665.

\bibitem{FedCM}
J.~Xu, S.~Wang, L.~Wang, and A.~C. Yao, ``Fedcm: Federated learning with
  client-level momentum,'' \emph{CoRR}, vol. abs/2106.10874, 2021.

\bibitem{favg_convergence1}
H.~Yang, M.~Fang, and J.~Liu, ``Achieving linear speedup with partial worker
  participation in non-iid federated learning,'' in \emph{9th International
  Conference on Learning Representations, {ICLR} 2021, Virtual Event, Austria,
  May 3-7, 2021}.\hskip 1em plus 0.5em minus 0.4em\relax OpenReview.net, 2021.

\bibitem{favg_convergence2}
X.~Li, K.~Huang, W.~Yang, S.~Wang, and Z.~Zhang, ``On the convergence of fedavg
  on non-iid data,'' in \emph{8th International Conference on Learning
  Representations, {ICLR} 2020, Addis Ababa, Ethiopia, April 26-30,
  2020}.\hskip 1em plus 0.5em minus 0.4em\relax OpenReview.net, 2020.

\bibitem{favg_convergence3}
T.~Lin, S.~U. Stich, K.~K. Patel, and M.~Jaggi, ``Don't use large mini-batches,
  use local {SGD},'' in \emph{8th International Conference on Learning
  Representations, {ICLR} 2020, Addis Ababa, Ethiopia, April 26-30,
  2020}.\hskip 1em plus 0.5em minus 0.4em\relax OpenReview.net, 2020.

\bibitem{FedProx}
T.~Li, A.~K. Sahu, M.~Zaheer, M.~Sanjabi, A.~Talwalkar, and V.~Smith,
  ``Federated optimization in heterogeneous networks,'' in \emph{Proceedings of
  Machine Learning and Systems 2020, MLSys 2020, Austin, TX, USA, March 2-4,
  2020}, I.~S. Dhillon, D.~S. Papailiopoulos, and V.~Sze, Eds.\hskip 1em plus
  0.5em minus 0.4em\relax mlsys.org, 2020.

\bibitem{SVRG}
R.~Johnson and T.~Zhang, ``Accelerating stochastic gradient descent using
  predictive variance reduction,'' in \emph{Advances in Neural Information
  Processing Systems 26: 27th Annual Conference on Neural Information
  Processing Systems 2013. Proceedings of a meeting held December 5-8, 2013,
  Lake Tahoe, Nevada, United States}, C.~J.~C. Burges, L.~Bottou,
  Z.~Ghahramani, and K.~Q. Weinberger, Eds., 2013, pp. 315--323.

\bibitem{FedNova}
J.~Wang, Q.~Liu, H.~Liang, G.~Joshi, and H.~V. Poor, ``Tackling the objective
  inconsistency problem in heterogeneous federated optimization,'' in
  \emph{Advances in Neural Information Processing Systems 33: Annual Conference
  on Neural Information Processing Systems 2020, NeurIPS 2020, December 6-12,
  2020, virtual}, H.~Larochelle, M.~Ranzato, R.~Hadsell, M.~Balcan, and H.~Lin,
  Eds., 2020.

\bibitem{FedPD}
X.~Zhang, M.~Hong, S.~V. Dhople, W.~Yin, and Y.~Liu, ``Fedpd: {A} federated
  learning framework with adaptivity to non-iid data,'' \emph{{IEEE} Trans.
  Signal Process.}, vol.~69, pp. 6055--6070, 2021.

\bibitem{local_momentum}
H.~Yu, R.~Jin, and S.~Yang, ``On the linear speedup analysis of communication
  efficient momentum {SGD} for distributed non-convex optimization,'' in
  \emph{Proceedings of the 36th International Conference on Machine Learning,
  {ICML} 2019, 9-15 June 2019, Long Beach, California, {USA}}, ser. Proceedings
  of Machine Learning Research, K.~Chaudhuri and R.~Salakhutdinov, Eds.,
  vol.~97.\hskip 1em plus 0.5em minus 0.4em\relax {PMLR}, 2019, pp. 7184--7193.

\bibitem{L2GD}
F.~Hanzely and P.~Richt{\'{a}}rik, ``Federated learning of a mixture of global
  and local models,'' \emph{CoRR}, vol. abs/2002.05516, 2020.

\bibitem{t1}
G.~Malinovskiy, D.~Kovalev, E.~Gasanov, L.~Condat, and P.~Richt{\'{a}}rik,
  ``From local {SGD} to local fixed-point methods for federated learning,'' in
  \emph{Proceedings of the 37th International Conference on Machine Learning,
  {ICML} 2020, 13-18 July 2020, Virtual Event}, ser. Proceedings of Machine
  Learning Research, vol. 119.\hskip 1em plus 0.5em minus 0.4em\relax {PMLR},
  2020, pp. 6692--6701.

\bibitem{t2}
K.~Ji, Z.~Wang, B.~Weng, Y.~Zhou, W.~Zhang, and Y.~Liang, ``History-gradient
  aided batch size adaptation for variance reduced algorithms,'' in
  \emph{Proceedings of the 37th International Conference on Machine Learning,
  {ICML} 2020, 13-18 July 2020, Virtual Event}, ser. Proceedings of Machine
  Learning Research, vol. 119.\hskip 1em plus 0.5em minus 0.4em\relax {PMLR},
  2020, pp. 4762--4772.

\bibitem{t4}
A.~Spiridonoff, A.~Olshevsky, and I.~C. Paschalidis, ``Local {SGD} with a
  communication overhead depending only on the number of workers,''
  \emph{CoRR}, vol. abs/2006.02582, 2020.

\bibitem{GD}
A.~Khaled, K.~Mishchenko, and P.~Richt{\'{a}}rik, ``First analysis of local
  {GD} on heterogeneous data,'' \emph{CoRR}, vol. abs/1909.04715, 2019.

\bibitem{appendix_1}
T.~Lin, S.~U. Stich, K.~K. Patel, and M.~Jaggi, ``Don't use large mini-batches,
  use local {SGD},'' in \emph{8th International Conference on Learning
  Representations, {ICLR} 2020, Addis Ababa, Ethiopia, April 26-30,
  2020}.\hskip 1em plus 0.5em minus 0.4em\relax OpenReview.net, 2020.

\bibitem{appendix_2}
B.~E. Woodworth, K.~K. Patel, S.~U. Stich, Z.~Dai, B.~Bullins, H.~B. McMahan,
  O.~Shamir, and N.~Srebro, ``Is local {SGD} better than minibatch sgd?'' in
  \emph{Proceedings of the 37th International Conference on Machine Learning,
  {ICML} 2020, 13-18 July 2020, Virtual Event}, ser. Proceedings of Machine
  Learning Research, vol. 119.\hskip 1em plus 0.5em minus 0.4em\relax {PMLR},
  2020, pp. 10\,334--10\,343.

\bibitem{yurochkin2019bayesian}
M.~Yurochkin, M.~Agarwal, S.~Ghosh, K.~Greenewald, N.~Hoang, and Y.~Khazaeni,
  ``Bayesian nonparametric federated learning of neural networks,'' in
  \emph{International conference on machine learning}.\hskip 1em plus 0.5em
  minus 0.4em\relax PMLR, 2019, pp. 7252--7261.

\bibitem{appendix_3}
S.~U. Stich, ``Local {SGD} converges fast and communicates little,'' in
  \emph{7th International Conference on Learning Representations, {ICLR} 2019,
  New Orleans, LA, USA, May 6-9, 2019}.\hskip 1em plus 0.5em minus 0.4em\relax
  OpenReview.net, 2019.

\bibitem{FedSplit}
R.~Pathak and M.~J. Wainwright, ``Fedsplit: an algorithmic framework for fast
  federated optimization,'' in \emph{Advances in Neural Information Processing
  Systems 33: Annual Conference on Neural Information Processing Systems 2020,
  NeurIPS 2020, December 6-12, 2020, virtual}, H.~Larochelle, M.~Ranzato,
  R.~Hadsell, M.~Balcan, and H.~Lin, Eds., 2020.

\bibitem{FedDR}
N.~H. Pham, L.~M. Nguyen, D.~T. Phan, and Q.~Tran{-}Dinh, ``Federated learning
  with randomized douglas-rachford splitting methods,'' \emph{CoRR}, vol.
  abs/2103.03452, 2021.

\bibitem{FedMAGA}
Z.~Hu, K.~Shaloudegi, G.~Zhang, and Y.~Yu, ``Fedmgda+: Federated learning meets
  multi-objective optimization,'' \emph{CoRR}, vol. abs/2006.11489, 2020.

\bibitem{FedPuring}
H.~Peng, J.~Wu, S.~Chen, and J.~Huang, ``Collaborative channel pruning for deep
  networks,'' in \emph{Proceedings of the 36th International Conference on
  Machine Learning, {ICML} 2019, 9-15 June 2019, Long Beach, California,
  {USA}}, ser. Proceedings of Machine Learning Research, K.~Chaudhuri and
  R.~Salakhutdinov, Eds., vol.~97.\hskip 1em plus 0.5em minus 0.4em\relax
  {PMLR}, 2019, pp. 5113--5122.

\bibitem{FedHN}
A.~Shamsian, A.~Navon, E.~Fetaya, and G.~Chechik, ``Personalized federated
  learning using hypernetworks,'' in \emph{Proceedings of the 38th
  International Conference on Machine Learning, {ICML} 2021, 18-24 July 2021,
  Virtual Event}, ser. Proceedings of Machine Learning Research, M.~Meila and
  T.~Zhang, Eds., vol. 139.\hskip 1em plus 0.5em minus 0.4em\relax {PMLR},
  2021, pp. 9489--9502.

\bibitem{t3}
H.~B. McMahan and M.~J. Streeter, ``Adaptive bound optimization for online
  convex optimization,'' in \emph{{COLT} 2010 - The 23rd Conference on Learning
  Theory, Haifa, Israel, June 27-29, 2010}, A.~T. Kalai and M.~Mohri,
  Eds.\hskip 1em plus 0.5em minus 0.4em\relax Omnipress, 2010, pp. 244--256.

\bibitem{appendix_4}
D.~Zhou, Y.~Tang, Z.~Yang, Y.~Cao, and Q.~Gu, ``On the convergence of adaptive
  gradient methods for nonconvex optimization,'' \emph{CoRR}, vol.
  abs/1808.05671, 2018.

\bibitem{distributed_adam}
T.~Chen, Z.~Guo, Y.~Sun, and W.~Yin, ``{CADA:} communication-adaptive
  distributed adam,'' in \emph{The 24th International Conference on Artificial
  Intelligence and Statistics, {AISTATS} 2021, April 13-15, 2021, Virtual
  Event}, ser. Proceedings of Machine Learning Research, A.~Banerjee and
  K.~Fukumizu, Eds., vol. 130.\hskip 1em plus 0.5em minus 0.4em\relax {PMLR},
  2021, pp. 613--621.

\bibitem{decentralized_adaptive}
Q.~Tong, G.~Liang, and J.~Bi, ``Effective federated adaptive gradient methods
  with non-iid decentralized data,'' \emph{CoRR}, vol. abs/2009.06557, 2020.

\bibitem{local_adaptive}
J.~Wang, Z.~Xu, Z.~Garrett, Z.~Charles, L.~Liu, and G.~Joshi, ``Local
  adaptivity in federated learning: Convergence and consistency,'' \emph{CoRR},
  vol. abs/2106.02305, 2021.

\bibitem{adam_rmsprop}
F.~Zou, L.~Shen, Z.~Jie, W.~Zhang, and W.~Liu, ``A sufficient condition for
  convergences of adam and rmsprop,'' in \emph{{IEEE} Conference on Computer
  Vision and Pattern Recognition, {CVPR} 2019, Long Beach, CA, USA, June 16-20,
  2019}.\hskip 1em plus 0.5em minus 0.4em\relax Computer Vision Foundation /
  {IEEE}, 2019, pp. 11\,127--11\,135.

\bibitem{quantized_adam}
C.~Chen, L.~Shen, H.~Huang, and W.~Liu, ``Quantized adam with error feedback,''
  \emph{{ACM} Trans. Intell. Syst. Technol.}, vol.~12, no.~5, pp. 56:1--56:26,
  2021.

\bibitem{adam_convergence_convexity}
C.~Chen, L.~Shen, F.~Zou, and W.~Liu, ``Towards practical adam: Non-convexity,
  convergence theory, and mini-batch acceleration,'' \emph{CoRR}, vol.
  abs/2101.05471, 2021.

\bibitem{yu2021fed2}
F.~Yu, W.~Zhang, Z.~Qin, Z.~Xu, D.~Wang, C.~Liu, Z.~Tian, and X.~Chen, ``Fed2:
  Feature-aligned federated learning,'' in \emph{Proceedings of the 27th ACM
  SIGKDD conference on knowledge discovery \& data mining}, 2021, pp.
  2066--2074.

\bibitem{li2022federated}
X.-C. Li, Y.-C. Xu, S.~Song, B.~Li, Y.~Li, Y.~Shao, and D.-C. Zhan, ``Federated
  learning with position-aware neurons,'' in \emph{Proceedings of the IEEE/CVF
  Conference on Computer Vision and Pattern Recognition}, 2022, pp.
  10\,082--10\,091.

\bibitem{FedNL}
M.~Safaryan, R.~Islamov, X.~Qian, and P.~Richt{\'{a}}rik, ``Fednl: Making
  newton-type methods applicable to federated learning,'' \emph{CoRR}, vol.
  abs/2106.02969, 2021.

\bibitem{hessian1}
R.~Islamov, X.~Qian, and P.~Richt{\'{a}}rik, ``Distributed second order methods
  with fast rates and compressed communication,'' in \emph{Proceedings of the
  38th International Conference on Machine Learning, {ICML} 2021, 18-24 July
  2021, Virtual Event}, ser. Proceedings of Machine Learning Research, M.~Meila
  and T.~Zhang, Eds., vol. 139.\hskip 1em plus 0.5em minus 0.4em\relax {PMLR},
  2021, pp. 4617--4628.

\bibitem{adagrad}
J.~C. Duchi, E.~Hazan, and Y.~Singer, ``Adaptive subgradient methods for online
  learning and stochastic optimization,'' \emph{J. Mach. Learn. Res.}, vol.~12,
  pp. 2121--2159, 2011.

\bibitem{adapt1}
H.~B. McMahan and M.~J. Streeter, ``Adaptive bound optimization for online
  convex optimization,'' in \emph{{COLT} 2010 - The 23rd Conference on Learning
  Theory, Haifa, Israel, June 27-29, 2010}, A.~T. Kalai and M.~Mohri,
  Eds.\hskip 1em plus 0.5em minus 0.4em\relax Omnipress, 2010, pp. 244--256.

\bibitem{adam}
D.~P. Kingma and J.~Ba, ``Adam: {A} method for stochastic optimization,'' in
  \emph{3rd International Conference on Learning Representations, {ICLR} 2015,
  San Diego, CA, USA, May 7-9, 2015, Conference Track Proceedings}, Y.~Bengio
  and Y.~LeCun, Eds., 2015.

\bibitem{adapt2}
X.~Li and F.~Orabona, ``On the convergence of stochastic gradient descent with
  adaptive stepsizes,'' in \emph{The 22nd International Conference on
  Artificial Intelligence and Statistics, {AISTATS} 2019, 16-18 April 2019,
  Naha, Okinawa, Japan}, ser. Proceedings of Machine Learning Research,
  K.~Chaudhuri and M.~Sugiyama, Eds., vol.~89.\hskip 1em plus 0.5em minus
  0.4em\relax {PMLR}, 2019, pp. 983--992.

\bibitem{adapt3}
X.~Wu, S.~S. Du, and R.~Ward, ``Global convergence of adaptive gradient methods
  for an over-parameterized neural network,'' \emph{CoRR}, vol. abs/1902.07111,
  2019.

\bibitem{adadelta}
M.~D. Zeiler, ``{ADADELTA:} an adaptive learning rate method,'' \emph{CoRR},
  vol. abs/1212.5701, 2012.

\bibitem{amsgrad}
S.~J. Reddi, S.~Kale, and S.~Kumar, ``On the convergence of adam and beyond,''
  in \emph{6th International Conference on Learning Representations, {ICLR}
  2018, Vancouver, BC, Canada, April 30 - May 3, 2018, Conference Track
  Proceedings}.\hskip 1em plus 0.5em minus 0.4em\relax OpenReview.net, 2018.

\bibitem{average_v}
X.~Chen, X.~Li, and P.~Li, ``Toward communication efficient adaptive gradient
  method,'' in \emph{{FODS} '20: {ACM-IMS} Foundations of Data Science
  Conference, Virtual Event, USA, October 19-20, 2020}, J.~M. Wing and
  D.~Madigan, Eds.\hskip 1em plus 0.5em minus 0.4em\relax {ACM}, 2020, pp.
  119--128.

\bibitem{local_preconditioner}
J.~Wang, Z.~Xu, Z.~Garrett, Z.~Charles, L.~Liu, and G.~Joshi, ``Local
  adaptivity in federated learning: Convergence and consistency,'' \emph{CoRR},
  vol. abs/2106.02305, 2021.

\bibitem{CIFAR100}
A.~Krizhevsky, G.~Hinton \emph{et~al.}, ``Learning multiple layers of features
  from tiny images,'' 2009.

\bibitem{Dirichlet}
T.~H. Hsu, H.~Qi, and M.~Brown, ``Measuring the effects of non-identical data
  distribution for federated visual classification,'' \emph{CoRR}, vol.
  abs/1909.06335, 2019.

\bibitem{resnet}
K.~He, X.~Zhang, S.~Ren, and J.~Sun, ``Deep residual learning for image
  recognition,'' in \emph{2016 {IEEE} Conference on Computer Vision and Pattern
  Recognition, {CVPR} 2016, Las Vegas, NV, USA, June 27-30, 2016}.\hskip 1em
  plus 0.5em minus 0.4em\relax {IEEE} Computer Society, 2016, pp. 770--778.

\bibitem{GN}
Y.~Wu and K.~He, ``Group normalization,'' \emph{Int. J. Comput. Vis.}, vol.
  128, no.~3, pp. 742--755, 2020.

\bibitem{hendrycks2016gaussian}
D.~Hendrycks and K.~Gimpel, ``Gaussian error linear units (gelus),''
  \emph{arXiv preprint arXiv:1606.08415}, 2016.

\bibitem{biswas2021smu}
K.~Biswas, S.~Kumar, S.~Banerjee, and A.~K. Pandey, ``Smu: smooth activation
  function for deep networks using smoothing maximum technique,'' \emph{arXiv
  preprint arXiv:2111.04682}, 2021.

\bibitem{FLGN}
K.~Hsieh, A.~Phanishayee, O.~Mutlu, and P.~B. Gibbons, ``The non-iid data
  quagmire of decentralized machine learning,'' in \emph{Proceedings of the
  37th International Conference on Machine Learning, {ICML} 2020, 13-18 July
  2020, Virtual Event}, ser. Proceedings of Machine Learning Research, vol.
  119.\hskip 1em plus 0.5em minus 0.4em\relax {PMLR}, 2020, pp. 4387--4398.

\end{thebibliography}

 \clearpage
\appendices
\onecolumn 
\section{More Experiments}
\subsection{Test of Communication Rounds}
\begin{table}[H]
\caption{Performance of FedAvg, FedCM, FedAdam, and FedLADA. The upper part is the number of communication rounds required for convergence and the test set accuracy at the corresponding round $t$. The lower part is the performance of training after 1500 rounds.}
\vskip 0.15in
\centering
\begin{tabular}{|c|cccccccccccc}
\hline
        & \multicolumn{3}{c|}{FedAvg}                                                   & \multicolumn{3}{c|}{FedCM}                                                    & \multicolumn{3}{c|}{FedAdam}                                                  & \multicolumn{3}{c|}{FedLADA}                                                  \\ \hline
Dataset & \multicolumn{1}{c|}{C10} & \multicolumn{1}{c|}{C100} & \multicolumn{1}{c|}{Ti} & \multicolumn{1}{c|}{C10} & \multicolumn{1}{c|}{C100} & \multicolumn{1}{c|}{Ti} & \multicolumn{1}{c|}{C10} & \multicolumn{1}{c|}{C100} & \multicolumn{1}{c|}{Ti} & \multicolumn{1}{c|}{C10} & \multicolumn{1}{c|}{C100} & \multicolumn{1}{c|}{Ti} \\ \hline
Acc.    & 81.8                     & 45.4                      & 34.5                   & 86.8                     & 52.5                      & 41.1                   & 86.4                     & 52.6                      & 41.3                   & 86.0                     & 51.4                      & \multicolumn{1}{c|}{41.9}                   \\ \cline{1-1}
Round   & 1511                     & 1664                      & 1474                   & 2089                     & 1698                      & 1754                   & 3031                     & 2324                      & 2834                   & 1313                     & 885                       & \multicolumn{1}{c|}{878}                    \\ \hline
Acc.    & 81.8                     & 45.3                      & 34.5                   & 84.9                     & 52.0                      & 40.6                   & 83.7                     & 51.3                      & 38.8                   & 86.1                     & 52.6                      & \multicolumn{1}{c|}{41.9}                  \\ \cline{1-1}
Round   & \multicolumn{12}{c|}{1500}                                                                                                                                                                                                                                                                                                     \\ \hline
\end{tabular}
\label{at1}
\end{table}
In Table~\ref{at1}, we compare the performance of FedAvg, FedCM, FedAdam and FedLADA on three datasets. $\textbf{C10}$ represents for CIFAR10, $\textbf{C100}$ represents for CIFAR100 and $\textbf{Ti}$ represents for the TinyImageNet. The top part records the communication rounds required for each algorithm to converge on the training set and the corresponding test set accuracy when the training set converges. The bottom part is the test accuracy after 1500 communication rounds. FedLADA performs excellently, which retains the high convergence speed of the local adaptive method while reducing the problem of heterogeneous over-fitting. FedLADA still has a good generalization guarantee after all the algorithms converge.
% Please add the following required packages to your document preamble:
% \usepackage{multirow}
\begin{table}[H]
\caption{The number of communication rounds required to achieve the specific accuracy on CIFAR10/100 and TinyImageNet dataset.}
\vskip 0.15in
\centering
\begin{tabular}{|c|c|cc|c|cc|c|cc|}
\hline
                 &                          & \multicolumn{2}{c|}{Rounds}                         &                           & \multicolumn{2}{c|}{Rounds}                         &                               & \multicolumn{2}{c|}{Rounds}                          \\ \hline
Target Acc.             & \multirow{8}{*}{CIFAR10} & \multicolumn{1}{c|}{70\%}          & 80\%           & \multirow{8}{*}{CIFAR100} & \multicolumn{1}{c|}{40\%}          & 45\%           & \multirow{8}{*}{TinyImageNet} & \multicolumn{1}{c|}{28\%}           & 35\%           \\ \cline{1-1} \cline{3-4} \cline{6-7} \cline{9-10} 
FedAvg           &                          & \multicolumn{1}{c|}{94.0}          & 536.9          &                           & \multicolumn{1}{c|}{299.1}         & 535.7          &                               & \multicolumn{1}{c|}{207.7}          & 421.1          \\ \cline{1-1} \cline{3-4} \cline{6-7} \cline{9-10} 
FedProx          &                          & \multicolumn{1}{c|}{90.9}          & 511.1          &                           & \multicolumn{1}{c|}{289.5}         & 501.6          &                               & \multicolumn{1}{c|}{214.6}          & 456.5          \\ \cline{1-1} \cline{3-4} \cline{6-7} \cline{9-10} 
SCAFFOLD         &                          & \multicolumn{1}{c|}{85.1}          & 262.5          &                           & \multicolumn{1}{c|}{135.7}         & 351.4          &                               & \multicolumn{1}{c|}{191.3}          & 362.2          \\ \cline{1-1} \cline{3-4} \cline{6-7} \cline{9-10} 
FedCM            &                          & \multicolumn{1}{c|}{71.5}          & 286.7          &                           & \multicolumn{1}{c|}{120.5}         & 200.0          &                               & \multicolumn{1}{c|}{115.9}          & 245.8          \\ \cline{1-1} \cline{3-4} \cline{6-7} \cline{9-10} 
FedAdam          &                          & \multicolumn{1}{c|}{96.1}          & 611.2            &                           & \multicolumn{1}{c|}{254.2}         & 407.4          &                               & \multicolumn{1}{c|}{242.1}          & 618.7          \\ \cline{1-1} \cline{3-4} \cline{6-7} \cline{9-10} 
LocalAdam        &                          & \multicolumn{1}{c|}{65.5}          & $\infty$            &                           & \multicolumn{1}{c|}{169.1}         & 361.9          &                               & \multicolumn{1}{c|}{142.4}          & $\infty$            \\ \cline{1-1} \cline{3-4} \cline{6-7} \cline{9-10} 
\textbf{FedLADA} &                          & \multicolumn{1}{c|}{\textbf{44.6}} & \textbf{186.4} &                           & \multicolumn{1}{c|}{\textbf{93.9}} & \textbf{143.7} &                               & \multicolumn{1}{c|}{\textbf{110.2}} & \textbf{170.2} \\ \hline
\end{tabular}
\label{at2}
\end{table}
Table~\ref{at2} shows the convergence speed of our proposed FedLADA and other baselines. $\infty$ represents impossible accuracy in the whole training rounds. All numbers in the table are averaged by 5 results with different random seeds. The target accuracy given in the table is for the early and middle stages of training, which indicates that our proposed FedLADA has a very efficient convergence speed on both three datasets. In the early stage, FedLADA is about 1.3$\times$ faster than the second in average and 1.5$\times$ faster in the middle stage.

\subsection{Test of Smooth Function}
We use the LeNet as the model to test the selected baselines and our proposed FedLADA. LeNet is a simple model consisting of 2 convolutional layers and 2 full-connection layers. There is an activation function in front of each layer. The vanilla activation is ReLU. We select another 2 smooth activation functions instead of ReLU to satisfy the smoothness. The test is shown as follows. Both of them could be considered as the smooth approximation of ReLU.\\

(a) Smooth Activation Functions
\begin{table}[H]
%\small % if necessary
  \caption{Smooth activation functions.}
  \vspace{-0.2cm}
  \label{activation}
  \centering
  \begin{threeparttable}
  \begin{tabular}{cc}
    \toprule
    Method     & Formulation   \\
    \midrule
    GeLU~\cite{hendrycks2016gaussian} & $\frac{w}{2}\left(1+\tanh\left(\sqrt{\frac{2}{\pi}}\left(w+0.044715w^3\right)\right)\right)$\\
    SMU~\cite{biswas2021smu} & $\frac{1}{2}\left(\mathbf{x} + \mathbf{x} \cdot \text{erf}(\mu\mathbf{x})\right)$ \\
    \bottomrule
  \end{tabular}
  \vskip 0.2in
  \end{threeparttable}
  %\vspace{-0.5cm}
\end{table}
GeLU is a classical smooth approximation of ReLU. SMU is a novel smooth approximation that selects a proper $\mu$ to approximate the ReLU activation. We use these two smooth activation functions in the LeNet to test our experiments.\\

(b) Experiments\\
\begin{figure}[h]
	\centering
	\begin{minipage}[b]{0.49\textwidth}
		\subfigure[Accuracy and loss of LeNet~(GeLU).]{
			\includegraphics[width=0.49\textwidth]{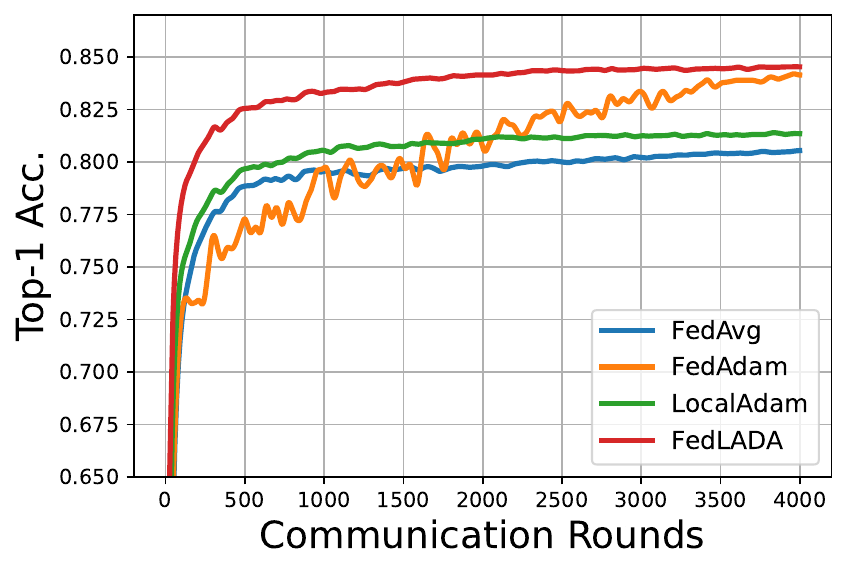} 
			\includegraphics[width=0.49\textwidth]{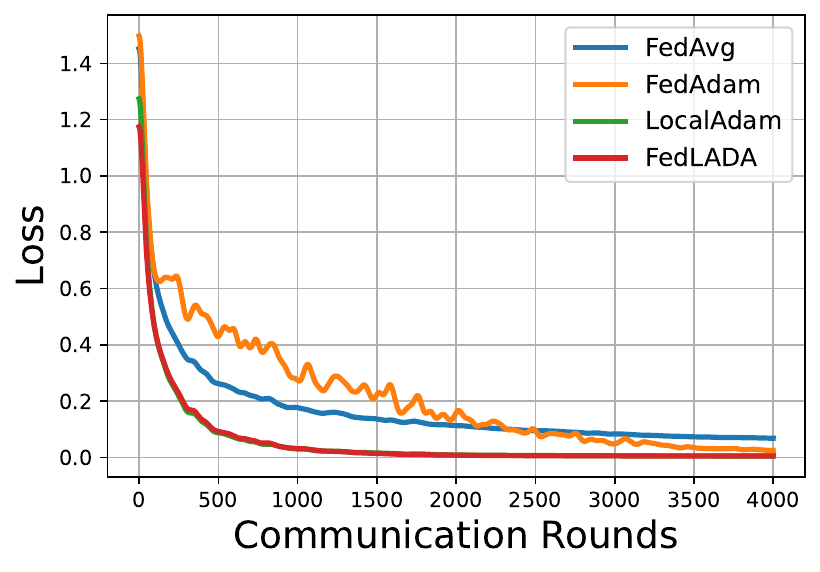}
			\label{fig:hor_4figs_1cap_2subcap_1}
	}
	\end{minipage}
	\begin{minipage}[b]{0.49\textwidth}
		\subfigure[Accuracy and loss of LeNet~(SMU).]{
			\includegraphics[width=0.49\textwidth]{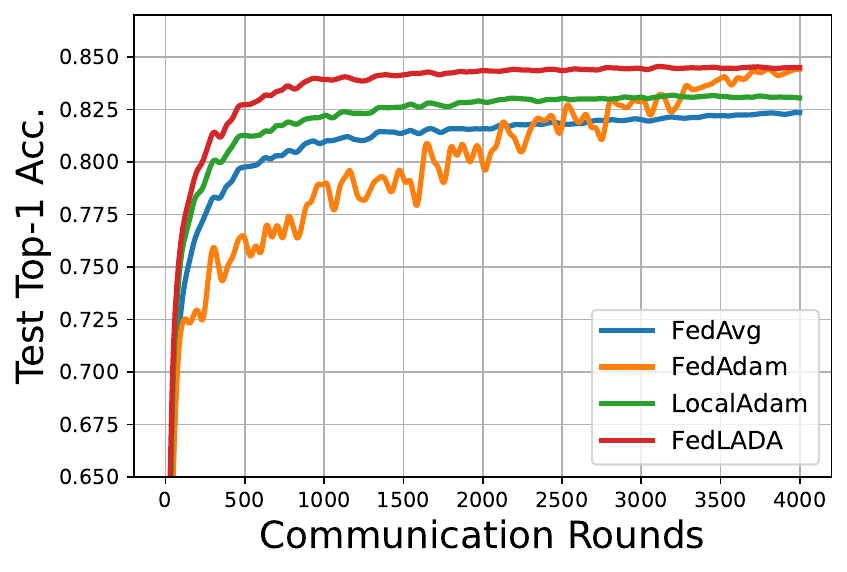} 
			\includegraphics[width=0.49\textwidth]{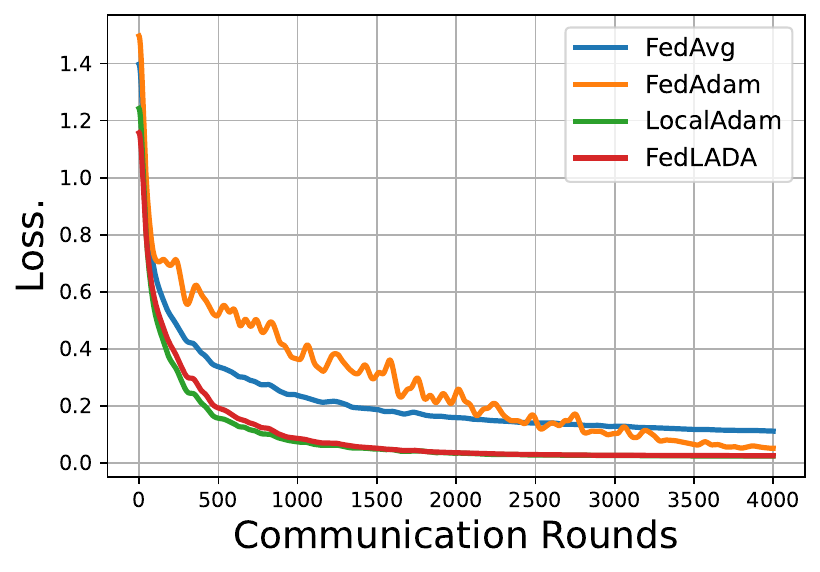}
			\label{fig:hor_4figs_1cap_2subcap_2}
	}
	\end{minipage}
	\caption{We use the simple LeNet with the smooth activation to test on the CIFAR-10 dataset Dirichlet-0.3 splitting. We test different hyperparameters and report the best results here. The total communication round is set as $T=4000$.  }
	\label{fig:hor_4figs_1cap_2subcap}
\end{figure}

In the Fig~\ref{fig:hor_4figs_1cap_2subcap}, we clearly see the rugged convergence of adopting the global adaptive optimizer in FL. This is consistent with what we observed with adopting the ReLU activation. In the loss figure, we can see that the global adaptive converges very slowly. While LocalAdam achieves much faster convergence than the FedAvg. Our proposed FedLADA achieves the best results both on the test accuracy and the training loss.

Though the ReLU is non-smooth activation, the test on the GeLU and SMU shows a similar phenomenon in our paper. In the deep models, which is a general problem in the FL paradigm. We also explain it in section 5 and discuss it in detail. Our proposed FedLADA tries to adopt the global correction to enhance local consistency, which helps to achieve a better convergence speed and higher test accuracy. Actually, in our experiments, the other advanced methods, i.e. SCAFFOLD~\cite{SCAFFOLD} and FedCM~\cite{FedCM}, may diverge in training and become very unstable and sensitive to the selection of the hyperparameters. The local adaptive method is very stable to the hyperparameters. We almost do not change the selections as introduced in our paper.

\section{Proofs}
\label{proof}
\allowdisplaybreaks[3]
\renewcommand{\qedsymbol}{}
In this part, we introduce the proof of the main Theorem~\ref{convergence_rate}. \cite{local_momentum} provide proof for local momentum to combine local intervals and communication rounds as total iterations and decompose momentum terms as recursion. We use the proof framework of FedAvg \cite{favg_convergence1} to separate the local and global training processes. We refer to the proof of FedDyn \cite{FedDyn} to bound the sum of the norm of parameters offsets and momentum offsets instead of them separately by constructing a sum sequence. We propose to bound the sequence $\{\epsilon_{\tau}^{t}+q\mathbf{S}_{\tau}^{t}\}$, where $\epsilon_{k}^{t}=\sum_{i,\tau}\mathbb{E}\Vert \mathbf{x}_{i,\tau}^{t}-\mathbf{x}^{t}\Vert^{2}$ and $\mathbf{S}_{k}^{t}=\sum_{i,\tau}\mathbb{E}\Vert \mathbf{m}_{i,\tau}^{t}\odot\vartheta_{i,\tau}^{t}\Vert^{2}$ and $q$ is positive constant. We also apply the growth boundedness of the upper bound of the second-order momenta term $\hat{\mathbf{v}}$ in AMSgrad to bound the difference term for the second-order momenta term.
\subsection{Preliminary lemmas}
Firstly we introduce some common lemmas used in our proof.
\begin{lemma}\label{lemma_product}
    For $\forall \ \mathbf{x}, \mathbf{y} \in \mathbb{R}^{d}$,   
    \begin{equation}
        \langle \mathbf{x} , \mathbf{y} \rangle = \frac{1}{2} (\Vert \mathbf{x} \Vert^{2} + \Vert \mathbf{y} \Vert^{2} - \Vert \mathbf{x} - \mathbf{y} \Vert^{2} ).
    \end{equation}
\end{lemma}
\begin{lemma}\label{lemma_hadamard_norm}
    For $\forall \ \mathbf{x}, \mathbf{y} \in \mathbb{R}^{d}$,   
    \begin{equation}
        \Vert\mathbf{x}\odot\mathbf{y}\Vert^{2} \leq \Vert\mathbf{x}\Vert^{2}\Vert\mathbf{y}\Vert_{\infty}^{2}.
    \end{equation}
\end{lemma}
\begin{proof}
    Let $\mathbf{x}=[x_{1}, x_{2}, \cdots, x_{d}]$, $\mathbf{y}=[y_{1}, y_{2}, \cdots, y_{d}]$, then
    \begin{align*}
        \Vert\mathbf{x}\odot\mathbf{y}\Vert^{2}
        &= \sum_{j=1}^{d}(x_{j}y_{j})^{2}
        \leq \sum_{j=1}^{d}x_{j}^{2}\cdot \max_{i}(y_{i}^{2})=\Vert\mathbf{x}\Vert^{2}\Vert\mathbf{y}\Vert_{\infty}^{2}.\\
    \end{align*}
\end{proof}
\begin{lemma}\label{lemma_hadamard_largernorm}
    For $\forall \ \mathbf{x}, \mathbf{y}, \mathbf{z} \in \mathbb{R}^{d}$, if \ $\forall \ \mathbf{z}_{j} \geq \mathbf{y}_{j} \geq 0 \ for \ j \in \{1, 2, \cdots, d\}$,
    \begin{equation}
        \Vert\mathbf{x}\odot\mathbf{y}\Vert^{2} \leq \Vert\mathbf{x}\odot\mathbf{z}\Vert^{2}.
    \end{equation}
\end{lemma}
\begin{proof}
    Let $\mathbf{x}=[x_{1}, x_{2}, \cdots, x_{d}]$, $\mathbf{y}=[y_{1}, y_{2}, \cdots, y_{d}]$, $\mathbf{z}=[z_{1}, z_{2}, \cdots, z_{d}]$ then
    \begin{align*}
        \Vert\mathbf{x}\odot\mathbf{y}\Vert^{2}
        &= \sum_{j=1}^{d}(x_{j}y_{j})^{2}
        \leq \sum_{j=1}^{d}x_{j}^{2}\cdot z_{j}^{2} = \Vert\mathbf{x}\odot\mathbf{z}\Vert^{2}.\\
    \end{align*}
\end{proof}
\begin{lemma}\label{lemma_youngs}
    For $\forall \ \mathbf{x}, \mathbf{y}\in \mathbb{R}^{d}$ and $\mu \ > \ 0$,
    \begin{equation}
        \Vert\mathbf{x} + \mathbf{y}\Vert^{2} \leq (1+\mu)\Vert\mathbf{x}\Vert^{2} + (1+\frac{1}{\mu})\Vert\mathbf{y}\Vert^{2}.
    \end{equation}
\end{lemma}
\begin{proof}
    Let $\mathbf{x}=[x_{1}, x_{2}, \cdots, x_{d}]$, $\mathbf{y}=[y_{1}, y_{2}, \cdots, y_{d}]$, then
    \begin{align*}
        (1+\mu)\Vert\mathbf{x}\Vert^{2} + (1+\frac{1}{\mu})\Vert\mathbf{y}\Vert^{2}
        &= \sum_{j=1}^{d}(1+\mu)x_{j}^{2} + (1+\frac{1}{\mu})y_{j}^{2}\\
        &= \sum_{j=1}^{d}x_{j}^{2}+y_{j}^{2}+\mu x_{j}^{2} + \frac{1}{u}y_{j}^{2}\\
        &\geq \sum_{j=1}^{d}x_{j}^{2}+y_{j}^{2}+ 2x_{j}y_{j}= \Vert\mathbf{x} + \mathbf{y}\Vert^{2}.\\
    \end{align*}
\end{proof}
\begin{lemma}\label{lemma_cauchy}
    For $\forall \ \mathbf{x}_{1}, \mathbf{x}_{2}. \cdots, \mathbf{x}_{n}\in \mathbb{R}^{d}$,
    \begin{equation}
        \Vert\sum_{i=1}^{n}\mathbf{x}_{i}\Vert^{2} \leq n\sum_{i=1}^{n}\Vert\mathbf{x}\Vert^{2}.
    \end{equation}
\end{lemma}
\begin{proof}
This inequality can be derived from the Cauchy–Schwarz inequality.
\end{proof}
\begin{lemma}\label{mean_norm}
    For $\forall \ \mathbf{x}_{1}, \mathbf{x}_{2}. \cdots, \mathbf{x}_{n}\in \mathbb{R}^{d}$ are random variables. Suppose that \{$\mathbf{x}_{i}$\} is a zero mean sequence, then:
    \begin{equation}
        \mathbb{E}\Vert\sum_{i=1}^{n}\mathbf{x}_{i}\Vert^{2}=\mathbb{E}[\sum_{i=1}^{n}\Vert \mathbf{x}_{i}\Vert^{2}].
    \end{equation}
\end{lemma}
\begin{proof}
    Let $\mathbf{x}_{(j)}$ represents for the $j$-th element of $\mathbf{x}$.
    \begin{align*}
        \mathbb{E}\Vert\sum_{i=1}^{n}\mathbf{x}_{i}\Vert^{2}
        &= \mathbb{E}[\sum_{j}^{d}\big(\sum_{i}^{n}(\mathbf{x}_{i})_{(j)}\big)^{2}]\\
        &= \mathbb{E}[\sum_{j}^{d}\big(\sum_{i}^{n}(\mathbf{x}_{i})_{(j)}^{2} + \sum_{i\neq k}^{n}(\mathbf{x}_{i})_{(j)}(\mathbf{x}_{k})_{(j)}\big)]\\
        &= \mathbb{E}[\sum_{i=1}^{n}\Vert \mathbf{x}_{i}\Vert^{2}] + \sum_{j}^{d}\sum_{i\neq k}^{n}\mathbb{E}(\mathbf{x}_{i})_{(j)}\mathbb{E}(\mathbf{x}_{k})_{(j)}=\mathbb{E}[\sum_{i=1}^{n}\Vert \mathbf{x}_{i}\Vert^{2}].\\
    \end{align*}
\end{proof}
\begin{lemma}\label{lemma_tpt}
    For $\forall \ 0 \ < \ m \ < \ 1$,
    \begin{equation}
        \sum_{\tau=1}^{K}\tau m^{\tau} = \frac{m}{(1-m)^{2}} - \frac{m^{K+1}}{(1-m)^{2}}- \frac{Km^{K+1}}{1-m}.
    \end{equation}
\end{lemma}
\begin{proof}
    Let $S = \sum_{\tau=1}^{K}\tau m^{\tau}$, we have $mS = \sum_{\tau=1}^{K}\tau m^{\tau+1}$,
    \begin{align*}
        S = \frac{S - mS}{1-m} 
        &= \frac{1}{1-m} \Big(\sum_{\tau=1}^{K}\tau m^{\tau} - \sum_{\tau=1}^{K}\tau m^{\tau+1}\Big)\\
        &= \frac{1}{1-m} \Big(\sum_{\tau=1}^{K} m^{\tau} - Km^{K+1}\Big)\\
        &= \frac{1}{1-m} \Big(\frac{m(1-m^{K})}{1-m} - Km^{K+1}\Big)\\
        &= \frac{m}{(1-m)^{2}} - \frac{m^{K+1}}{(1-m)^{2}}- \frac{Km^{K+1}}{1-m}.\\
    \end{align*}
\end{proof}
\subsection{Proof of the Theorem \ref{convergence_rate}}
Then we introduce an additional variable $\mathbf{z}$ to facilitate the proof:
\begin{equation}
    \mathbf{z}^{t+1} = \frac{1}{\alpha}\mathbf{x}^{t+1} - \frac{1-\alpha}{\alpha}\mathbf{x}^{t}.
\end{equation}
Then $\mathbf{z}_{t}$ satisfies the update of:
\begin{lemma}\label{lemma_update}
    \begin{equation}
        \mathbf{z}^{t+1} = \mathbf{z}^{t} - \eta\eta_{l}\frac{1}{SK}\sum_{i\in \mathcal{S}^{t}}\sum_{\tau=1}^{K}\mathbf{m}_{i,\tau}^{t}\odot\vartheta_{i,\tau}^{t}.
    \end{equation}
\end{lemma}
\begin{proof}
    \begin{align*}
        \mathbf{z}^{t+1}
        &= \frac{1}{\alpha}\mathbf{x}^{t+1} - \frac{1-\alpha}{\alpha}\mathbf{x}^{t}\\
        &= \frac{1}{\alpha}\big(\mathbf{x}^{t} - \eta_{g}\frac{1}{S}\sum_{i\in\mathcal{S}^{t}}(\mathbf{x}_{i,0}^{t} - \mathbf{x}_{i,K}^{t})\big) -  \frac{1-\alpha}{\alpha}\mathbf{x}^{t}\\
        &= \frac{1}{\alpha}\mathbf{x}^{t} - \frac{1-\alpha}{\alpha}\mathbf{x}^{t-1} + \frac{1-\alpha}{\alpha}\mathbf{x}^{t-1} -\eta_{g}\frac{1}{\alpha S}\sum_{i\in\mathcal{S}^{t}}(\mathbf{x}_{i,0}^{t} - \mathbf{x}_{i,K}^{t}) -  \frac{1-\alpha}{\alpha}\mathbf{x}^{t}\\
        &= \mathbf{z}^{t} - \eta_{g}\frac{1}{\alpha S}\sum_{i\in\mathcal{S}^{t}}(\mathbf{x}_{i,0}^{t} - \mathbf{x}_{i,K}^{t}) - \frac{1-\alpha}{\alpha}(\mathbf{x}_{t} - \mathbf{x}_{t-1})\\
        &= \mathbf{z}^{t} - \eta_{g}\frac{1}{\alpha S}\sum_{i\in\mathcal{S}^{t}}\sum_{\tau=1}^{K}\eta_{l}(\alpha\mathbf{m}_{i,\tau}^{t}\odot\vartheta_{i,\tau}^{t}+(1-\alpha)\mathbf{g}_{a}^{t}) - \frac{1-\alpha}{\alpha}(-\eta_{g}\eta_{l}K\mathbf{g}_{a}^{t})\\
        &= \mathbf{z}^{t} - \eta_{g}\eta_{l}\frac{1}{S}\sum_{i\in\mathcal{S}^{t}}\sum_{\tau=1}^{K}\mathbf{m}_{i,\tau}^{t}\odot\vartheta_{i,\tau}^{t} + \Big(\frac{1-\alpha}{\alpha}\eta_{g}\eta_{l}K\mathbf{g}_{a}^{t}- \eta_{g}\eta_{l}\frac{1-\alpha}{\alpha}K\mathbf{g}_{a}^{t}\Big)\\
        &= \mathbf{z}^{t} - \eta\eta_{l}\frac{1}{SK}\sum_{i\in\mathcal{S}^{t}}\sum_{\tau=1}^{K}\mathbf{m}_{i,\tau}^{t}\odot\vartheta_{i,\tau}^{t}.\\
    \end{align*}
\end{proof}
Here it satisfies $\eta=K\eta_{g}$ to ensure it can still maintain a change equivalent to $K$ steps on the global server after each local client optimizes $K$ iterations. Our purpose in introducing the variable $\mathbf{z}$ is to reduce some redundant terms in the following proof process. In our works, we execute the local adaptive optimizer and use the global adaptive gradient estimation to 
reduce the local gradient error introduced by the heterogeneity dataset. According to the update of $\mathbf{z}$, we give the following important theorems.\\\\
We denote $\mathbb{E}_{\tau|t}$ as the conditional expectation of iterations from $1$ to $\tau$ in the communication round $t$ given the rounds 0, 1, $\cdots$, $t-1$. According to the Assumption \ref{smoothness}, we have:
\begin{align*}
        \mathbb{E}_{K|t}[F(\mathbf{z}^{t+1})] 
        &\overset{(a)}{\leq} \mathbb{E}_{K|t}[F(\mathbf{z}^{t})] + \mathbb{E}_{K|t}\langle\nabla F(\mathbf{z}^{t}), {\mathbf{z}^{t+1}-\mathbf{z}^{t}}\rangle + \frac{L}{2}\mathbb{E}_{K|t}\Vert\mathbf{z}^{t+1}-\mathbf{z}^{t}\Vert^{2}\\
        &= F(\mathbf{z}^{t}) + \mathbb{E}_{K|t}\langle\nabla F(\mathbf{z}^{t}), \underbrace{- \eta\eta_{l}\frac{1}{SK}\sum_{i\in\mathcal{S}^{t}}\sum_{\tau=1}^{K}\mathbf{m}_{i,\tau}^{t}\odot\vartheta_{i,\tau}^{t}}_{\textbf{partial participation}}\rangle + \frac{L}{2}\mathbb{E}_{K|t}\Vert\mathbf{z}^{t+1}-\mathbf{z}^{t}\Vert^{2}\\
        &\overset{(b)}{=} F(\mathbf{z}^{t}) + \mathbb{E}_{K|t}\langle\nabla F(\mathbf{z}^{t}), - \eta\eta_{l}\frac{1}{mK}\sum_{i}\sum_{\tau=1}^{K}\mathbf{m}_{i,\tau}^{t}\odot\vartheta_{i,\tau}^{t}\rangle + \frac{L}{2}\mathbb{E}_{K|t}\Vert\mathbf{z}^{t+1}-\mathbf{z}^{t}\Vert^{2}\\
        &= F(\mathbf{z}^{t}) + \frac{L}{2}\underbrace{\mathbb{E}_{K|t}\Vert\mathbf{z}^{t+1}-\mathbf{z}^{t}\Vert^{2}}_{\mathbf{R}_{1}} + \underbrace{\mathbb{E}_{K|t}\langle\nabla F(\mathbf{z}^{t}), - \eta\eta_{l}\frac{1}{mK}\sum_{i}\sum_{\tau=1}^{K}\mathbf{m}_{i,\tau}^{t}\odot\vartheta_{i,0}^{t}\rangle}_{\mathbf{R}_{2}}\\
        &\quad + \underbrace{\mathbb{E}_{K|t}\langle\nabla F(\mathbf{z}^{t}), - \eta\eta_{l}\frac{1}{mK}\sum_{i}\sum_{\tau=1}^{K}\mathbf{m}_{i,\tau}^{t}\odot\big(\vartheta_{i,\tau}^{t}-\vartheta_{i,0}^{t}\big)\rangle}_{\mathbf{R}_{3}}.\\
\end{align*}
$(a)$ applies the $L$-smooth assumption of the function $F$; $(b)$ takes the expectation on clients sampling with the same probability. Then we bound the $\mathbf{R}_{1}$, $\mathbf{R}_{2}$ and $\mathbf{R}_{3}$ respectively.
\subsubsection{Bounded R1}
Firstly we bound the term $\mathbf{R}_{1}$. We define a indicator function $\mathbb{I}_{i}^{t}$ for client $i$ is selected to be active at round $t$,
\begin{align*}
    \mathbb{I}_{i}^{t}=
    \begin{cases}
        0& \text{client $i$ is inactive at round $t$}\\
        1& \text{client $i$ is active at round $t$}\\
\end{cases}.
\end{align*}
Partial participation means that each client is selected with the same probability at each round $t$. Thus we have probability of $P(\mathbb{I}_{i}^{t}=1)=\frac{S}{m}$ for $\forall i \in [m]$. According to the lemma \ref{lemma_update},
\begin{align*}
    \mathbf{R}_{1}
    &= \mathbb{E}_{K|t}\Vert\mathbf{z}^{t+1}-\mathbf{z}^{t}\Vert^{2}\\
    &= \mathbb{E}_{K|t}\Vert - \eta\eta_{l}\frac{1}{SK}\sum_{i\in \mathcal{S}^{t}}\sum_{\tau=1}^{K}\mathbf{m}_{i,\tau}^{t}\odot\vartheta_{i,\tau}^{t} \Vert^{2}\\
    &= \mathbb{E}_{K|t}\Vert\eta\eta_{l}\frac{1}{SK}\sum_{i\in \mathcal{S}^{t}}\sum_{\tau=1}^{K}\mathbf{m}_{i,\tau}^{t}\odot(\vartheta_{i,\tau}^{t}-\vartheta_{i,0}^{t}+\vartheta_{i,0}^{t}) \Vert^{2}\\
    &\overset{(a)}{\leq} 2\eta^{2}\eta_{l}^{2}\underbrace{\mathbb{E}_{K|t}\Vert\frac{1}{SK}\sum_{i\in \mathcal{S}^{t}}\sum_{\tau=1}^{K}\mathbf{m}_{i,\tau}^{t}\odot\vartheta_{i,0}^{t} \Vert^{2}}_{\mathbf{M}^{t}} + 2\eta^{2}\eta_{l}^{2}\underbrace{\mathbb{E}_{K|t}\Vert\frac{1}{SK}\sum_{i\in \mathcal{S}^{t}}\sum_{\tau=1}^{K}\mathbf{m}_{i,\tau}^{t}\odot(\vartheta_{i,\tau}^{t}-\vartheta_{i,0}^{t}) \Vert^{2}}_{\Delta^{t}}.\\
\end{align*}
(a) applies the lemma \ref{lemma_cauchy}.
Then we can bound $\mathbf{M}^{t}$ as follows. By defining the $\overline{\mathbf{m}}_{i,\tau}^t = (1-\beta_1)\sum_{j=1}^{\tau}\beta_1^{\tau-j+1}\nabla F_i(\mathbf{x}_{i,\tau}^t)$, applying the triangle inequality and the zero-mean expectation $\mathbb{E}\Vert \sum_i x_i\Vert^2=\sum_i \mathbb{E}\Vert x_i\Vert^2$ if $\mathbb{E}\left[x_i\vert x_{i-1},x_{i-2},\cdots,x_{1}\right] = 0$, we have:
\begin{align*}
    \mathbf{M}^{t}
    &= \mathbb{E}_{K|t}\Vert\frac{1}{SK}\sum_{i,\tau} \mathbf{m}_{i,\tau}^{t}\odot\vartheta_{i,0}^{t}\Vert^{2}\\
    &\leq 2\mathbb{E}_{K|t}\Vert\frac{1}{SK}\sum_{i,\tau} (\mathbf{m}_{i,\tau}^{t}-\overline{\mathbf{m}}_{i,\tau}^{t})\odot\vartheta_{i,0}^{t}\Vert^{2} + 2\mathbb{E}_{K|t}\Vert\frac{1}{SK}\sum_{i,\tau} \overline{\mathbf{m}}_{i,\tau}^{t}\odot\vartheta_{i,0}^{t}\Vert^{2}\\
    &= 2\mathbb{E}_{K|t}\Vert\frac{1}{SK}\sum_{i,\tau} \sum_{j=1}^{\tau}(1-\beta_{1})\beta_{1}^{\tau-j+1}\big(\mathbf{g}_{i,j}^{t}-\nabla F_{i}(\mathbf{x}_{i,j}^{t})\big)\odot\vartheta_{i,0}^{t}\Vert^{2} + 2\mathbb{E}_{K|t}\Vert\frac{1}{SK}\sum_{i,\tau} \overline{\mathbf{m}}_{i,\tau}^{t}\odot\vartheta_{i,0}^{t}\Vert^{2}\\
    &= 2\mathbb{E}_{K|t}\Vert\frac{1}{SK}\sum_{i,\tau} (\sum_{j=\tau}^{K}(1-\beta_{1})\beta_{1}^{K-j+1})\big(\mathbf{g}_{i,\tau}^{t}-\nabla F_{i}(\mathbf{x}_{i,\tau}^{t})\big)\odot\vartheta_{i,0}^{t}\Vert^{2} + 2\mathbb{E}_{K|t}\Vert\frac{1}{SK}\sum_{i,\tau} \overline{\mathbf{m}}_{i,\tau}^{t}\odot\vartheta_{i,0}^{t}\Vert^{2}\\
    &= 2\mathbb{E}_{K|t}\Vert\frac{1}{SK}\sum_{i,\tau} \beta_{1}(1-\beta_{1}^{\tau})\big(\mathbf{g}_{i,\tau}^{t}-\nabla F_{i}(\mathbf{x}_{i,\tau}^{t})\big)\odot\vartheta_{i,0}^{t}\Vert^{2} + 2\mathbb{E}_{K|t}\Vert\frac{1}{SK}\sum_{i,\tau} \overline{\mathbf{m}}_{i,\tau}^{t}\odot\vartheta_{i,0}^{t}\Vert^{2}\\
    &= \frac{2\beta_{1}^{2}G_{\vartheta}^{2}}{S^{2}K^{2}}\mathbb{E}_{K|t}\Vert\sum_{i,\tau} (1-\beta_{1}^{\tau})\big(\mathbf{g}_{i,\tau}^{t}-\nabla F_{i}(\mathbf{x}_{i,\tau}^{t})\big)\Vert^{2} + 2\mathbb{E}_{K|t}\Vert\frac{1}{SK}\sum_{i,\tau} \overline{\mathbf{m}}_{i,\tau}^{t}\odot\vartheta_{i,0}^{t}\Vert^{2}\\
    &= \frac{2\beta_{1}^{2}G_{\vartheta}^{2}}{S^{2}K^{2}}\sum_{i,\tau}\mathbb{E}_{K|t}\Vert (1-\beta_{1}^{\tau})\big(\mathbf{g}_{i,\tau}^{t}-\nabla F_{i}(\mathbf{x}_{i,\tau}^{t})\big)\Vert^{2} + 2\mathbb{E}_{K|t}\Vert\frac{1}{SK}\sum_{i,\tau} \overline{\mathbf{m}}_{i,\tau}^{t}\odot\vartheta_{i,0}^{t}\Vert^{2}\\
    &= \frac{2\beta_{1}^{2}G_{\vartheta}^{2}}{S^{2}K^{2}}\sum_{i,\tau}(1-\beta_{1}^{\tau})^{2}\mathbb{E}_{K|t}\Vert \mathbf{g}_{i,\tau}^{t}-\nabla F_{i}(\mathbf{x}_{i,\tau}^{t})\Vert^{2} + 2\mathbb{E}_{K|t}\Vert\frac{1}{SK}\sum_{i,\tau} \overline{\mathbf{m}}_{i,\tau}^{t}\odot\vartheta_{i,0}^{t}\Vert^{2}\\
    &\overset{(a)}{\leq} \frac{2\beta_{1}^{2}G_{\vartheta}^{2}\sigma_{l}^{2}}{S^{2}K^{2}}\sum_{i,\tau}(1-\beta_{1}^{\tau})^{2} + 2\mathbb{E}_{K|t}\Vert\frac{1}{SK}\sum_{i,\tau} \overline{\mathbf{m}}_{i,\tau}^{t}\odot\vartheta_{i,0}^{t}\Vert^{2}\\
    &\leq \frac{2\beta_{1}^{2}G_{\vartheta}^{2}\sigma_{l}^{2}}{SK} + 2\underbrace{\mathbb{E}_{K|t}\Vert\frac{1}{SK}\sum_{i,\tau} \overline{\mathbf{m}}_{i,\tau}^{t}\odot\vartheta_{i,0}^{t}\Vert^{2}}_{\mathbf{W}^{t}}.\\
\end{align*}
$(a)$ applies the assumption of bounded gradient $\Vert\mathbf{g}_{i,\tau}^{t} - \nabla F(\mathbf{x}_{i,\tau}^{t}) \Vert^{2} \leq \sigma_{l}^{2}$. Then We bound $\mathbf{W}^{t}$ as:
\begin{align*}
    \mathbf{W}^{t}
    &= \mathbb{E}_{K|t}\Vert\frac{1}{SK}\sum_{i\in \mathcal{S}^{t}}\sum_{\tau=1}^{K}\overline{\mathbf{m}}_{i,\tau}^{t}\odot\vartheta_{i,0}^{t} \Vert^{2}\\
    &= \frac{1}{S^{2}K^{2}}\mathbb{E}_{K|t} \langle \sum_{i\in \mathcal{S}^{t}}\sum_{\tau=1}^{K}\overline{\mathbf{m}}_{i,\tau}^{t}\odot\vartheta_{i,0}^{t}\mathbb{I}_{i}^{t}, \sum_{j\in \mathcal{S}^{t}}\sum_{\tau=1}^{K}\overline{\mathbf{m}}_{j,\tau}^{t}\odot\vartheta_{i,0}^{t}\mathbb{I}_{j}^{t}\rangle\\
    &= \frac{1}{S^{2}K^{2}}\mathbb{E}_{K|t}\Big( \sum_{i\neq j}\langle\sum_{\tau=1}^{K} \overline{\mathbf{m}}_{i,\tau}^{t}\odot\vartheta_{i,0}^{t}, \sum_{\tau=1}^{K}\overline{\mathbf{m}}_{j,\tau}^{t}\odot\vartheta_{i,0}^{t}\rangle\mathbb{E}[\mathbb{I}_{i}^{t}\mathbb{I}_{j}^{t}] + \sum_{i=j}\langle \sum_{\tau=1}^{K}\overline{\mathbf{m}}_{i,\tau}^{t}\odot\vartheta_{i,0}^{t}, \sum_{\tau=1}^{K}\overline{\mathbf{m}}_{j,\tau}^{t}\odot\vartheta_{i,0}^{t}\rangle\mathbb{E}[\mathbb{I}_{i}^{t}]\Big)\\
    &= \frac{1}{S^{2}K^{2}}\mathbb{E}_{K|t}\Big( \sum_{i\neq j}\langle \sum_{\tau=1}^{K}\overline{\mathbf{m}}_{i,\tau}^{t}\odot\vartheta_{i,0}^{t}, \sum_{\tau=1}^{K}\overline{\mathbf{m}}_{j,\tau}^{t}\odot\vartheta_{i,0}^{t}\rangle\frac{S(S-1)}{m(m-1)} + \sum_{i=j}\langle \sum_{\tau=1}^{K}\overline{\mathbf{m}}_{i,\tau}^{t}\odot\vartheta_{i,0}^{t}, \sum_{\tau=1}^{K}\overline{\mathbf{m}}_{j,\tau}^{t}\odot\vartheta_{i,0}^{t}\rangle\frac{S}{m}\Big)\\
    &= \frac{1}{S^{2}K^{2}}\mathbb{E}_{K|t}\Big( \sum_{i, j}\langle \sum_{\tau=1}^{K}\overline{\mathbf{m}}_{i,\tau}^{t}\odot\vartheta_{i,0}^{t}, \sum_{\tau=1}^{K}\overline{\mathbf{m}}_{j,\tau}^{t}\odot\vartheta_{i,0}^{t}\rangle\frac{S(S-1)}{m(m-1)} + \sum_{i}\Vert\sum_{\tau=1}^{K}\overline{\mathbf{m}}_{i,\tau}^{t}\odot\vartheta_{i,0}^{t}\Vert^{2}\frac{S(m-S)}{m(m-1)}\Big)\\
    &= \frac{(S-1)}{Sm(m-1)}\mathbb{E}_{K|t}\Vert\frac{1}{K}\sum_{i,\tau} \overline{\mathbf{m}}_{i,\tau}^{t}\odot\vartheta_{i,0}^{t}\Vert^{2} + \frac{(m-S)}{Sm(m-1)}\sum_{i}\mathbb{E}_{K|t}\Vert\frac{1}{K}\sum_{\tau}\overline{\mathbf{m}}_{i,\tau}^{t}\odot\vartheta_{i,0}^{t}\Vert^{2}\\
    &\leq \mathbb{E}_{K|t}\Vert\frac{1}{mK}\sum_{i,\tau} \overline{\mathbf{m}}_{i,\tau}^{t}\odot\vartheta_{i,0}^{t}\Vert^{2} + \frac{(m-S)}{Sm(m-1)}\sum_{i}\mathbb{E}_{K|t}\Vert\frac{1}{K}\sum_{\tau}\overline{\mathbf{m}}_{i,\tau}^{t}\odot\vartheta_{i,0}^{t}\Vert^{2}.\\
\end{align*}

About the $\Delta^{t}$, suppose that all the variable vector belongs to $\mathbb{R}^{d}$, and let $\mathbf{x}_{(j)}$ represents the $j$-th element in $\mathbf{x}$, we bound it as:
\begin{align*}
    \Delta^{t}
    &= \mathbb{E}_{K|t}\Vert\frac{1}{SK}\sum_{i,\tau} \mathbf{m}_{i,\tau}^{t}\odot(\vartheta_{i,\tau}^{t}-\vartheta_{i,0}^{t})\Vert^{2}\\
    &\overset{(a)}{\leq} \frac{1}{SK}\sum_{i,\tau}\mathbb{E}_{K|t}\Vert \mathbf{m}_{i,\tau}^{t}\odot(\vartheta_{i,\tau}^{t}-\vartheta_{i,0}^{t})\Vert^{2}\\
    &\overset{(b)}{\leq} \frac{G_{g}^{2}}{SK}\sum_{i,\tau}\mathbb{E}_{K|t}\Vert \vartheta_{i,\tau}^{t}-\vartheta_{i,0}^{t}\Vert^{2}\\
    &= \frac{G_{g}^{2}}{SK}\mathbb{E}_{K|t}\sum_{i,\tau}\sum_{j}^{d} \big((\vartheta_{i,\tau}^{t})_{(j)}-(\vartheta_{i,0}^{t})_{(j)}\big)^{2}\\
    &= \frac{G_{g}^{2}}{SK}\mathbb{E}_{K|t}\sum_{i,\tau}\sum_{j}^{d} \big((\frac{1}{\sqrt{\hat{\mathbf{v}}_{i,\tau}^{t}}})_{(j)}-(\frac{1}{\sqrt{\hat{\mathbf{v}}_{i,0}^{t}}})_{(j)}\big)^{2}\\
    &= \frac{G_{g}^{2}}{SK}\mathbb{E}_{K|t}\sum_{i,\tau}\sum_{j}^{d} \big(\frac{(\sqrt{\hat{\mathbf{v}}_{i,0}^{t}})_{(j)}-(\sqrt{\hat{\mathbf{v}}_{i,\tau}^{t}})_{(j)}}{(\sqrt{\hat{\mathbf{v}}_{i,\tau}^{t}})_{(j)}(\sqrt{\hat{\mathbf{v}}_{i,0}^{t}})_{(j)}}\big)^{2}\\
    &= \frac{G_{g}^{2}}{SK}\mathbb{E}_{K|t}\sum_{i,\tau}\sum_{j}^{d} \frac{(\hat{\mathbf{v}}_{i,0}^{t})_{(j)}+(\hat{\mathbf{v}}_{i,\tau}^{t})_{(j)}-2(\sqrt{\hat{\mathbf{v}}_{i,0}^{t}})_{(j)}(\sqrt{\hat{\mathbf{v}}_{i,\tau}^{t}})_{(j)}}{(\hat{\mathbf{v}}_{i,\tau}^{t})_{(j)}(\hat{\mathbf{v}}_{i,0}^{t})_{(j)}}\\
    &\overset{(c)}{\leq} \frac{G_{g}^{2}}{SK}\mathbb{E}_{K|t}\sum_{i,\tau}\sum_{j}^{d} \frac{(\hat{\mathbf{v}}_{i,0}^{t})_{(j)}+(\hat{\mathbf{v}}_{i,\tau}^{t})_{(j)}-2(\hat{\mathbf{v}}_{i,0}^{t})_{(j)}}{(\hat{\mathbf{v}}_{i,\tau}^{t})_{(j)}(\hat{\mathbf{v}}_{i,0}^{t})_{(j)}}\\
    &\overset{(d)}{\leq} \frac{G_{g}^{2}}{SK\epsilon_{v}^{4}}\mathbb{E}_{K|t}\sum_{i,\tau}\sum_{j}^{d} \big((\hat{\mathbf{v}}_{i,\tau}^{t})_{(j)}-(\hat{\mathbf{v}}_{i,0}^{t})_{(j)}\big)\\
    &\overset{(e)}{\leq} \frac{G_{g}^{2}}{\epsilon_{v}^{4}}\mathbb{E}_{K|t}\sum_{j}^{d} \big((\overline{\hat{\mathbf{v}}}_{K}^{t})_{(j)}-(\overline{\hat{\mathbf{v}}}_{0}^{t})_{(j)}\big),\\
\end{align*}
where $(\overline{\hat{\mathbf{v}}}_{\tau}^{t})_{(j)}=\frac{1}{m}\sum_{i}(\hat{\mathbf{v}}_{i,\tau}^{t})_{(j)}$. $(a)$ applies the lemma \ref{lemma_cauchy}; $(b)$ applies the lemma \ref{lemma_hadamard_norm} and the assumption of bounded gradient $\Vert \mathbf{m} \Vert_{\infty} \leq G_{g}$; $(c)$ applies the fact $\hat{\mathbf{v}}_{i,\tau}^{t} \geq \hat{\mathbf{v}}_{i,0}^{t}$ for $\tau \geq 1$; $(d)$ applies the fact that $(\mathbf{v}_{i,\tau}^{t})_{(j)} \geq \epsilon_{v}$ for $\forall (i,\tau,t)$; $(e)$ applies applies $(\overline{\hat{\mathbf{v}}}_{\tau}^{t})_{(j)} \geq (\overline{\hat{\mathbf{v}}}_{\tau-1}^{t})_{(j)} \geq \cdots \geq (\overline{\hat{\mathbf{v}}}_{0}^{t})_{(j)}$ ($\hat{\mathbf{v}}_{i,\tau}^{t} \geq \hat{\mathbf{v}}_{i,\tau-1}^{t} \geq \cdots \geq \hat{\mathbf{v}}_{i,0}^{t}$ for $\forall (i,\tau,t)$).\\\\
Combining the three terms, we can bound $\mathbf{R}_{1}$ as:
\begin{align*}
    \mathbf{R}_{1}
    &\leq  2\eta^{2}\eta_{l}^{2}\Delta^{t} + 2\eta^{2}\eta_{l}^{2}\mathbf{M}^{t}\\
    &\leq 4\eta^{2}\eta_{l}^{2}(\frac{\beta_{1}^{2}G_{\vartheta}^{2}\sigma_{l}^{2}}{SK} + \mathbf{W}^{t}) + 2\eta^{2}\eta_{l}^{2}\Delta^{t}\\
    &\leq 4\eta^{2}\eta_{l}^{2}\Big(\frac{\beta_{1}^{2}G_{\vartheta}^{2}\sigma_{l}^{2}}{SK}+\mathbb{E}_{K|t}\Vert\frac{1}{mK}\sum_{i,\tau} \overline{\mathbf{m}}_{i,\tau}^{t}\odot\vartheta_{i,0}^{t}\Vert^{2} + \frac{(m-S)}{Sm(m-1)}\sum_{i}\mathbb{E}_{K|t}\Vert\frac{1}{K}\sum_{\tau}\overline{\mathbf{m}}_{i,\tau}^{t}\odot\vartheta_{i,0}^{t}\Vert^{2}\Big)\\
    &\quad  + \frac{2\eta^{2}\eta_{l}^{2}G_{g}^{2}}{\epsilon_{v}^{4}}\mathbb{E}_{K|t}\sum_{j}^{d} \big((\overline{\hat{\mathbf{v}}}_{K}^{t})_{(j)}-(\overline{\hat{\mathbf{v}}}_{0}^{t})_{(j)}\big).\\
\end{align*}
\subsubsection{Bounded R2}
Next we bound the term $\mathbf{R}_{2}$, noticing that $\vartheta_{i,0}^{t}$ can be regarded as a constant under the conditional expectation $\mathbb{E}_{\tau|t}$ at time $\tau$ in $\mathbf{R}_{2}$ and $\vartheta_{i,0}^{t}=\frac{1}{\sqrt{\mathbf{v}^{t}}}$ is independent of $i$, We have:
\begin{align*}
    \mathbf{R}_{2}
    &= \mathbb{E}_{K|t}\langle\nabla F(\mathbf{z}^{t}), - \eta\eta_{l}\frac{1}{mK}\sum_{i}\sum_{\tau=1}^{K}\mathbf{m}_{i,\tau}^{t}\odot\vartheta_{i,0}^{t}\rangle\\
    &\overset{(a)}{=} -\eta\eta_{l}\langle\nabla F(\mathbf{z}^{t})\odot\sqrt{\vartheta_{i,0}^{t}}, \frac{1}{mK}\sum_{i}\sum_{\tau=1}^{K}\mathbb{E}_{K|t}\big(\mathbf{m}_{i,\tau}^{t}+\nabla F(\mathbf{z}^{t})-\nabla F_{i}(\mathbf{z}^{t})\big)\odot\sqrt{\vartheta_{i,0}^{t}}\rangle\\
    &= -\eta\eta_{l}\Vert\nabla F(\mathbf{z}^{t})\odot\sqrt{\vartheta_{i,0}^{t}}\Vert^{2}-\eta\eta_{l}\langle\nabla F(\mathbf{z}^{t})\odot\sqrt{\vartheta_{i,0}^{t}}, \frac{1}{mK}\sum_{i}\sum_{\tau=1}^{K}\mathbb{E}_{K|t}\big(\mathbf{m}_{i,\tau}^{t}-\nabla F_{i}(\mathbf{z}^{t})\big)\odot\sqrt{\vartheta_{i,0}^{t}}\rangle\\
    &\overset{(b)}{=} -\eta\eta_{l}\Vert\nabla F(\mathbf{z}^{t})\odot\sqrt{\vartheta_{i,0}^{t}}\Vert^{2}+\frac{\eta\eta_{l}}{2}\Vert\nabla F(\mathbf{z}^{t})\odot\sqrt{\vartheta_{i,0}^{t}}\Vert^{2}+\frac{\eta\eta_{l}}{2}\mathbb{E}_{K|t}\Vert\frac{1}{mK}\sum_{i}\sum_{\tau=1}^{K}\big(\overline{\mathbf{m}}_{i,\tau}^{t}-\nabla F_{i}(\mathbf{z}^{t})\big)\odot\sqrt{\vartheta_{i,0}^{t}}\Vert^{2}\\
    &\quad - \frac{\eta\eta_{l}}{2}\mathbb{E}_{K|t}\Vert\frac{1}{mK}\sum_{i}\sum_{\tau=1}^{K}\overline{\mathbf{m}}_{i,\tau}^{t}\odot\sqrt{\vartheta_{i,0}^{t}}\Vert^{2}\\
    &= \frac{\eta\eta_{l}}{2}\underbrace{\mathbb{E}_{K|t}\Vert\frac{1}{mK}\sum_{i}\sum_{\tau=1}^{K}\big(\overline{\mathbf{m}}_{i,\tau}^{t}-\nabla F_{i}(\mathbf{z}^{t})\big)\odot\sqrt{\vartheta_{i,0}^{t}}\Vert^{2}}_{\mathbf{R}_{2.a}}-\frac{\eta\eta_{l}}{2}\underbrace{\mathbb{E}_{K|t}\Vert\frac{1}{mK}\sum_{i}\sum_{\tau=1}^{K}\overline{\mathbf{m}}_{i,\tau}^{t}\odot\sqrt{\vartheta_{i,0}^{t}}\Vert^{2}}_{\mathbf{R}_{2.b}}\\
    &\quad -\frac{\eta\eta_{l}}{2}\Vert\nabla F(\mathbf{z}^{t})\odot\sqrt{\vartheta_{i,0}^{t}}\Vert^{2}.\\
\end{align*}
$(a)$ applies the fact that $\nabla F(\mathbf{z}^{t})=\frac{1}{m}\sum_{i}\nabla F_{i}(\mathbf{z}^{t})$; $(b)$ applies lemma \ref{lemma_product}.\\\\
Next, we bound the term $\mathbf{R}_{2.a}$. By shortening $\sum_{i}\sum_{\tau=1}^{K}$ as $\sum_{i,\tau}$ we have:
\begin{align*}
    \mathbf{R}_{2.a}
    &= \underbrace{\mathbb{E}_{K|t}\Vert\frac{1}{mK}\sum_{i,\tau}\big(\overline{\mathbf{m}}_{i,\tau}^{t}-\nabla F_{i}(\mathbf{z}^{t})\big)\odot\sqrt{\vartheta_{i,0}^{t}}\Vert^{2}}_{\mathbf{A}_{K}^{t}}\\
    &= \mathbb{E}_{K|t}\Vert\frac{1}{mK}\sum_{i,\tau}\Big((1-\beta_{1})\big(\nabla F_{i}(\mathbf{x}_{i,\tau}^{t})-\nabla F_{i}(\mathbf{z}^{t})\big)+\beta_{1}\big(\overline{\mathbf{m}}_{i,\tau-1}^{t}-\nabla F_{i}(\mathbf{z}^{t})\big)\Big)\odot\sqrt{\vartheta_{i,0}^{t}}\Vert^{2}\\
    &\overset{(a)}{\leq} \beta_{1}\underbrace{\mathbb{E}_{K|t}\Vert\frac{1}{mK}\sum_{i,\tau}\big(\overline{\mathbf{m}}_{i,\tau-1}^{t}-\nabla F_{i}(\mathbf{z}^{t})\big)\odot\sqrt{\vartheta_{i,0}^{t}}\Vert^{2}}_{\mathbf{A}_{K-1}^{t}}\\
    &\quad + (1-\beta_{1})\underbrace{\mathbb{E}_{K|t}\Vert\frac{1}{mK}\sum_{i,\tau}\big(\nabla F_{i}(\mathbf{x}_{i,\tau}^{t})-\nabla F_{i}(\mathbf{z}^{t})\big)\odot\sqrt{\vartheta_{i,0}^{t}}\Vert^{2}}_{\mathbf{B}_{K}^{t}}.\\
\end{align*}
$(a)$ applies the Jensen's inequality.\\\\
According to the restart momentum in algorithm\ref{algorithm_fedLada}, $\mathbf{m}_{i,0}^{t}=\mathbf{0}$. Here we add the definition of $\mathbf{m}_{i,\tau}^{t}$ as:
\begin{align*}
    \mathbf{m}_{i,\tau}^{t}=
    \begin{cases}
        0& \text{$\tau \leq 0$}\\
        \beta_{1}\mathbf{m}_{i,\tau-1}^{t} + (1-\beta_{1})\mathbf{g}_{i,\tau}^{t}& \text{$\tau \geq 1$}\\
\end{cases}.
\end{align*}
The addition here makes $\mathbf{A}_{K}$ still have a recursion formulation when $\tau \leq 0$ and obviously we have $\mathbf{A}_{0}^{t}=\mathbb{E}_{K|t}\Vert\frac{1}{mK}\sum_{i,-K+1}^{0}\big(\overline{\mathbf{m}}_{i,\tau}^{t}-\nabla F_{i}(\mathbf{z}^{t})\big)\odot\sqrt{\vartheta_{i,0}^{t}}\Vert^{2}=\Vert\nabla F(\mathbf{z}^{t})\odot\sqrt{\vartheta_{i,0}^{t}}\Vert^{2}$. And we have the recursion:
\begin{align*}
    \mathbf{A}_{K}^{t} &\leq \beta_{1}\mathbf{A}_{K-1}^{t} + (1-\beta_{1})\mathbf{B}_{K}^{t},\\
    \beta_{1}\mathbf{A}_{K-1}^{t} &\leq \beta_{1}^{2}\mathbf{A}_{K-2}^{t} + (1-\beta_{1})\beta_{1}\mathbf{B}_{K-1}^{t},\\
    \beta_{1}^{2}\mathbf{A}_{K-2}^{t} &\leq \beta_{1}^{3}\mathbf{A}_{K-3}^{t} + (1-\beta_{1})\beta_{1}^{2}\mathbf{B}_{K-2}^{t},\\
    &\quad \cdots,\\
    \beta_{1}^{K-1}\mathbf{A}_{1}^{t} &\leq \beta_{1}^{K}\mathbf{A}_{0} + (1-\beta_{1})\beta_{1}^{K-1}\mathbf{B}_{1}^{t}.\\
\end{align*}
Add up the above inequalities:
\begin{equation}\label{ak}
    \mathbf{A}_{K}^{t} \leq \beta_{1}^{K}\mathbf{A}_{0}^{t}+(1-\beta_{1})\sum_{\tau=1}^{K}\beta_{1}^{K-\tau}\mathbf{B}_{\tau}^{t},\\
\end{equation}
where $\mathbf{B}_{k}^{t}=\mathbb{E}_{K|t}\Vert\frac{1}{mK}\sum_{i,\tau=1+k-K}^{k}\big(\nabla F_{i}(\mathbf{x}_{i,\tau}^{t})-\nabla F_{i}(\mathbf{z}^{t})\big)\odot\sqrt{\vartheta_{i,0}^{t}}\Vert^{2}$.\\\\
Same as $\mathbf{m}_{i,\tau}^{t}$, we add the definition of $\nabla F_{i}(\mathbf{x}_{i,\tau}^{t})$ as:
\begin{align*}
    \nabla F_{i}(\mathbf{x}_{i,\tau}^{t})=
    \begin{cases}
        0& \text{$\tau \leq 0$}\\
        \nabla F_{i}(\mathbf{x}_{i,\tau}^{t})& \text{$\tau \geq 1$}\\
\end{cases}.
\end{align*}
Then we bound $\mathbf{B}_{k}^{t}$ as:
\begin{align*}
    \mathbf{B}_{k}^{t}
    &= \mathbb{E}_{K|t}\Vert\frac{1}{mK}\sum_{i,\tau=k+1-K}^{k}\big(\nabla F_{i}(\mathbf{x}_{i,\tau}^{t})-\nabla F_{i}(\mathbf{z}^{t})\big)\odot\sqrt{\vartheta_{i,0}^{t}}\Vert^{2}\\
    &= \mathbb{E}_{K|t}\Vert\frac{1}{mK}\Big(\sum_{i,\tau=1}^{k}\big(\nabla F_{i}(\mathbf{x}_{i,\tau}^{t})-\nabla F_{i}(\mathbf{z}^{t})\big)+\sum_{i,\tau=k+1-K}^{0}\big(-\nabla F_{i}(\mathbf{z}^{t})\big)\Big)\odot\sqrt{\vartheta_{i,0}^{t}}\Vert^{2}\\
    &\overset{(a)}{\leq} \frac{1}{mK}\sum_{i,\tau=1}^{k}\mathbb{E}_{k|t}\Vert\big(\nabla F_{i}(\mathbf{x}_{i,\tau}^{t})-\nabla F_{i}(\mathbf{z}^{t})\big)\odot\sqrt{\vartheta_{i,0}^{t}}\Vert^{2} + \frac{K-k}{K}\Vert\nabla F(\mathbf{z}^{t})\odot\sqrt{\vartheta_{i,0}^{t}}\Vert^{2}\\
    &\overset{(b)}{\leq} \frac{G_{\vartheta}}{mK}\sum_{i,\tau=1}^{k}\mathbb{E}_{k|t}\Vert\nabla F_{i}(\mathbf{x}_{i,\tau}^{t}) - \nabla F_{i}(\mathbf{x}^{t}) + \nabla F_{i}(\mathbf{x}^{t}) -\nabla F_{i}(\mathbf{z}^{t})\Vert^{2} + (1-\frac{k}{K})\Vert\nabla F(\mathbf{z}^{t})\odot\sqrt{\vartheta_{i,0}^{t}}\Vert^{2}\\
    &\overset{(c)}{\leq} (1+\mu)\frac{G_{\vartheta}}{mK}\sum_{i,\tau=1}^{k}\mathbb{E}_{k|t}\Vert\nabla F_{i}(\mathbf{x}_{i,\tau}^{t})-\nabla F_{i}(\mathbf{x}^{t})\Vert^{2} + (1+\frac{1}{\mu})\frac{G_{\vartheta}}{mK}\sum_{i,\tau=1}^{k}\mathbb{E}_{k|t}\Vert\nabla F_{i}(\mathbf{x}^{t})-\nabla F_{i}(\mathbf{z}^{t})\Vert^{2}\\
    &\quad + (1-\frac{k}{K})\Vert\nabla F(\mathbf{z}^{t})\odot\sqrt{\vartheta_{i,0}^{t}}\Vert^{2}\\
    &\overset{(d)}{\leq} (1+\mu)\frac{G_{\vartheta}L^{2}}{mK}\sum_{i,\tau=1}^{k}\mathbb{E}_{k|t}\Vert \mathbf{x}_{i,\tau}^{t}-\mathbf{x}^{t}\Vert^{2} + (1+\frac{1}{\mu})\frac{G_{\vartheta}L^{2}}{mK}\sum_{i,\tau=1}^{k}\mathbb{E}_{k|t}\Vert \mathbf{x}^{t}- \mathbf{z}^{t}\Vert^{2} + (1-\frac{k}{K})\Vert\nabla F(\mathbf{z}^{t})\odot\sqrt{\vartheta_{i,0}^{t}}\Vert^{2}\\
    &= (1+\mu)G_{\vartheta}L^{2}\frac{1}{mK}\sum_{i,\tau=1}^{k}\mathbb{E}_{k|t}\Vert \mathbf{x}_{i,\tau}^{t}-\mathbf{x}^{t}\Vert^{2} + (1+\frac{1}{\mu})G_{\vartheta}L^{2}(\frac{1-\alpha}{\alpha})^{2}\frac{1}{mK}\sum_{i,\tau=1}^{k}\mathbb{E}_{k|t}\Vert \mathbf{x}^{t}-\mathbf{x}^{t-1}\Vert^{2}\\
    &\quad + (1-\frac{k}{K})\Vert\nabla F(\mathbf{z}^{t})\odot\sqrt{\vartheta_{i,0}^{t}}\Vert^{2}\\
    &= (1+\mu)G_{\vartheta}L^{2}\frac{1}{mK}\underbrace{\sum_{i,\tau=1}^{k}\mathbb{E}_{k|t}\Vert \mathbf{x}_{i,\tau}^{t}-\mathbf{x}^{t}\Vert^{2}}_{\mathbf{\epsilon}_{k}^{t}} + (1+\frac{1}{\mu})(\frac{1-\alpha}{\alpha})^{2}G_{\vartheta}L^{2}\eta^{2}\eta_{l}^{2}\frac{k}{K}\mathbb{E}_{k|t}\Vert\mathbf{g}_{a}^{t}\Vert^{2}\\
    &\quad +(1-\frac{k}{K})\Vert\nabla F(\mathbf{z}^{t})\odot\sqrt{\vartheta_{i,0}^{t}}\Vert^{2}.\\
\end{align*}
$(a)$ applies lemma \ref{lemma_cauchy}; $(b)$ applies lemma \ref{lemma_hadamard_norm} and let $G_{\vartheta} = \max\{\Vert\vartheta_{i,0}^{t}\Vert_{\infty}\}$; $(c)$ applies lemma \ref{lemma_youngs}; $(d)$ applies the Assumption \ref{smoothness}.\\\\
$\mathbf{\epsilon}_{k}^{t}$ measures the average moving of $\mathbf{x}$ during the entire $k$ training iterations on the clients. In the vanilla SGD optimization, this part contributes most of the error due to the heterogeneity of the dataset (usually a constant bound). Many variance reduction techniques play an important role in controlling the variance of the offset $\mathbf{\epsilon}_{k}^{t}$ and achieve better experiment results and faster convergence. We will give the theoretical bound of $\mathbf{\epsilon}_{k}^{t}$ in our algorithm in the next part.
\begin{align*}
    \mathbf{\epsilon}_{k}^{t}
    &= \sum_{i,\tau=1}^{k}\mathbb{E}_{k|t}\Vert \mathbf{x}_{i,\tau}^{t}-\mathbf{x}^{t}\Vert^{2}\\
    &= \sum_{i,\tau=1}^{k}\mathbb{E}_{k|t}\Vert\mathbf{x}_{i,\tau-1}^{t}-\eta_{l}(\alpha\mathbf{m}_{i,\tau}^{t}+(1-\alpha)\mathbf{g}_{a}^{t})\odot \vartheta_{i,\tau}^{t}-\mathbf{x}^{t}\Vert^{2}\\
    &\overset{(a)}{\leq} (1+a)\sum_{i,\tau=1}^{k}\mathbb{E}_{k|t}\Vert\mathbf{x}_{i,\tau-1}^{t}-\mathbf{x}^{t}\Vert^{2} + (1+\frac{1}{a})\eta_{l}^{2}\sum_{i,\tau=1}^{k}\mathbb{E}_{k|t}\Vert\big(\alpha\mathbf{m}_{i,\tau}^{t}+(1-\alpha)\mathbf{g}_{a}^{t}\big)\odot \vartheta_{i,\tau}^{t}\Vert^{2}\\
    &= (1+a)\sum_{i,\tau=0}^{k-1}\mathbb{E}_{k|t}\Vert\mathbf{x}_{i,\tau}^{t}-\mathbf{x}^{t}\Vert^{2} + (1+\frac{1}{a})\eta_{l}^{2}\sum_{i,\tau=1}^{k}\mathbb{E}_{k|t}\Vert\big(\alpha\mathbf{m}_{i,\tau}^{t}+(1-\alpha)\mathbf{g}_{a}^{t}\big)\odot \vartheta_{i,\tau}^{t}\Vert^{2}\\
    &\overset{(b)}{=} (1+a)\sum_{i,\tau=1}^{k-1}\mathbb{E}_{k|t}\Vert\mathbf{x}_{i,\tau}^{t}-\mathbf{x}^{t}\Vert^{2} + (1+\frac{1}{a})\eta_{l}^{2}\sum_{i,\tau=1}^{k}\mathbb{E}_{k|t}\Vert\big(\alpha\mathbf{m}_{i,\tau}^{t}+(1-\alpha)\mathbf{g}_{a}^{t}\big)\odot \vartheta_{i,\tau}^{t}\Vert^{2}\\
    &\overset{(c)}{\leq} (1+a)\sum_{i,\tau=1}^{k-1}\mathbb{E}_{k|t}\Vert\mathbf{x}_{i,\tau}^{t}-\mathbf{x}^{t}\Vert^{2}+\alpha(1+\frac{1}{a})\eta_{l}^{2}\sum_{i,\tau=1}^{k}\mathbb{E}_{k|t}\Vert\mathbf{m}_{i,\tau}^{t}\odot \vartheta_{i,\tau}^{t}\Vert^{2}\\
    &\quad +(1-\alpha)(1+\frac{1}{a})\eta_{l}^{2}\sum_{i,\tau=1}^{k}\mathbb{E}_{k|t}\Vert\mathbf{g}_{a}^{t}\odot \vartheta_{i,\tau}^{t}\Vert^{2}\\
    &\overset{(d)}{\leq} (1+a)\underbrace{\sum_{i,\tau=1}^{k-1}\mathbb{E}_{k|t}\Vert\mathbf{x}_{i,\tau}^{t}-\mathbf{x}^{t}\Vert^{2}}_{\mathbf{\epsilon}_{k-1}^{t}} +\alpha(1+\frac{1}{a})\eta_{l}^{2}\underbrace{\sum_{i,\tau=1}^{k}\mathbb{E}_{k|t}\Vert\mathbf{m}_{i,\tau}^{t}\odot \vartheta_{i,\tau}^{t}\Vert^{2}}_{\mathbf{S}_{k}^{t}}\\
    &\quad +(1-\alpha)(1+\frac{1}{a})G_{\vartheta}^{2}\eta_{l}^{2}mk\Vert\mathbf{g}_{a}^{t}\Vert^{2}.\\
\end{align*}
$(a)$ applies the lemma \ref{lemma_youngs}; $(b)$ applies the fact that $\mathbf{x}_{i,0}^{t} = \mathbf{x}^{t}$ defined in Algorithm \ref{algorithm_fedLada} and eliminate the term at $\tau=0$; $(c)$ applies the Jensen's inequality; $(d)$ applies the lemma \ref{lemma_hadamard_norm} and lemma \ref{lemma_hadamard_largernorm} and $G_{\vartheta}$ is defined above.\\\\
Then we bound the $\mathbf{S}_{k}^{t}$. This term describes the average norm of the adaptive gradient for each local iteration. Applying a simple transformation, we can get the following inequality:
\begin{align*}
        \mathbf{S}_{k}^{t}
        &= \sum_{i,\tau=1}^{k}\mathbb{E}_{k|t}\Vert \mathbf{m}_{i,\tau}^{t}\odot \vartheta_{i,\tau}^{t}\Vert^{2}\\
        &= \sum_{i,\tau=1}^{k}\mathbb{E}_{k|t}\Vert\big((1-\beta_{1})\mathbf{g}_{i,\tau}^{t}+\beta_{1}\mathbf{m}_{i,\tau-1}^{t}\big)\odot \vartheta_{i,\tau}^{t}\Vert^{2}\\
        &\overset{(a)}{\leq} (1-\beta_{1})\sum_{i,\tau=1}^{k}\mathbb{E}_{k|t}\Vert\mathbf{g}_{i,\tau}^{t}\odot \vartheta_{i,\tau}^{t}\Vert^{2} + \beta_{1}\sum_{i,\tau=1}^{k}\mathbb{E}_{k|t}\Vert\mathbf{m}_{i,\tau-1}^{t}\odot \vartheta_{i,\tau}^{t}\Vert^{2}\\
        &= (1-\beta_{1})\sum_{i,\tau=1}^{k}\mathbb{E}_{k|t}\Vert\big(\mathbf{g}_{i,\tau}^{t}-\nabla F_{i}(\mathbf{x}_{i,\tau}^{t})+\nabla F_{i}(\mathbf{x}_{i,\tau}^{t})-\nabla F_{i}(\mathbf{x}^{t})+\nabla F_{i}(\mathbf{x}^{t})-\nabla F(\mathbf{x}^{t})+\nabla F(\mathbf{x}^{t})\big)\odot\vartheta_{i,\tau}^{t}\Vert^{2}\\
        &\quad + \beta_{1}\sum_{i,\tau=0}^{k-1}\mathbb{E}_{k|t}\Vert\mathbf{m}_{i,\tau}^{t}\odot \vartheta_{i,\tau+1}^{t}\Vert^{2}\\
        &\overset{(b)}{\leq} (1-\beta_{1})\sum_{i,\tau=1}^{k}\mathbb{E}_{k|t}\Vert\big(\mathbf{g}_{i,\tau}^{t}-\nabla F_{i}(\mathbf{x}_{i,\tau}^{t})+\nabla F_{i}(\mathbf{x}_{i,\tau}^{t})-\nabla F_{i}(\mathbf{x}^{t})+\nabla F_{i}(\mathbf{x}^{t})-\nabla F(\mathbf{x}^{t})+\nabla F(\mathbf{x}^{t})\big)\odot\vartheta_{i,\tau}^{t}\Vert^{2}\\
        &\quad + \beta_{1}\sum_{i,\tau=1}^{k-1}\mathbb{E}_{k|t}\Vert\mathbf{m}_{i,\tau}^{t}\odot \vartheta_{i,\tau}^{t}\Vert^{2}\\
        &\overset{(c)}{\leq} \beta_{1}\sum_{i,\tau=1}^{k-1}\mathbb{E}_{k|t}\Vert\mathbf{m}_{i,\tau}^{t}\odot \vartheta_{i,\tau}^{t}\Vert^{2} + 4(1-\beta_{1})\sum_{i,\tau=1}^{k}\Big( \mathbb{E}_{k|t}\Vert\big(\mathbf{g}_{i,\tau}^{t}-\nabla F_{i}(\mathbf{x}_{i,\tau}^{t})\big)\odot \vartheta_{i,\tau}^{t}\Vert^{2}+\mathbb{E}_{k|t}\Vert F(\mathbf{x}^{t})\odot\vartheta_{i,\tau}^{t}\Vert^{2}\\
        &\quad +\mathbb{E}_{k|t}\Vert\big(F_{i}(\mathbf{x}_{i,\tau}^{t})-F_{i}(\mathbf{x}^{t})\big)\odot\vartheta_{i,\tau}^{t}\Vert^{2}+\mathbb{E}_{k|t}\Vert\big(F_{i}(\mathbf{x}^{t})-F(\mathbf{x}^{t})\big)\odot\vartheta_{i,\tau}^{t}\Vert^{2}\Big)\\
        &\overset{(d)}{\leq} \beta_{1}\underbrace{\sum_{i,\tau=1}^{k-1}\mathbb{E}_{k|t}\Vert\mathbf{m}_{i,\tau}^{t}\odot \vartheta_{i,\tau}^{t}\Vert^{2}}_{\mathbf{S}_{i,k-1}^{t}} + 4(1-\beta_{1})G_{\vartheta}^{2}L^{2}\underbrace{\sum_{i,\tau=1}^{k}\mathbb{E}_{k|t}\Vert \mathbf{x}_{i,\tau}^{t}-\mathbf{x}^{t}\Vert^{2}}_{\mathbf{\epsilon}_{i,k}^{t}} + 4(1-\beta_{1})G_{\vartheta}^{2}\sigma^{2}mk.\\
\end{align*}
$(a)$ applies the Jensen's inequality; $(b)$ applies the fact that $\mathbf{m}_{i,0}^{t}=\mathbf{0}$ to eliminate the term at $\tau=0$ and the fact $\vartheta_{i,\tau}^{t} \geq \vartheta_{i,\tau+1}^{t}$ for $\tau \geq 0$; $(c)$ applies the lemma \ref{lemma_cauchy}; $(d)$ applies the lemma \ref{lemma_hadamard_norm}, lemma \ref{lemma_hadamard_largernorm}, the bounded local gradient Assumption \ref{bounded_stochastic_gradient_I}, the bounded global gradient Assumption \ref{bounded_stochastic_gradient_II} and the bounded full gradient Assumption \ref{bounded_stochastic_gradient_II}. We abbreviate $(\sigma_{l}^{2} + \sigma_{g}^{2} + \sigma_{u}^{2})$ as $\sigma^{2}$.\\\\
Here we get two relationships of $\mathbf{\epsilon}_{k}^{t}$ and $\mathbf{S}_{k}^{t}$:
\begin{align*}
    \epsilon_{k}^{t}
    &\leq (1+a)\epsilon_{i,k-1}^{t} + \alpha(1+\frac{1}{a})\eta_{l}^{2}\mathbf{S}_{k}^{t}+(1-\alpha)(1+\frac{1}{a})G_{\vartheta}^{2}\eta_{l}^{2}mk\Vert\mathbf{g}_{a}^{t}\Vert^{2},\\\\
    \mathbf{S}_{k}^{t}
    &\leq \beta_{1}\mathbf{S}_{k-1}^{t} + 4(1-\beta_{1})G_{\vartheta}^{2}L^{2}\epsilon_{k}^{t}+4(1-\beta_{1})G_{\vartheta}^{2}\sigma^{2}mk,\\
\end{align*}
where $a$ is a constant to be chosen later. Noting that $\mathbf{g}_{a}^{t}$ can be considered as a constant when we take the conditional expectation at round $t$. We add the first inequality to the second inequality multiplied by a positive parameter $\gamma$ to construct the recursive relationship, when $\gamma$ satisfies the equation : 
\begin{equation}\label{condition_equation}
    4\beta_{1}(1-\beta_{1})G_{\vartheta}^{2}L^{2}\gamma^{2}+(1+a-\beta_{1})\gamma-\alpha(1+a)(1+\frac{1}{a})\eta_{l}^{2}=0,
\end{equation}
and then we have:
\begin{equation}
    \epsilon_{k}^{t}+q\mathbf{S}_{k}^{t}\leq p(\epsilon_{k-1}^{t}+q\mathbf{S}_{k-1}^{t})+mkC,\\
\end{equation}
$$ \left\{
\begin{aligned}
C & = \frac{(1-\alpha)(1+\frac{1}{a})G_{\vartheta}^{2}\eta_{l}^{2}\Vert\mathbf{g}_{a}^{t}\Vert^{2}+4\gamma(1-\beta_{1})G_{\vartheta}^{2}\sigma^{2}}{1-4\gamma(1-\beta_{1})G_{\vartheta}^{2}L^{2}}\\
q & = \frac{\gamma-\alpha(1+\frac{1}{a})\eta_{l}^{2}}{1-4\gamma(1-\beta_{1})G_{\vartheta}^{2}L^{2}}=\frac{\gamma\beta_{1}}{1+a} \\
p & = \frac{1+a}{1-4\gamma(1-\beta_{1})G_{\vartheta}^{2}L^{2}}\\
\end{aligned}.
\right.
$$\\\\
In this part we will demonstrate there is an exact $a$ which makes $\gamma$ satisfy the conditional equation (\ref{condition_equation}) and some other special properties of the sequence \{$\epsilon_{k}^{t}+q\mathbf{S}_{k}^{t}$\}. Obviously, we can know $4\beta_{1}(1-\beta_{1})G_{\vartheta}^{2}L^{2} \ > \ 0$, $1+a-\beta_{1} \ > \ 0$ and $-\alpha(1+a)(1+\frac{1}{a})\eta_{l}^{2} \ < \ 0$. Simple analysis about solution of quadratic equations states that the condition equation (\ref{condition_equation}) must have two solutions of one positive $\gamma^{+}$ and one negative $\gamma^{-}$. The choice of the solution is closely related to the above sequence. We let $\gamma=\gamma^{+}$ which ensures that the second inequality still holds when multiplied by $\gamma$ and let $1-4\gamma^{+}(1-\beta_{1})G_{\vartheta}^{2}L^{2} \ > \ 0$ hold which ensures $p \ > \ 0$ (In fact, it can be observed that $p \ > \ 1$ when this condition is met). Then we take a sample analysis on $q$. As $q=\frac{\gamma^{+}\beta_{1}}{1+a} \ > \ 0$ and $1-4\gamma^{+}(1-\beta_{1})G_{\vartheta}^{2}L^{2} \ > \ 0$,
the $\gamma^{+}-\alpha(1+\frac{1}{a})\eta_{l}^{2}$ must be greater than 0 simultaneously. Thus the learning rate $\eta_{l}$ must satisfy the condition : $\eta_{l} \ < \ \frac{1}{2\sqrt{\alpha(1-\beta_{1})(1+\frac{1}{a})}G_{\vartheta}L}$ for an appropriate positive constant $a$ which must keep $\gamma^{+} \ < \ \frac{1}{4(1-\beta_{1})G_{\vartheta}^{2}L^{2}}$ hold simultaneously. $\gamma^{+}$ is a solution of the equation (\ref{condition_equation}), thus the following inequality must be satisfied:
\begin{equation}\label{condition_equation2}
    \frac{4\beta_{1}(1-\beta_{1})G_{\vartheta}^{2}L^{2}}{(4(1-\beta_{1})G_{\vartheta}^{2}L^{2})^{2}}+\frac{1+a-\beta_{1}}{4(1-\beta_{1})G_{\vartheta}^{2}L^{2}}-\alpha(1+a)(1+\frac{1}{a})\eta_{l}^{2} \ > \ 0.\\
\end{equation}
For $\forall \ a \ > \ 0$ and $\eta_{l}$ satisfies the previous condition, inequality (\ref{condition_equation2}) holds, which confirms that the construction of the sequence $\{ \epsilon_{k}^{t}+q\mathbf{S}_{k}^{t} \}$ exists.\\\\
Recursively on $k$,
\begin{align*}
    \epsilon_{k}^{t}+q\mathbf{S}_{k}^{t} \leq  p^{k}(\epsilon_{0}^{t} + q\mathbf{S}_{0}^{t}) + mC\sum_{\tau=1}^{k}\tau p^{k-\tau} = mC\sum_{\tau=1}^{k}\tau p^{k-\tau},\\
\end{align*}
where $\epsilon_{0}^{t}=\sum_{i}\mathbb{E}_{K|t}\Vert\mathbf{x}_{i,0}^{t}-\mathbf{x}^{t}\Vert^{2}=0$ and $\mathbf{S}_{0}^{t}=\sum_{i}\mathbb{E}_{K|t}\Vert \mathbf{m}_{i,0}^{t}\odot \vartheta_{i,0}^{t}\Vert^{2}=0$ for $\mathbf{x}_{i,0}^{t}=\mathbf{x}^{t}$ and $\mathbf{m}_{i,0}^{t}=\textbf{0}$ are defined in Algorithm \ref{algorithm_fedLada}. And $\mathbf{S}_{\tau}^{t} \geq 0$ for $\forall (i, \tau, t)$, then:
\begin{align*}
    \epsilon_{k}^{t} 
    &\leq mC\sum_{\tau=1}^{k}\tau p^{k-\tau} - q\mathbf{S}_{k}^{t} \leq mC\sum_{\tau=1}^{k}\tau p^{k-\tau} - q\mathbf{S}_{k}^{t} = mCp^{k}\sum_{\tau=1}^{k}\tau(\frac{1}{p})^{\tau}-q\mathbf{S}_{k}^{t}\\
    &\leq mC\Big(\frac{p^{k+1}}{(p-1)^{2}}- \frac{p}{(p-1)^{2}}\Big) - q\mathbf{S}_{k}^{t}=mC\frac{p(p^{k}-1)}{(p-1)^{2}}-q\mathbf{S}_{k}^{t},\\
\end{align*}
where $p = \frac{1+a}{1-4\gamma^{+}(1-\beta_{1})G_{\vartheta}^{2}L^{2}} \ > \ 1$ as $a \ > \ 0$, $1-4\gamma^{+}(1-\beta_{1})G_{\vartheta}^{2}L^{2} \ > \ 0$ and $4\gamma^{+}(1-\beta_{1})G_{\vartheta}^{2}L^{2} \ > \ 0$. The last inequality applies the lemma\ref{lemma_tpt}.\\\\
Here we let $p = 1+\frac{1}{K-1} \ > \ 1$ is a fixed constant (We have previously proved that when $\gamma^{+}$ and $\eta_{l}$ satisfy their corresponding conditions, $\exists a \ > \ 0$ which makes the equation (\ref{condition_equation}) holds when $p \ > \ 1$). As $1+a \ > \ 1$, there are new conditions for $\gamma^{+}$ and $a$ that $1-4\gamma^{+}(1-\beta_{1})G_{\vartheta}^{2}L^{2} \geq \frac{1}{2}$ and $a \leq \frac{1}{K-1}$ which makes $p$ exists. Thus, we have $\gamma^{+} \leq \frac{1}{8(1-\beta_{1})G_{\vartheta}^{2}L^{2}}$ and $\eta_{l} \leq \frac{1}{2\sqrt{2\alpha(1-\beta_{1})(1+\frac{1}{a})}G_{\vartheta}L}$. And noting that $(K-1)\Big[\Big(1+\frac{1}{K-1}\Big)^{K}-1\Big] \leq 3K$ for $K \geq 2$ (when $K=1$, $\epsilon_{1}^{t} \leq C \ < \ 3K^{2}C$ still holds), then we have:
\begin{align*}
    \frac{p(p^{k}-1)}{(p-1)^{2}} 
    &= \frac{(1+\frac{1}{K-1})\big((1+\frac{1}{K-1})^{k}-1\big)}{(1+\frac{1}{K-1}-1)^{2}}\\
    &= (K-1)(1+\frac{1}{K-1})(K-1)\big((1+\frac{1}{K-1})^{k}-1\big)\\
    &\overset{(a)}{\leq} (K-1)(1+\frac{1}{K-1})(K-1)\big((1+\frac{1}{K-1})^{K}-1\big)\leq 3K^{2}.\\
\end{align*}
$(a)$ applies $(1+\frac{1}{K-1})^{k} \leq (1+\frac{1}{K-1})^{K}$ for $k \leq K$.\\\\
We also need to give an upper bound on $\gamma^{+}$. It should be noted that when $p$ and $a$ are both fixed, $\gamma^{+}$ is actually fixed. Let $\gamma_{++}=\frac{\alpha(1+a)(1+\frac{1}{a})}{1+a-\beta_{1}}\eta_{l}^{2}=C_{1}\eta_{l}^{2}$, we have:
\begin{align*}
    &\quad 4\beta_{1}(1-\beta_{1})G_{\vartheta}^{2}L^{2}\gamma_{++}^{2}+(1+a-\beta_{1})\gamma_{++}-\alpha(1+a)(1+\frac{1}{a})\eta_{l}^{2}\\
    &= \frac{4\beta_{1}(1-\beta_{1})G_{\vartheta}^{2}L^{2}\alpha^{2}(1+a)^{2}(1+\frac{1}{a})^{2}\eta_{l}^{4}}{(1+\alpha-\beta_{1})^{2}} + \frac{(1+a-\beta_{1})\alpha(1+a)(1+\frac{1}{a})\eta_{l}^{2}}{1+a-\beta_{1}} - \alpha(1+a)(1+\frac{1}{a})\eta_{l}^{2}\\
    &= \frac{4\beta_{1}(1-\beta_{1})G_{\vartheta}^{2}L^{2}\alpha^{2}(1+a)^{2}(1+\frac{1}{a})^{2}\eta_{l}^{4}}{(1+\alpha-\beta_{1})^{2}} \ > \ 0.\\
\end{align*}
For $\gamma^{+} \ > \ 0$, $\gamma_{++} \ > \ 0$ and the above inequality holds, we have $\gamma^{+} \ < \ \gamma_{++}$. For $\gamma^{+} \leq \frac{1}{8(1-\beta_{1})G_{\vartheta}^{2}L^{2}}$.
We denote $C_{0}=\frac{KG_{\vartheta}^{2}}{1-4\gamma^{+}(1-\beta_{1})G_{\vartheta}^{2}L^{2}}\leq 2KG_{\vartheta}^{2}$. Then we have:
\begin{align*}
    \epsilon_{k}^{t} 
    &\leq mC\frac{p(p^{k}-1)}{(p-1)^{2}}- q\mathbf{S}_{k}^{t}=3mK^{2}C- q\mathbf{S}_{k}^{t}\\
    &= 3mK(1-\alpha)(1+\frac{1}{a})C_{0}\eta_{l}^{2}\Vert\mathbf{g}_{a}^{t}\Vert^{2}+12mK\gamma^{+}(1-\beta_{1})C_{0}\sigma^{2}- q\mathbf{S}_{k}^{t}\\
    &\leq 6mK(1-\alpha)(1+\frac{1}{a})KG_{\vartheta}^{2}\eta_{l}^{2}\Vert\mathbf{g}_{a}^{t}\Vert^{2}+24mK\gamma^{+}(1-\beta_{1})KG_{\vartheta}^{2}\sigma^{2}- q\mathbf{S}_{k}^{t}\\
    &\overset{(a)}{\leq} mKC_{2}\eta_{l}^{2}\Vert\mathbf{g}_{a}^{t}\Vert^{2}+24mK\gamma_{++}(1-\beta_{1})KG_{\vartheta}^{2}\sigma^{2}- q\mathbf{S}_{k}^{t}\\
    &= mKC_{2}\eta_{l}^{2}\Vert\mathbf{g}_{a}^{t}\Vert^{2}+mKC_{3}\eta_{l}^{2}\sigma^{2}- q\mathbf{S}_{k}^{t},\\
\end{align*}
where $C_{1} = \frac{\alpha(1+a)(1+\frac{1}{a})}{1+a-\beta_{1}}$, $C_{2} = 6(1-\alpha)(1+\frac{1}{a})KG_{\vartheta}^{2}$ and $C_{3} = 24(1-\beta_{1})KG_{\vartheta}^{2}C_{1}$. $(a)$ applies the fact $\gamma^{+} \ < \ \gamma_{++}$.\\\\
Combining $\mathbf{B}_{k}^{t}$ and $\mathbf{\epsilon}_{k}^{t}$, we have:
\begin{align*}
    \mathbf{B}_{k}^{t}
    &\leq (1+\mu)G_{\vartheta}L^{2}\frac{1}{mK}\mathbf{\epsilon}_{k}^{t} + (1+\frac{1}{\mu})(\frac{1-\alpha}{\alpha})^{2}G_{\vartheta}L^{2}\eta^{2}\eta_{l}^{2}\frac{k}{K}\mathbb{E}_{k|t}\Vert\mathbf{g}_{a}^{t}\Vert^{2} + (1-\frac{k}{K})\Vert\nabla F(\mathbf{z}^{t})\odot\sqrt{\vartheta_{i,0}^{t}}\Vert^{2}\\
    &\leq (1+\mu)G_{\vartheta}L^{2}\frac{1}{mK}\big(mKC_{2}\eta_{l}^{2}\Vert\mathbf{g}_{a}^{t}\Vert^{2}+mKC_{3}\eta_{l}^{2}\sigma^{2}- q\mathbf{S}_{k}^{t}\big) + (1+\frac{1}{\mu})(\frac{1-\alpha}{\alpha})^{2}G_{\vartheta}L^{2}\eta^{2}\eta_{l}^{2}\frac{k}{K}\mathbb{E}_{k|t}\Vert\mathbf{g}_{a}^{t}\Vert^{2}\\
    &\quad + (1-\frac{k}{K})\Vert\nabla F(\mathbf{z}^{t})\odot\sqrt{\vartheta_{i,0}^{t}}\Vert^{2}\\
    &\overset{(a)}{\leq}\underbrace{\big((1+\mu)G_{\vartheta}L^{2}C_{2}\eta_{l}^{2}+(1+\frac{1}{\mu})(\frac{1-\alpha}{\alpha})^{2}G_{\vartheta}L^{2}\eta^{2}\eta_{l}^{2}\big)\Vert\mathbf{g}_{a}^{t}\Vert^{2} + (1+\mu)G_{\vartheta}L^{2}C_{3}\eta_{l}^{2}\sigma^{2}}_{\mathbf{B}} - (1+\mu)G_{\vartheta}L^{2}\frac{q}{mK}\mathbf{S}_{k}^{t}\\
    &\quad + (1-\frac{k}{K})\Vert\nabla F(\mathbf{z}^{t})\odot\sqrt{\vartheta_{i,0}^{t}}\Vert^{2}.\\
\end{align*}
$(a)$ applies the fact $\frac{k}{K} \leq 1$ for $k \leq K$.\\\\
Noting in above inequality $\mathbf{B}$ is independent of $k$, according to the inequality (\ref{ak}), we can bound $\mathbf{A}_{K}^{t}$ as:
\begin{align*}
    \mathbf{A}_{K}^{t}
    &\leq \beta_{1}^{K}\mathbf{A}_{0}^{t} + (1-\beta_{1})\sum_{\tau=1}^{K}\beta_{1}^{K-\tau}\mathbf{B}_{\tau}^{t}\\
    &\leq \beta_{1}^{K}\mathbf{A}_{0}^{t} + (1-\beta_{1})\beta_{1}^{K}\sum_{\tau=1}^{K}(\frac{1}{\beta_{1}})^\tau\big(\mathbf{B}-(1+\mu)G_{\vartheta}L^{2}\frac{q}{mK}\mathbf{S}_{\tau}^{t}+(1-\frac{\tau}{K})\Vert\nabla F(\mathbf{z}^{t})\odot\sqrt{\vartheta_{i,0}^{t}}\Vert^{2}\big)\\
    &= \beta_{1}^{K}\mathbf{A}_{0}^{t} + (1-\beta_{1})\beta_{1}^{K}\Big(\mathbf{B}\sum_{\tau=1}^{K}(\frac{1}{\beta_{1}})^\tau-(1+\mu)G_{\vartheta}L^{2}\frac{q}{mK}\sum_{\tau=1}^{K}(\frac{1}{\beta_{1}})^{\tau}\mathbf{S}_{\tau}^{t}+\Vert\nabla F(\mathbf{z}^{t})\odot\sqrt{\vartheta_{i,0}^{t}}\Vert^{2}\sum_{\tau=1}^{K}(\frac{1}{\beta_{1}})^\tau(1-\frac{\tau}{K})\Big)\\
    &= \beta_{1}^{K}\mathbf{A}_{0}^{t} + (1-\beta_{1})\beta_{1}^{K}\mathbf{B}\frac{\frac{1}{\beta_{1}}\big(1-(\frac{1}{\beta_{1}})^{K}\big)}{1-\frac{1}{\beta_{1}}}-(1-\beta_{1})\beta_{1}^{K}(1+\mu)G_{\vartheta}L^{2}\frac{q}{mK}\sum_{\tau=1}^{K}(\frac{1}{\beta_{1}})^{\tau}\mathbf{S}_{\tau}^{t}\\
    &\quad + \Vert\nabla F(\mathbf{z}^{t})\odot\sqrt{\vartheta_{i,0}^{t}}\Vert^{2}(1-\beta_{1})\beta_{1}^{K}\sum_{\tau=1}^{K}(\frac{1}{\beta_{1}})^\tau(1-\frac{\tau}{K})\\
    &= \beta_{1}^{K}\mathbf{A}_{0}^{t} + \mathbf{B}(1-\beta_{1}^{K})-(1-\beta_{1})\beta_{1}^{K}(1+\mu)G_{\vartheta}L^{2}\frac{q}{mK}\sum_{\tau=1}^{K}(\frac{1}{\beta_{1}})^{\tau}\mathbf{S}_{\tau}^{t} + \Vert\nabla F(\mathbf{z}^{t})\odot\sqrt{\vartheta_{i,0}^{t}}\Vert^{2}(1-\beta_{1})\sum_{\tau=0}^{K-1}\frac{\tau\beta_{1}^{\tau}}{K}\\
    &= \beta_{1}^{K}\mathbf{A}_{0}^{t} + \mathbf{B} - (1-\beta_{1})(1+\mu)G_{\vartheta}L^{2}q\underbrace{\frac{1}{mK}\sum_{\tau=1}^{K}(\beta_{1})^{K-\tau}\mathbf{S}_{\tau}^{t}}_{\mathbf{A}} + \Vert\nabla F(\mathbf{z}^{t})\odot\sqrt{\vartheta_{i,0}^{t}}\Vert^{2}(\frac{\beta_{1}-\beta_{1}^{K}}{(1-\beta_{1})K}-\beta_{1}^{K})\\
    &\leq \frac{\beta_{1}}{(1-\beta_{1})K}\Vert\nabla F(\mathbf{z}^{t})\odot\sqrt{\vartheta_{i,0}^{t}}\Vert^{2} + \mathbf{B} - (1-\beta_{1})(1+\mu)G_{\vartheta}L^{2}q\underbrace{\frac{1}{mK}\sum_{\tau=1}^{K}(\beta_{1})^{K-\tau}\mathbf{S}_{\tau}^{t}}_{\mathbf{A}}.\\
\end{align*}
The term $\mathbf{A}$, we have the following transformation:
\begin{align*}
    \mathbf{A}
    &= \frac{1}{mK}\sum_{\tau=1}^{K}(\beta_{1})^{K-\tau}\mathbf{S}_{\tau}^{t}\\
    &= \frac{1}{mK}\sum_{\tau=1}^{K}(\beta_{1})^{K-\tau}\Big(\sum_{i,\tau^{'}=1}^{\tau}\mathbb{E}_{k|t}\Vert \mathbf{m}_{i,\tau^{'}}^{t}\odot \vartheta_{i,\tau^{'}}^{t}\Vert^{2}\Big)\\
    &= \frac{1}{mK}\sum_{i,\tau=1}^{K}\Big(\sum_{\tau^{'}=\tau}^{K}\beta_{1}^{K-\tau^{'}} \Big)\mathbb{E}_{k|t}\Vert \mathbf{m}_{i,\tau}^{t}\odot \vartheta_{i,\tau}^{t}\Vert^{2}\\
    &= \frac{1}{mK}\sum_{i,\tau=1}^{K}\Big(\frac{1-\beta_{1}^{K+1-\tau}}{1-\beta_{1}} \Big)\mathbb{E}_{k|t}\Vert \mathbf{m}_{i,\tau}^{t}\odot \vartheta_{i,\tau}^{t}\Vert^{2}\\
    &\overset{(a)}{\geq} \frac{1}{mK}\sum_{i,\tau=1}^{K}\mathbb{E}_{k|t}\Vert \mathbf{m}_{i,\tau}^{t}\odot \vartheta_{i,\tau}^{t}\Vert^{2}.\\
\end{align*}
$(a)$ applies the fact that $1-\beta_{1}^{K}\geq 1-\beta_{1}$ for $0 \ < \ \beta_{1} \ < \ 1$ and $K \geq 1$.\\\\
Then we give the bound of $\mathbf{R}_{2.a}$ as:
\begin{align*}
    \mathbf{R}_{2.a}
    &= \mathbf{A}_{K}^{t}\\
    &\leq \frac{\beta_{1}}{(1-\beta_{1})K}\Vert\nabla F(\mathbf{z}^{t})\odot\sqrt{\vartheta_{i,0}^{t}}\Vert^{2} + \mathbf{B} - (1-\beta_{1})(1+\mu)G_{\vartheta}L^{2}q\mathbf{A}\\
    &\leq \frac{\beta_{1}}{(1-\beta_{1})K}\Vert\nabla F(\mathbf{z}^{t})\odot\sqrt{\vartheta_{i,0}^{t}}\Vert^{2} + \big((1+\mu)G_{\vartheta}L^{2}C_{2}\eta_{l}^{2}+(1+\frac{1}{\mu})(\frac{1-\alpha}{\alpha})^{2}G_{\vartheta}L^{2}\eta^{2}\eta_{l}^{2}\big)\Vert\mathbf{g}_{a}^{t}\Vert^{2}\\
    &\quad + (1+\mu)G_{\vartheta}L^{2}C_{3}\eta_{l}^{2}\sigma^{2} - (1-\beta_{1})(1+\mu)G_{\vartheta}L^{2}q\frac{1}{mK}\sum_{i,\tau=1}^{K}\mathbb{E}_{k|t}\Vert \mathbf{m}_{i,\tau}^{t}\odot \vartheta_{i,\tau}^{t}\Vert^{2}\\
    &\overset{(a)}{\leq} \frac{\beta_{1}}{(1-\beta_{1})K}\Vert\nabla F(\mathbf{z}^{t})\odot\sqrt{\vartheta_{i,0}^{t}}\Vert^{2} + \big((1+\mu)G_{\vartheta}L^{2}C_{2}\eta_{l}^{2}+(1+\frac{1}{\mu})(\frac{1-\alpha}{\alpha})^{2}G_{\vartheta}L^{2}\eta^{2}\eta_{l}^{2}\big)\Vert\mathbf{g}_{a}^{t}\Vert^{2}\\
    &\quad + (1+\mu)G_{\vartheta}L^{2}C_{3}\eta_{l}^{2}\sigma^{2} - \frac{(1+\mu)\beta_{1}}{8G_{\vartheta}}\frac{1}{mK}\sum_{i,\tau=1}^{K}\mathbb{E}_{k|t}\Vert \mathbf{m}_{i,\tau}^{t}\odot \vartheta_{i,\tau}^{t}\Vert^{2}.\\
\end{align*}
$(a)$ applies that $q=\frac{\gamma^{+}\beta_{1}}{1+a}\leq\gamma^{+}\beta_{1} \leq \frac{\beta_{1}}{8(1-\beta_{1})G_{\vartheta}^{2}L^{2}}$.\\\\
Then we bound $\mathbf{g}_{a}^{t}$, which represents the global average change of each local training iteration at round $t$. In vanilla SGD optimization, $\mathbf{g}_{a}^{t}$ means the average of local gradients. In our algorithm, it denotes the average of historical adaptive gradients. According to the algorithm\ref{algorithm_fedLada}, we have:
\begin{align*}
    \mathbb{E}\Vert\mathbf{g}_{a}^{t+1}\Vert^{2}
    &= \mathbb{E}\Vert \frac{1}{\eta_{l}\eta_{g}K} \sum_{i \in \mathcal{S}^{t}}(\mathbf{x}^{t}-\mathbf{x}^{t+1})\Vert^{2}\\
    &= \mathbb{E}\Vert \frac{1}{SK}\sum_{i\in\mathcal{S}^{t}}\sum_{\tau=1}^{K}(\alpha\mathbf{m}_{i,\tau}^{t}\odot\vartheta_{i,\tau}^{t}+(1-\alpha)\mathbf{g}_{a}^{t})\Vert^{2}\\
    &\leq \alpha\mathbb{E}\Vert\frac{1}{SK}\sum_{i\in\mathcal{S}^{t}}\sum_{\tau=1}^{K}\mathbf{m}_{i,\tau}^{t}\odot\vartheta_{i,\tau}^{t}\Vert^{2}+(1-\alpha)\mathbb{E}\Vert\mathbf{g}_{a}^{t}\Vert^{2}.\\
\end{align*}
Noticing that $\Vert \overline{\mathbf{m}}_{i,0}^{t}\Vert^{2}=\Vert \mathbf{m}_{i,0}^{t}\Vert^{2}=0$, then we have:
\begin{align*}
    \mathbb{E}_{K|t}\Vert\overline{\mathbf{m}}_{i,k}^{t}\Vert^{2}
    &= \mathbb{E}_{K|t}\Vert\beta_{1}\overline{\mathbf{m}}_{i,k-1}^{t} + (1-\beta_{1})\nabla F_{i}(\mathbf{x}_{i,\tau}^{t})\Vert^{2} \\
    &\overset{(a)}{\leq} \beta_{1}\mathbb{E}_{K|t}\Vert\overline{\mathbf{m}}_{i,k-1}^{t}\Vert^{2} + (1-\beta_{1})\mathbb{E}_{K|t}\Vert\nabla F_{i}(\mathbf{x}_{i,\tau}^{t})\Vert^{2}\\
    &\overset{(b)}{\leq} \beta_{1}\mathbb{E}_{K|t}\Vert\overline{\mathbf{m}}_{i,k-1}^{t}\Vert^{2} + 2(1-\beta_{1})(\mathbb{E}_{K|t}\Vert\nabla F_{i}(\mathbf{x}_{i,k}^{t}) - \nabla F(\mathbf{x}_{i,k}^{t})\Vert^{2}+\mathbb{E}_{K|t}\Vert\nabla F(\mathbf{x}_{i,k}^{t})\Vert^{2})\\
    &\overset{(c)}{\leq} \beta_{1}\mathbb{E}_{K|t}\Vert\overline{\mathbf{m}}_{i,k-1}^{t}\Vert^{2} + 2(1-\beta_{1})(\sigma_{g}^{2}+\sigma_{u}^{2})\\
    &\leq \beta_{1}^{k}\mathbb{E}_{K|t}\Vert\overline{\mathbf{m}}_{i,0}^{t}\Vert^{2} + 2(1-\beta_{1})\sum_{\tau=1}^{k}\beta_{1}^{k-\tau}(\sigma_{g}^{2}+\sigma_{u}^{2})\\
    &= 2(1-\beta_{1}^{k})(\sigma_{g}^{2}+\sigma_{u}^{2}) \leq 2(\sigma_{g}^{2}+\sigma_{u}^{2}).\\
\end{align*}
$(a)$ applies the Jensen's inequality; $(b)$ applies the lemma \ref{lemma_cauchy}; $(c)$ applies the assumption of bounded gradient.\\\\
This is a very loose upper bound which gives a constant bound. Since this part is not the dominant error in the final conclusion, the loose upper bound is simple and practical.
We can bound the first part $\mathbb{E}\Vert\frac{1}{SK}\sum_{i\in\mathcal{S}^{t}}\sum_{\tau=1}^{K}\mathbf{m}_{i,\tau}^{t}\odot\vartheta_{i,\tau}^{t}\Vert^{2}$ (looser upper bound of $\mathbf{R}_{1}$) as:
\begin{align*}
    &\quad \mathbb{E}_{K|t}\Vert\frac{1}{SK}\sum_{i\in\mathcal{S}^{t}}\sum_{\tau=1}^{K}\mathbf{m}_{i,\tau}^{t}\odot\vartheta_{i,\tau}^{t}\Vert^{2}\\
    &\leq \frac{2G_{g}^{2}}{\epsilon_{v}^{4}}\mathbb{E}_{K|t}\sum_{j}^{d} \big((\overline{\hat{\mathbf{v}}}_{K}^{t})_{(j)}-(\overline{\hat{\mathbf{v}}}_{0}^{t})_{(j)}\big) +  \frac{2\beta_{1}^{2}G_{\vartheta}^{2}\sigma_{l}^{2}}{SK} + 2\mathbb{E}_{K|t}\Vert\frac{1}{SK}\sum_{i,\tau} \overline{\mathbf{m}}_{i,\tau}^{t}\odot\vartheta_{i,0}^{t}\Vert^{2}\\
    &\overset{(a)}{\leq} \frac{2G_{g}^{2}}{\epsilon_{v}^{4}}\mathbb{E}_{K|t}\sum_{j}^{d} \big((\overline{\hat{\mathbf{v}}}_{K}^{t})_{(j)}-(\overline{\hat{\mathbf{v}}}_{0}^{t})_{(j)}\big) +  \frac{2\beta_{1}^{2}G_{\vartheta}^{2}\sigma_{l}^{2}}{SK} + \frac{2}{SK}\sum_{i,\tau}\mathbb{E}_{K|t}\Vert \overline{\mathbf{m}}_{i,\tau}^{t}\odot\vartheta_{i,0}^{t}\Vert^{2}\\
    &\overset{(b)}{\leq} \frac{2G_{g}^{2}d(G_{g}^{2}-\epsilon_{v}^{2})}{\epsilon_{v}^{4}} + \frac{2\beta_{1}^{2}G_{\vartheta}^{2}\sigma_{l}^{2}}{SK} + \frac{2G_{\vartheta}^{2}}{SK}\sum_{i,\tau}\mathbb{E}_{K|t}\Vert \overline{\mathbf{m}}_{i,\tau}^{t}\Vert^{2}\\
    &\leq \frac{2dG_{g}^{4}}{\epsilon_{v}^{4}} + \frac{2\beta_{1}^{2}G_{\vartheta}^{2}\sigma_{l}^{2}}{SK} + 4G_{\vartheta}^{2}(\sigma_{g}^{2}+\sigma_{u}^{2}).\\
\end{align*}
$(a)$ applies lemma \ref{lemma_hadamard_largernorm}; $(b)$ applies the assumption of bounded $\mathbf{v}$ the $\epsilon_{v} \leq \mathbf{v}_{(j)} \leq G_{g}$; $(c)$ bound as the first part of $\mathbf{M}^{t}$ and the assumption of bounded gradient $\Vert \nabla F\Vert^{2} \leq \sigma_{u}^{2}$.\\\\
Noticing that $\Vert \mathbf{g}_{a}^{0}\Vert^{2}=0$ and the loose constant bounded $\mathbf{R}_{1}$, we can bound $\mathbb{E}\Vert \mathbf{g}_{a}^{t+1}\Vert^{t}$ as:
\begin{align*}
    \mathbb{E}\Vert \mathbf{g}_{a}^{t+1}\Vert^{t}
    &\leq \alpha\mathbb{E}\Vert\frac{1}{SK}\sum_{i\in\mathcal{S}^{t}}\sum_{\tau=1}^{K}\mathbf{m}_{i,\tau}^{t}\odot\vartheta_{i,\tau}^{t}\Vert^{2}+(1-\alpha)\mathbb{E}\Vert\mathbf{g}_{a}^{t}\Vert^{2}\\
    &= (1-\alpha)^{t+1}\Vert\mathbf{g}_{a}^{0}\Vert^{2} + \alpha\sum_{t^{'}}^{t}(1-\alpha)^{t-t^{'}}\mathbb{E}\Vert\frac{1}{SK}\sum_{i\in\mathcal{S}^{t}}\sum_{\tau=1}^{K}\mathbf{m}_{i,\tau}^{t^{'}}\odot\vartheta_{i,\tau}^{t^{'}}\Vert^{2}\\
    &\leq 2\Big(\frac{dG_{g}^{4}}{\epsilon_{v}^{4}} + \frac{\beta_{1}^{2}G_{\vartheta}^{2}\sigma_{l}^{2}}{SK} + 2G_{\vartheta}^{2}(\sigma_{g}^{2}+\sigma_{u}^{2})\Big)\frac{1-(1-\alpha)^{t}}{\alpha}\\
    &\leq 2\Big(\frac{dG_{g}^{4}}{\epsilon_{v}^{4}} + \frac{\beta_{1}^{2}G_{\vartheta}^{2}\sigma_{l}^{2}}{SK} + 2G_{\vartheta}^{2}(\sigma_{g}^{2}+\sigma_{u}^{2})\Big).\\
\end{align*}
Although the upper bound of $\Vert\mathbf{g}_{a}^{t}\Vert^{2}$ is a loose bound, since it contains a second-order term of the learning rate $\eta\eta_{l}$, as mentioned above, it is not the dominant influence of the convergence rate.\\\\
Thus we have the bound of $\mathbf{R}_{2}$:
\begin{align*}
    \mathbf{R}_{2}
    &= -\frac{\eta\eta_{l}}{2}\Vert \nabla F(\mathbf{z}^{t})\odot\sqrt{\vartheta_{i,0}^{t}}\Vert^{2} + \frac{\eta\eta_{l}}{2}\mathbf{R}_{2.a} - \frac{\eta\eta_{l}}{2}\mathbf{R}_{2.b}\\
    &\leq \frac{\eta\eta_{l}}{2}\Big(\frac{\beta_{1}}{(1-\beta_{1})K}\Vert\nabla F(\mathbf{z}^{t})\odot\sqrt{\vartheta_{i,0}^{t}}\Vert^{2} + \big((1+\mu)G_{\vartheta}L^{2}C_{2}\eta_{l}^{2}+(1+\frac{1}{\mu})(\frac{1-\alpha}{\alpha})^{2}G_{\vartheta}L^{2}\eta^{2}\eta_{l}^{2}\big)\Vert\mathbf{g}_{a}^{t}\Vert^{2}\\
    &\quad + (1+\mu)G_{\vartheta}L^{2}C_{3}\eta_{l}^{2}\sigma^{2} - \frac{(1+\mu)\beta_{1}}{8G_{\vartheta}}\frac{1}{mK}\sum_{i,\tau=1}^{K}\mathbb{E}_{k|t}\Vert \mathbf{m}_{i,\tau}^{t}\odot \vartheta_{i,\tau}^{t}\Vert^{2}\Big)-\frac{\eta\eta_{l}}{2}\Vert \nabla F(\mathbf{z}^{t})\odot\sqrt{\vartheta_{i,0}^{t}}\Vert^{2}\\
    &\quad -\frac{\eta\eta_{l}}{2}\mathbb{E}_{K|t}\Vert\frac{1}{mK}\sum_{i}\sum_{\tau=1}^{K}\overline{\mathbf{m}}_{i,\tau}^{t}\odot\sqrt{\vartheta_{i,0}^{t}}\Vert^{2}\\
    &\leq \frac{\eta\eta_{l}}{2}(\frac{\beta_{1}}{(1-\beta_{1})K}-1)\Vert\nabla F_{i}(\mathbf{z}^{t})\odot\sqrt{\vartheta_{i,0}^{t}}\Vert^{2} - \frac{(1+\mu)\beta_{1}\eta\eta_{l}}{16G_{\vartheta}}\frac{1}{mK}\sum_{i,\tau=1}^{K}\mathbb{E}_{k|t}\Vert \mathbf{m}_{i,\tau}^{t}\odot \vartheta_{i,\tau}^{t}\Vert^{2}\\
    &\quad + \big((1+\mu)G_{\vartheta}L^{2}C_{2}\eta\eta_{l}^{3}+(1+\frac{1}{\mu})(\frac{1-\alpha}{\alpha})^{2}G_{\vartheta}L^{2}\eta^{3}\eta_{l}^{3}\big)2\Big(\frac{dG_{g}^{4}}{\epsilon_{v}^{4}} + \frac{\beta_{1}^{2}G_{\vartheta}^{2}\sigma_{l}^{2}}{SK} + 2G_{\vartheta}^{2}(\sigma_{g}^{2}+\sigma_{u}^{2})\Big)\\
    &\quad -\frac{\eta\eta_{l}}{2}\mathbb{E}_{K|t}\Vert\frac{1}{mK}\sum_{i}\sum_{\tau=1}^{K}\overline{\mathbf{m}}_{i,\tau}^{t}\odot\sqrt{\vartheta_{i,0}^{t}}\Vert^{2} + \frac{(1+\mu)G_{\vartheta}L^{2}C_{3}\eta\eta_{l}^{3}\sigma^{2}}{2}.\\
\end{align*}
\subsubsection{Bounded R3}
Then we bound the difference term $\mathbf{R}_{3}$. we have:
\begin{align*}
    \mathbf{R}_{3}
    &=\mathbb{E}_{K|t}\langle\nabla F(\mathbf{z}^{t}), - \eta\eta_{l}\frac{1}{mK}\sum_{i}\sum_{\tau=1}^{K}\mathbf{m}_{i,\tau}^{t}\odot\big(\vartheta_{i,\tau}^{t}-\vartheta_{i,0}^{t}\big)\rangle\\
    &= -\frac{\eta\eta_{l}}{mK}\mathbb{E}_{K|t}\sum_{i,\tau}\sum_{j}^{d} \nabla F(\mathbf{z}^{t})_{(j)} \times (\mathbf{m}_{i,\tau}^{t})_{(j)} \times \big((\vartheta_{i,\tau}^{t})_{(j)} - (\vartheta_{i,0}^{t})_{(j)}\big)\\
    &\overset{(a)}{\leq} \frac{\eta\eta_{l}}{mK}\mathbb{E}_{K|t}\sum_{i,\tau}\sum_{j}^{d} \vert \nabla F(\mathbf{z}^{t})_{(j)} \vert \times \vert (\mathbf{m}_{i,\tau}^{t})_{(j)} \vert \times \vert(\vartheta_{i,\tau}^{t})_{(j)} - (\vartheta_{i,0}^{t})_{(j)}\vert\\
    &\overset{(b)}{\leq} \frac{\eta\eta_{l}G_{g}^{2}}{mK}\mathbb{E}_{K|t}\sum_{i,\tau}\sum_{j}^{d} \vert(\vartheta_{i,\tau}^{t})_{(j)} - (\vartheta_{i,0}^{t})_{(j)}\vert\\
    &= \frac{\eta\eta_{l}G_{g}^{2}}{mK}\mathbb{E}_{K|t}\sum_{i,\tau}\sum_{j}^{d} \vert(\frac{1}{\sqrt{\hat{\mathbf{v}}_{i,\tau}^{t}}})_{(j)} - (\frac{1}{\sqrt{\hat{\mathbf{v}}_{i,0}^{t}}})_{(j)}\vert\\
    &= \frac{\eta\eta_{l}G_{g}^{2}}{mK}\mathbb{E}_{K|t}\sum_{i,\tau}\sum_{j}^{d} \vert\frac{(\sqrt{\hat{\mathbf{v}}_{i,0}^{t}})_{(j)}-(\sqrt{\hat{\mathbf{v}}_{i,\tau}^{t}})_{(j)}}{(\sqrt{\hat{\mathbf{v}}_{i,\tau}^{t}})_{(j)}(\sqrt{\hat{\mathbf{v}}_{i,0}^{t}})_{(j)}}\vert\\
    &= \frac{\eta\eta_{l}G_{g}^{2}}{mK}\mathbb{E}_{K|t}\sum_{i,\tau}\sum_{j}^{d} \vert\frac{(\hat{\mathbf{v}}_{i,0}^{t})_{(j)}-(\hat{\mathbf{v}}_{i,\tau}^{t})_{(j)}}{(\sqrt{\hat{\mathbf{v}}_{i,\tau}^{t}})_{(j)}(\sqrt{\hat{\mathbf{v}}_{i,0}^{t}})_{(j)}\big((\sqrt{\hat{\mathbf{v}}_{i,\tau}^{t}})_{(j)}+(\sqrt{\hat{\mathbf{v}}_{i,0}^{t}})_{(j)}\big)}\vert\\
    &= \frac{\eta\eta_{l}G_{g}^{2}}{mK}\mathbb{E}_{K|t}\sum_{i,\tau}\sum_{j}^{d} \Big(\frac{(\hat{\mathbf{v}}_{i,\tau}^{t})_{(j)}-(\hat{\mathbf{v}}_{i,0}^{t})_{(j)}}{(\sqrt{\hat{\mathbf{v}}_{i,\tau}^{t}})_{(j)}(\sqrt{\hat{\mathbf{v}}_{i,0}^{t}})_{(j)}\big((\sqrt{\hat{\mathbf{v}}_{i,\tau}^{t}})_{(j)}+(\sqrt{\hat{\mathbf{v}}_{i,0}^{t}})_{(j)}\big)}\Big)\\
    &\overset{(c)}{\leq} \frac{\eta\eta_{l}G_{g}^{2}}{2mK\epsilon_{v}^{3}}\mathbb{E}_{K|t}\sum_{i,\tau}\sum_{j}^{d} \big((\hat{\mathbf{v}}_{i,\tau}^{t})_{(j)}-(\hat{\mathbf{v}}_{i,0}^{t})_{(j)}\big)\\
    &\overset{(d)}{\leq} \frac{\eta\eta_{l}G_{g}^{2}}{2\epsilon_{v}^{3}}\mathbb{E}_{K|t}\sum_{j}^{d} \big((\overline{\hat{\mathbf{v}}}_{K}^{t})_{(j)}-(\overline{\hat{\mathbf{v}}}_{0}^{t})_{(j)}\big).\\
\end{align*}
$(a)$ applies $\langle \mathbf{x}, \mathbf{y}\rangle \leq \vert \mathbf{x} \vert \times \vert \mathbf{y} \vert$; $(b)$ applies the assumption of bounded gradient $\Vert \nabla F \Vert_{\infty} \leq G_{g}$ and $\Vert \mathbf{m} \Vert_{\infty} \leq G_{g}$; $(c)$ applies the fact that $(\mathbf{v}_{i,\tau}^{t})_{(j)} \geq \epsilon_{v}$ for $\forall (i,\tau,t)$; $(d)$ is the same as $(e)$ in proof of bounded $\Delta^{t}$.\\\\
Thus, for the general non-convex case, we expand on $\mathbf{z}^{t}$ and take the full expectation on all round $t$ and local iteration $\tau$:
\begin{align*}
    \mathbb{E}[F(\mathbf{z}^{t+1})] 
    &\leq \mathbb{E}[F(\mathbf{z}^{t})] + \mathbb{E}\langle\nabla F(\mathbf{z}^{t}), {\mathbf{z}^{t+1}-\mathbf{z}^{t}}\rangle + \frac{L}{2}\mathbb{E}\Vert\mathbf{z}^{t+1}-\mathbf{z}^{t}\Vert^{2}\\
    &= \mathbb{E}[F(\mathbf{z}^{t})] + \frac{L}{2}\mathbb{E}[\mathbf{R}_{1}] + \mathbb{E}[\mathbf{R}_{2}] + [\mathbb{E}\mathbf{R}_{3}]\\
    &\leq \mathbb{E}[F(\mathbf{z}^{t})] + \Big(\frac{2\eta^{2}\eta_{l}^{2}(m-S)L}{S(m-1)m}\sum_{i}\mathbb{E}_{K|t}\Vert\frac{1}{K}\sum_{\tau}\overline{\mathbf{m}}_{i,\tau}^{t}\odot\vartheta_{i,0}^{t}\Vert^{2}- \frac{(1+\mu)\beta_{1}\eta\eta_{l}}{16G_{\vartheta}}\frac{1}{mK}\sum_{i,\tau}\mathbb{E}\Vert\mathbf{m}_{i,\tau}^{t}\odot\vartheta_{i,\tau}^{t}\Vert^{2}\Big)\\
    &\quad +\Big(2L\eta^{2}\eta_{l}^{2}-\frac{\eta\eta_{l}}{2}\Big)\mathbb{E}\Vert\frac{1}{mK}\sum_{i,\tau}\overline{\mathbf{m}}_{i,\tau}^{t}\odot\vartheta_{i,0}^{t}\Vert^{2} + \Big(\frac{\eta^{2}\eta_{l}^{2}G_{g}^{2}L}{\epsilon_{v}^{4}}+\frac{\eta\eta_{l}G_{g}^{2}}{2\epsilon_{v}^{3}}\Big)\mathbb{E}\sum_{j}^{d} \big((\overline{\hat{\mathbf{v}}}_{K}^{t})_{(j)}-(\overline{\hat{\mathbf{v}}}_{0}^{t})_{(j)}\big)\\
    &\quad + 2\big(\eta\eta_{l}^{3}(1+\mu)G_{\vartheta}L^{2}C_{2}+\eta^{3}\eta_{l}^{3}(1+\frac{1}{\mu})(\frac{1-\alpha}{\alpha})^{2}G_{\vartheta}L^{2}\big)\Big(\frac{dG_{g}^{4}}{\epsilon_{v}^{4}} + \frac{\beta_{1}^{2}G_{\vartheta}^{2}\sigma_{l}^{2}}{SK} + 2G_{\vartheta}^{2}(\sigma_{g}^{2}+\sigma_{u}^{2})\Big)\\
    &\quad + \frac{\eta\eta_{l}}{2}(\frac{\beta_{1}}{(1-\beta_{1})K}-1)G_{\vartheta}\mathbb{E}\Vert\nabla F(\mathbf{z}^{t})\Vert^{2} + \frac{(1+\mu)G_{\vartheta}L^{2}C_{3}\eta\eta_{l}^{3}\sigma^{2}}{2}+\frac{2\eta^{2}\eta_{l}^{2}\beta_{1}^{2}G_{\vartheta}^{2}L\sigma_{l}^{2}}{SK}.\\
\end{align*}
We can bound the difference term as:
\begin{align*}
    &\quad \Big(\frac{2\eta^{2}\eta_{l}^{2}(m-S)L}{S(m-1)m}\sum_{i}\mathbb{E}_{K|t}\Vert\frac{1}{K}\sum_{\tau}\overline{\mathbf{m}}_{i,\tau}^{t}\odot\vartheta_{i,0}^{t}\Vert^{2}- \frac{(1+\mu)\beta_{1}\eta\eta_{l}}{16G_{\vartheta}}\frac{1}{mK}\sum_{i,\tau}\mathbb{E}\Vert\mathbf{m}_{i,\tau}^{t}\odot\vartheta_{i,\tau}^{t}\Vert^{2}\Big)\\
    &\leq \Big(\frac{2\eta^{2}\eta_{l}^{2}(m-S)LG_{\vartheta}^{2}}{S(m-1)m}\sum_{i}\mathbb{E}_{K|t}\Vert\frac{1}{K}\sum_{\tau}\overline{\mathbf{m}}_{i,\tau}^{t}\Vert^{2}- \frac{(1+\mu)\beta_{1}\eta\eta_{l}\min\{G_{\vartheta}^{2}\}}{16G_{\vartheta}}\frac{1}{mK}\sum_{i,\tau}\mathbb{E}\Vert\mathbf{m}_{i,\tau}^{t}\Vert^{2}\Big)\\
    &\overset{(a)}{\leq} \Big(\frac{2\eta^{2}\eta_{l}^{2}(m-S)LG_{\vartheta}^{2}}{S(m-1)m}\sum_{i}\mathbb{E}_{K|t}\Vert\frac{1}{K}\sum_{\tau}\overline{\mathbf{m}}_{i,\tau}^{t}\Vert^{2}- \frac{(1+\mu)\beta_{1}\eta\eta_{l}\min\{G_{\vartheta}^{2}\}}{16G_{\vartheta}}\frac{1}{m}\sum_{i}\mathbb{E}\Vert\frac{1}{K}\sum_{\tau}\mathbf{m}_{i,\tau}^{t}\Vert^{2}\Big)\\
    &\overset{(b)}{=} \Big(\frac{2\eta^{2}\eta_{l}^{2}(m-S)LG_{\vartheta}^{2}}{S(m-1)}- \frac{(1+\mu)\beta_{1}\eta\eta_{l}\min\{G_{\vartheta}^{2}\}}{16G_{\vartheta}}\Big)\frac{1}{m}\sum_{i}\mathbb{E}\Vert\frac{1}{K}\sum_{\tau}\mathbf{m}_{i,\tau}^{t}\Vert^{2}\\
    &\quad -\frac{2\eta^{2}\eta_{l}^{2}(m-S)LG_{\vartheta}^{2}}{S(m-1)m}\sum_{i}\mathbb{E}_{K|t}\Vert\frac{1}{K}\sum_{\tau}\mathbf{m}_{i,\tau}^{t}-\overline{\mathbf{m}}_{i,\tau}^{t}\Vert^{2}\\
    &\leq \Big(\frac{2\eta^{2}\eta_{l}^{2}(m-S)LG_{\vartheta}^{2}}{S(m-1)}- \frac{(1+\mu)\beta_{1}\eta\eta_{l}\min\{G_{\vartheta}^{2}\}}{16G_{\vartheta}}\Big)\frac{1}{m}\sum_{i}\mathbb{E}\Vert\frac{1}{K}\sum_{\tau}\mathbf{m}_{i,\tau}^{t}\Vert^{2}.\\
\end{align*}
Combining the two inequalities above, we have:
\begin{align*}
    \mathbb{E}[F(\mathbf{z}^{t+1})] 
    &\leq \mathbb{E}[F(\mathbf{z}^{t})] + \Big(\frac{2\eta^{2}\eta_{l}^{2}(m-S)LG_{\vartheta}^{2}}{S(m-1)}- \frac{(1+\mu)\beta_{1}\eta\eta_{l}\min\{G_{\vartheta}^{2}\}}{16G_{\vartheta}}\Big)\frac{1}{m}\sum_{i}\mathbb{E}\Vert\frac{1}{K}\sum_{\tau}\mathbf{m}_{i,\tau}^{t}\Vert^{2}\\
    &\quad +\Big(2L\eta^{2}\eta_{l}^{2}-\frac{\eta\eta_{l}}{2}\Big)\mathbb{E}\Vert\frac{1}{mK}\sum_{i,\tau}\overline{\mathbf{m}}_{i,\tau}^{t}\odot\vartheta_{i,0}^{t}\Vert^{2} + \Big(\frac{\eta^{2}\eta_{l}^{2}G_{g}^{2}L}{\epsilon_{v}^{4}}+\frac{\eta\eta_{l}G_{g}^{2}}{2\epsilon_{v}^{3}}\Big)\mathbb{E}\sum_{j}^{d} \big((\overline{\hat{\mathbf{v}}}_{K}^{t})_{(j)}-(\overline{\hat{\mathbf{v}}}_{0}^{t})_{(j)}\big)\\
    &\quad + 2\big(\eta\eta_{l}^{3}(1+\mu)G_{\vartheta}L^{2}C_{2}+\eta^{3}\eta_{l}^{3}(1+\frac{1}{\mu})(\frac{1-\alpha}{\alpha})^{2}G_{\vartheta}L^{2}\big)\Big(\frac{dG_{g}^{4}}{\epsilon_{v}^{4}} + \frac{\beta_{1}^{2}G_{\vartheta}^{2}\sigma_{l}^{2}}{SK} + 2G_{\vartheta}^{2}(\sigma_{g}^{2}+\sigma_{u}^{2})\Big)\\
    &\quad + \frac{\eta\eta_{l}}{2}(\frac{\beta_{1}}{(1-\beta_{1})K}-1)G_{\vartheta}\mathbb{E}\Vert\nabla F(\mathbf{z}^{t})\Vert^{2} + \frac{(1+\mu)G_{\vartheta}L^{2}C_{3}\eta\eta_{l}^{3}\sigma^{2}}{2}+\frac{2\eta^{2}\eta_{l}^{2}\beta_{1}^{2}G_{\vartheta}^{2}L\sigma_{l}^{2}}{SK}.\\
\end{align*}

Here we make some special and reasonable limitations on $\eta\eta_{l}$ and $\eta_{l}$ to simplify some of the terms in the above formula. Let $\Big(2L\eta^{2}\eta_{l}^{2}-\frac{\eta\eta_{l}}{2}\Big) \leq 0$ which means $\eta\eta_{l} \leq \frac{1}{4L}$. This is a very general assumption in the proof of convergence of stochastic optimization on convex cases. Also let $\Big(\frac{2\eta^{2}\eta_{l}^{2}(m-S)LG_{\vartheta}^{2}}{S(m-1)}- \frac{(1+\mu)\beta_{1}\eta\eta_{l}\min\{G_{\vartheta}^{2}\}}{16G_{\vartheta}}\Big) \leq 0$ which means $\eta\eta_{l} \leq \frac{\beta_{1}S(m-1)(1+\mu)\min\{G_{\vartheta}^{2}\}}{32(m-S)LG_{\vartheta}^{3}}\leq\frac{\beta_{1}S(m-1)(1+\mu)}{32(m-S)LG_{\vartheta}}$ where $\mu \ > \ 0$. Later we will discuss the choice of the value of $\mu$. Let $\beta_{1} \leq \frac{K}{K+1}$, which keeps the term $(\frac{\beta_{1}}{(1-\beta_{1})K}-1)$ is negative. In the algorithm\ref{algorithm_fedLada}, the momentum $\mathbf{m}$ is set to $0$ at the initial stage of each local training stage. This setting also indirectly requires that the value of local iteration $K$ should be large enough to ensure the effective training. In fact, a simple conclusion can be made from the update of $\mathbf{m}_{i,\tau}^{t}=\beta_{1}\mathbf{m}_{i,\tau-1}^{t}+(1-\beta_1)g_{i,\tau}^{t}$. Supposing stochastic gradient $g_{i,\tau}^{t}$ can take its maximum value at each iteration $\tau$, the local training iterations $K$ requires at least $\frac{1}{1-\beta_{1}}$ updates to ensure $\Vert m\Vert^{2}\approx\Vert g\Vert^{2}$. In order to achieve the better performance, in our experiments we set $K \geq \frac{2}{1-\beta_{1}}$. Under the above conditions, let $C_{4}=\frac{1}{2}(1-\frac{\beta_{1}}{(1-\beta_{1})K}) \ > \ 0$ is a constant by choose a positive $\beta_{1}$ and $K$, we have:
\begin{align*}
    &\quad \eta\eta_{l}C_{4}G_{\vartheta}\mathbb{E}\Vert\nabla F(\mathbf{z}^{t})\Vert^{2}\\
    &\leq \Big(\mathbb{E}[F(\mathbf{z}^{t})] - \mathbb{E}[F(\mathbf{z}^{t+1})]\Big) + \frac{2\eta^{2}\eta_{l}^{2}\beta_{1}^{2}G_{\vartheta}^{2}L\sigma_{l}^{2}}{SK}+\Big(\frac{\eta^{2}\eta_{l}^{2}G_{g}^{2}L}{\epsilon_{v}^{4}}+\frac{\eta\eta_{l}G_{g}^{2}}{2\epsilon_{v}^{3}}\Big)\mathbb{E}\sum_{j}^{d} \big((\overline{\hat{\mathbf{v}}}_{K}^{t})_{(j)}-(\overline{\hat{\mathbf{v}}}_{0}^{t})_{(j)}\big)\\
    &\quad + 2\big(\eta\eta_{l}^{3}(1+\mu)G_{\vartheta}L^{2}C_{2}+\eta^{3}\eta_{l}^{3}(1+\frac{1}{\mu})(\frac{1-\alpha}{\alpha})^{2}G_{\vartheta}L^{2}\big)\Big(\frac{dG_{g}^{4}}{\epsilon_{v}^{4}} + \frac{\beta_{1}^{2}G_{\vartheta}^{2}\sigma_{l}^{2}}{SK} + 2G_{\vartheta}^{2}(\sigma_{g}^{2}+\sigma_{u}^{2})\Big)\\
    &\quad + \frac{(1+\mu)G_{\vartheta}L^{2}C_{3}\eta\eta_{l}^{3}\sigma^{2}}{2}.\\
\end{align*}
Taking the recursion on $t$ and we have:
\begin{align*}
    &\quad \frac{1}{T}\sum_{t=0}^{T-1}\mathbb{E}\Vert\nabla F(\mathbf{z}^{t})\Vert^{2}\\
    &\leq \frac{\big(F(\mathbf{z}^{0})-f_{*}\big) - \big(F(\mathbf{z}^{T})-f_{*}\big)}{\eta\eta_{l}C_{4}G_{\vartheta}T} + \frac{2\eta\eta_{l}\beta_{1}^{2}G_{\vartheta}L\sigma_{l}^{2}}{C_{4}SK}+\Big(\frac{\eta\eta_{l}G_{g}^{2}L}{\epsilon_{v}^{4}C_{4}G_{\vartheta}T}+\frac{G_{g}^{2}}{2\epsilon_{v}^{3}C_{4}G_{\vartheta}T}\Big)\mathbb{E}\sum_{j}^{d} \big((\overline{\hat{\mathbf{v}}}_{K}^{T-1})_{(j)}-(\overline{\hat{\mathbf{v}}}_{0}^{0})_{(j)}\big)\\
    &\quad + \frac{\big((1+\mu)L^{2}C_{2}\eta_{l}^{2}+(1+\frac{1}{\mu})(\frac{1-\alpha}{\alpha})^{2}L^{2}\eta^{2}\eta_{l}^{2}\big)}{C_{4}} \Big(\frac{dG_{g}^{4}}{\epsilon_{v}^{4}} + \frac{\beta_{1}^{2}G_{\vartheta}^{2}\sigma_{l}^{2}}{SK} + 2G_{\vartheta}^{2}(\sigma_{g}^{2}+\sigma_{u}^{2})\Big)+ \frac{(1+\mu)L^{2}C_{3}\eta_{l}^{2}\sigma^{2}}{2C_{4}}\\
    &\leq \frac{F(\mathbf{z}^{0})-f_{*}}{\eta\eta_{l}C_{4}G_{\vartheta}T} + \eta\eta_{l}\frac{\beta_{1}^{2}G_{\vartheta}L\sigma_{l}^{2}}{C_{4}SK}+\eta\eta_{l}\frac{dG_{g}^{4}L}{\epsilon_{v}^{4}C_{4}G_{\vartheta}T}+\frac{dG_{g}^{4}}{2\epsilon_{v}^{3}C_{4}G_{\vartheta}T} + \eta_{l}^{2}\frac{(1+\mu)L^{2}C_{3}\sigma^{2}}{2C_{4}}\\
    &\quad + \frac{\big((1+\mu)L^{2}C_{2}\eta_{l}^{2}+(1+\frac{1}{\mu})(\frac{1-\alpha}{\alpha})^{2}L^{2}\eta^{2}\eta_{l}^{2}\big)}{C_{4}} \Big(\frac{dG_{g}^{4}}{\epsilon_{v}^{4}} + \frac{\beta_{1}^{2}G_{\vartheta}^{2}\sigma_{l}^{2}}{SK} + 2G_{\vartheta}^{2}(\sigma_{g}^{2}+\sigma_{u}^{2})\Big),\\
\end{align*}
where $f_{*}$ is the minimum value of the function $F$ and other constants are defined in the notation. Combining all the conditions above, let $\eta\eta_{l}$ satisfies $\eta\eta_{l} \leq \min\{\frac{1}{4L}, \frac{\beta_{1}S(m-1)(1+\mu)}{32(m-S)LG_{\vartheta}}\}$ as $\mu=1$ and we use $\eta\eta_{l}=\mathbf{O}(\frac{\sqrt{SK}}{\sqrt{T}})$; let $\eta_{l}$ satisfies $\eta_{l} \leq \frac{1}{2\sqrt{2\alpha(1-\beta_{1})(1+\frac{1}{a})}G_{\vartheta}L}\ < \ \frac{1}{2\sqrt{2\alpha(1-\beta_{1})K}G_{\vartheta}L}$ and we use $\eta_{l}=\mathbf{O}(\frac{1}{\sqrt{KT}})$, we have:
\begin{align*}
    &\quad \frac{1}{T}\sum_{t=0}^{T-1}\mathbb{E}\Vert\nabla F(\mathbf{z}^{t})\Vert^{2}\\
    &= \mathbf{O}\Big(\frac{\mathbb{E}F(\mathbf{z}^{0})-f_{*}}{\sqrt{SKT}} + \frac{\sigma_{l}^{2}}{\sqrt{SKT}} + \frac{\sigma^{2}}{KT} + dG_{g}^{3}(\frac{1}{T}+\frac{\sqrt{SK}}{T^{3/2}}) + (dG_{g}^{4}+\frac{\sigma_{l}^{2}+K(\sigma_{g}^{2}+\sigma_{u}^{2})}{S})(\frac{1}{KT} + \frac{SK}{T})\Big)\\
    &= \mathbf{O}\Big( \frac{1}{\sqrt{SKT}} + \frac{1}{KT} + \frac{1}{T} + \frac{1}{T^{3/2}} \Big).\\
\end{align*}

\newpage
\section{Additional Discussion}

\subsection{Discussion of the Convergence}
\begin{table}[t]
%\small % if necessary
  \caption{Convergence rate for non-convex smooth cases in some baselines and our proposed FedLADA.}
  \label{analysis_comparison}
  \centering
  \begin{threeparttable}
  \begin{tabular}{ccc}
    \toprule
    Method     & Convergence     & Additional Assumption  \\
    \midrule
    FedAvg\cite{favg_convergence1} & $\mathcal{O}\left(\frac{\sqrt{K}}{\sqrt{ST}}+\frac{1}{T}\right)$ &   -   \\
    FedAdam\cite{fedadam}     & $\mathcal{O}\left(\frac{1}{\sqrt{SKT}}+\frac{1}{T}+\frac{\sqrt{S}}{\sqrt{KT^{3}}}\right)$ & -      \\
    \!\!\!\!SCAFFOLD\cite{SCAFFOLD}\cite{favg_convergence1}     & $\mathcal{O}\left(\frac{1}{\sqrt{SKT}}+\frac{1}{T}\right)$ & -      \\
    FedCM\cite{FedCM}     & $\mathcal{O}\left(\sqrt{\frac{1}{ST}+\frac{1}{SKT}}\!+\!\sqrt[3]{\frac{1}{T^{2}}+\frac{1}{ST^{2}}+\frac{1}{SKT^{2}}}\right)$ & -     \\
    \midrule
    FedProx\cite{FedProx}     & $\mathcal{O}(\frac{1}{T})$ &   
    local exact solution   \\
    FedPD\cite{FedPD}     & $\mathcal{O}\left(\frac{1}{T}+\epsilon\right)$ &  $\text{local $\epsilon$-stationarity}^{2}$     \\
    FedDyn\cite{FedDyn}     & $\mathcal{O}\left(\frac{1}{T}\right)$ & local exact solution      \\
    \midrule
    FedLADA     & $\mathcal{O}\left(\frac{1}{\sqrt{SKT}}+\frac{1}{T}+\frac{1}{\sqrt{T^{3}}}\right)$ & -      \\
    \bottomrule
  \end{tabular}
  \vskip 0.02in
  \begin{tablenotes}
    \item[1] S : the number of clients, $K$ : the local interval, $T$ : the communication round.
    \item[2] solve the local sub-problem $F_{i}(\mathbf{x})$ to satisfy $\Vert\nabla\mathcal{L}_{i}(\mathbf{x}^{t})\Vert^{2}\leq \epsilon$.\\\\
  \end{tablenotes}
  \end{threeparttable}
  \vspace{-0.5cm}
\end{table}

For the general FedAvg method, under partial participation, the dominant term of the final convergence is $\mathcal{O}(\frac{\sqrt{K}}{\sqrt{ST}})$ which includes the stochastic variance $\sigma_l^2$ and the initialization bias $F(\mathbf{x^0})-F(\mathbf{x^\star})$. FedAdam method proves that the dominant term could be accelerated by the local interval $K$. However, the third term is affected by the number of clients $S$. In practical FL scenarios, this number could be very large. SCAFFOLD uses the variance reduction technique to correct the local bias, which achieves the same dominant term. Moreover, it releases the negative impact of the $S$. FedProx, FedPD, and FedDyn could achieve the fast $\mathcal{O}(\frac{1}{T})$ rate under the specific assumption of local exact-solution. They all require each client to approach the local optimum on each communication round. This is a strict condition that requires the local interval $K$ must be selected long enough. However, from the perspective of the communication round $T$, this type of method reduces the communication cost under the same convergence rate. FedLADA uses the locally amended adaptive optimizer to achieve the same dominant term $\mathcal{O}(\frac{1}{\sqrt{SKT}})$ with the linear speedup property. Furthermore, it expands the application of adaptive optimizers in FL. Compared to the FedAdam, it also releases the negative impact of the $S$.

\subsection{Discussion of the Correction Efficiency}
SCAFFOLD~\cite{SCAFFOLD} focuses on adopting the SGD optimizer on the local client, while FedLADA focuses on adopting the adaptive optimizer on the local client, which is the essential difference between these two methods. To generally compare the correction method in FedLADA and SCAFFOLD, we assume that the update $\mathbf{m}_{i,\tau}^t/\sqrt{\hat{v}_{i,\tau}^t}$ in adaptive optimizer plays the same role as the stochastic gradient in SGD. We use $g_{i,\tau}^{t}$ to represent the local update and use $\mathbf{g}_a^t$ to represent the global estimation broadly.
Differently, SCAFFOLD~\cite{SCAFFOLD} uses the variance reduction technique which corrects the local gradient as:
\begin{equation}
\label{scaffold}
    g_{\text{correct}} = g_{i,\tau}^t + \mathbf{g}_a^t - \overline{g}_{i}^{t-1},
\end{equation}
where $\overline{\mathbf{g}}_{i}^{t-1}$ is the average of the local updates on communication round $t-1$.\\
While FedLADA uses the local amended technique as:
\begin{equation}
\label{amended}
    g_{\text{correct}} = \alpha g_{i,\tau}^t + (1-\alpha)\mathbf{g}_a^t = g_{i,\tau}^t + (1-\alpha)(\mathbf{g}_a^t - g_{i,\tau}^t).
\end{equation}
SCAFFOLD~\cite{SCAFFOLD} indicates that the global estimation $\mathbf{g}_a$ is good enough, which could provide accurate guidance for the local updates~(it could be approximated as $\mathbf{g}_a^{t+1}\approx\mathbf{g}_a^{t}$). From the perspective of optimization, it adopts the SVRG on the local clients, while the past gradient in SVRG is replaced by the average of local updates. Its correction is fixed by the difference between the average of the global update and the local updates of the last round. FedLADA adopts a flexible manner whose correction is aimed to approach the global estimation. It uses a coefficient $\alpha$ to control the level of the correction. The advantages of this flexible correction include:\\\\
\textbf{(a) Reduce the impact of the inaccurate estimation:} Equation~(\ref{scaffold}) use the difference term $\mathbf{g}_a^t - \overline{g}_{i}^{t-1}$ to correct the current gradient. It implies that the local difference should maintain a high similarity condition, which is:
\begin{equation}
\label{scaffold_condition}
    \mathbf{g}_a^{t+1} - \overline{g}_{i}^{t} \approx \mathbf{g}_a^t - \overline{g}_{i}^{t-1}.
\end{equation}
Under Equation~(\ref{scaffold_condition}), for the current gradients, SCAFFOLD will approximate a very accurate global estimation on the local client. However, satisfying this condition is a little difficult in the FL paradigm. Equation~(\ref{scaffold_condition}) requires the following relationship:
\begin{equation}
    \mathbf{g}_a^{t+1} - \mathbf{g}_a^t \approx \overline{g}_{i}^{t} - \overline{g}_{i}^{t-1}.
\end{equation}
Due to the local heterogeneity, it is hard to measure these two terms. And generally, the difference of the global update between the two adjacent rounds is usually far away from it of the local update because their objectives are different. To reduce the impact of this inaccurate estimation, FedLADA expects that $g_{i.\tau}^t$ could be close enough to the global estimation $\mathbf{g}_a$. It uses a weighted average of the local update and global estimation to force local updates toward the global trajectory. Thus, FedLADA only relies on the condition that global estimation $\mathbf{g}_a$ is good enough (which could be considered as $\mathbf{g}_a^{t+1}\approx \mathbf{g}_a^{t}$). It would not be affected by the dissimilarity between the $\mathbf{g}_a^{t+1} - \mathbf{g}_a^t$ and $\overline{g}_{i}^{t} - \overline{g}_{i}^{t-1}$ terms.\\\\
\begin{figure}[t]
    \centering
    \setlength{\abovecaptionskip}{0.cm}
	\begin{minipage}[t]{0.4\linewidth}
		\centering
		\subfigure[Local Correction Schematic.]{
		\includegraphics[width=0.99\linewidth]{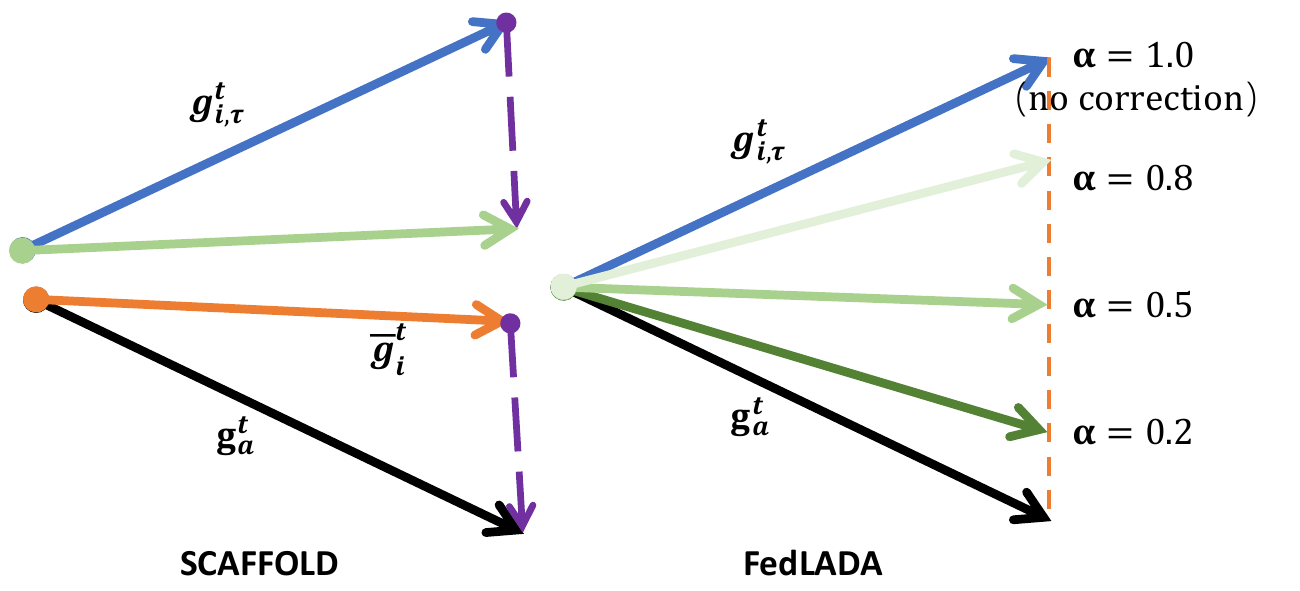}
		}
	\end{minipage}
	\begin{minipage}[t]{0.264\linewidth}
		\centering
		\subfigure[Test Accuracy.]{
		\includegraphics[width=0.99\linewidth]{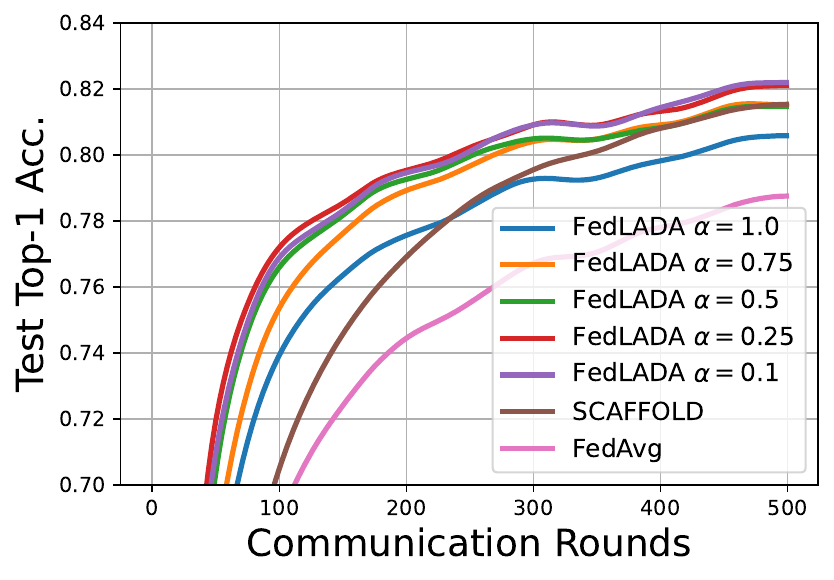}
		}
	\end{minipage}
 `      \begin{minipage}[t]{0.27\linewidth}
		\centering
		\subfigure[Consistency $\mathbb{E}\Vert\mathbf{x}_{i,K}^t-\mathbf{x}^{t+1}\Vert^2$.]{
		\includegraphics[width=0.99\linewidth]{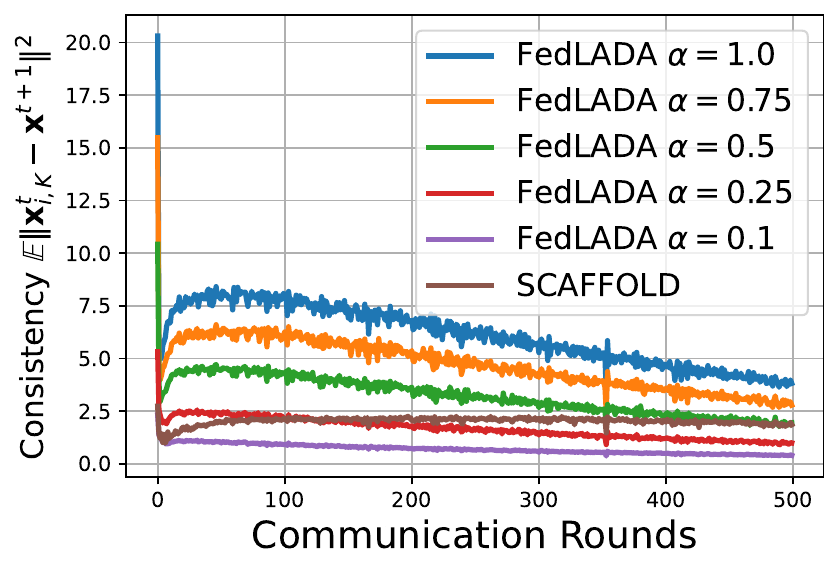}
		}
	\end{minipage}
	\vskip -0.05in
	\caption{(a) A simple schematic to introduce the difference between the correction of SCAFFOLD and FedLADA. (b) Comparison of test accuracy of different $\alpha$ in FedLADA with the SCAFFOLD method. (c) Comparison of consistency term of different $\alpha$ in FedLADA with the SCAFFOLD method.}
	\label{different_alpha}
	\vskip -0.1in
\end{figure}
\textbf{(b) The $\alpha$ coefficient help to improve the local consistency:} Consistency is very important in the FL paradigm. SCAFFOLD~\cite{SCAFFOLD} has studied that the ``client-drift" problem affects the performance of the FL framework seriously. The essential problem in ``client-drift" is that local objective is far away from each other due to the unknown local heterogeneity. Although SCAFFOLD has made corrections to reduce ``client-drift", there's still a lot of inconsistency across the local models. One success of the proposed FedLADA is that it uses the $\alpha$ to maintain very high consistency in practical training. We test different setups on CIFAR-10 dataset Dir-0.6 splitting as shown in Figure~\ref{different_alpha}. Figure~\ref{different_alpha} (b) indicates that when $\alpha$ is small enough, the performance of the correction in FedLADA outperforms SCAFFOLD. We also calculate the consistency term $\mathbb{E}\Vert\mathbf{x}_{i,K}^t-\mathbf{x}^{t+1}\Vert^2$ to bound the divergence level of the local solutions~(when global learning rate $\eta_g=1.0$, we have $\mathbf{x}^{t+1}=\frac{1}{S}\sum_{i}\mathbf{x}_{i,K}^t$). Figure~\ref{different_alpha} (c) indicates that small $\alpha$ leads to highly consistent local solutions. Due to these good properties, the global correction in FedLADA is better than it in SCAFFOLD. Actually, it uses the similarity of the adjacent global updates to maintain higher consistency during the total training process, which could be considered to be close to the centralized training.\\\\
\textbf{(c) Not a simple $\alpha$ can improve the performance:} Furthermore, we also test the following correction in SCAFFOLD:
\begin{equation}
\label{add_alpha}
    g_{\text{correct}} = g_{i,\tau}^t + \alpha(\mathbf{g}_a^t - \overline{g}_{i}^{t-1}),
\end{equation}
where we add a scaled coefficient $\alpha$ similar to it in FedLADA to explore its efficiency. The results of different $\alpha$ are shown in the following table.
\begin{table}[H]
\centering
%\vspace{-0.63cm}
%\small
%\renewcommand{\arraystretch}{1}
\caption{We test performance of adding scaled coefficient $\alpha$ in SCAFFOLD as introduced in Equation~(\ref{add_alpha}) under the same experimental setups as Fig~\ref{different_alpha}.}
%\vspace{0.1cm}
\label{scaffold_consistency}
%\begin{sc}
%\scalebox{0.9}{
\setlength{\tabcolsep}{1.1mm}{\begin{tabular}{@{}cccccccc@{}}
\toprule
    $\alpha$ & 0.0 & 0.25 & 0.5 & 0.75 & \textbf{1.0} & 1.25 & 1.5\\
    \midrule
    Accuracy~(\%)   & 78.32 & 80.74 & 81.09 & 81.72 & \textbf{81.77} & 81.05 & 80.62 \\
    \bottomrule
\end{tabular}}
%}
\end{table}
As shown in Table~\ref{scaffold_consistency}, $\alpha=0.0$ means the vanilla FedAvg method and $\alpha=1.0$ means the SCAFFOLD method. We can see the best selection of $\alpha$ is $1.0$. The main reason is that the local biases are different from the global biases. The scaled coefficient $\alpha$ can not effectively correct the local bias. Our proposed correction uses the weighted average to force the local update to be close to the global estimation which is not affected by the difference between the global bias and local bias.

\subsection{Discussion of the Communication Cost}
\begin{table}[h]
  \caption{Communication costs per communication round.}
  %\vskip 0.05in
  \label{appendix:communication}
  \centering
  \small
  \begin{tabular}{c|cr}
    \toprule
    Method & communication & ratio \\
    \midrule
    FedAvg    & $Nd$  & 1$\times$ \\
    FedAdam   & $Nd$  & 1$\times$ \\
    SCAFFOLD  & $2Nd$ & 2$\times$ \\
    FedCM     & $2Nd$ & 2$\times$ \\
    FedProx   & $Nd$  & 1$\times$ \\
    LocalAdam & $2Nd$ & 2$\times$ \\
    FedLADA   & $2Nd$ & 2$\times$ \\
    \bottomrule
  \end{tabular}
\end{table}

It could be seen in Table~\ref{appendix:communication}, the methods with correction, i.e. SCAFFOLD, FedCM, and FedLADA, require at least double the communication cost. However, we also calculate the total communication cost. We test the total communication rounds required to achieve the target accuracy on CIFAR-10 for each baseline in Table~\ref{appendix:com},
\begin{table}[h]
  \caption{Total rounds and total communication costs to achieve the target accuracy.}
  %\vskip 0.05in
  \label{appendix:com}
  \centering
  \small
  \begin{tabular}{c|ccr|ccr}
    \toprule
    Method & achieve 70\% & communication & ratio & achieve 80\% & communication & ratio \\
    \midrule
    FedAvg    & 94 & 94$Nd$  & 1$\times$ & 537 & 537$Nd$ & 1$\times$ \\
    FedAdam   & 96 & 96$Nd$  & 1.02$\times$ & 611 & 611$Nd$ & 1.13$\times$ \\
    SCAFFOLD  & 85 & 170$Nd$ & 1.80$\times$ & 262 & 524$Nd$ & 0.97$\times$ \\
    FedCM     & 71 & 142$Nd$ & 1.51$\times$ & 286 & 572$Nd$ & 1.06$\times$ \\
    FedProx   & 91 & 91$Nd$  & 0.97$\times$ & 511 & 511$Nd$ & 0.95$\times$ \\
    LocalAdam & 65 & 130$Nd$ & 1.38$\times$ & - & - & - \\
    FedLADA   & 45 & 90$Nd$  & \textbf{0.95$\times$} & 186 & 372$Nd$ & \textbf{0.69$\times$} \\
    \bottomrule
  \end{tabular}
\end{table}

In the low-precision phase, the local-SGD-based algorithms with local corrections, i.e. SCAFFOLD and FedCM obviously require more communication bits for their double costs. In the high-precision phase, the model is good enough, and the advantages of the correction help the local models to be close to the global one and significantly improve their performance. However, the FedLADA uses the local adaptive optimizer which could maintain high efficiency in the total training process. Though the communication costs per round are still doubled, it always requires fewer communication rounds for training. This is also one of the main successes of FedLADA.

\subsection{Disscussion of Some Concepts}
\textbf{Rugged Convergence on Global Adaptive.}
The adaptive optimizer often uses an additional vector or even a preconditioner to scale the gradients to achieve better performance. It demonstrates that some large values of the gradients bring an imbalance on the updates and some dimensions of the models are over-trained. Generally, it uses the square root of the accumulation of squared gradients to scale the stochastic gradients, and this has worked very well in many deep training tasks. When it is transferred to the global adaptive optimizer in the FL paradigm, it adopts the total local updates as the quasi-gradient on the global server. As mentioned in \cite{local_adaptive,local_adaAlter}, these quasi-gradients on the global server introduce biases to the scaled vector and usually reduce the precision of the training. These inexact global quasi-gradients lead to the inexact second-order momenta, which further brings the rugged convergence on the global server. We also validate this phenomenon in the practical training process. Though its final performance could be higher than the vanilla FedAvg, the global adaptive optimizer often requires more communication rounds to achieve the same accuracy.\\\\
\textbf{Client Drifts.} This is proposed in \cite{SCAFFOLD} that the local optimums are different and far away across the local clients due to the local heterogeneity. After the local training process, each local model is trained to fit the local dataset. The heterogeneous dataset yields huge gaps between aggregated local optimum and global optimum as client drifts mentioned in \cite{SCAFFOLD} by $\mathbf{x}^{*}\neq \frac{1}{m}\sum_{i}\mathbf{x}_{i}^{*}$ where $\mathbf{x}^{*}$ represents for the optimum of the objective function. Therefore, the local heterogeneous dataset causes the unsatisfactory performance of the global model.

\end{document}